\documentclass[conference]{IEEEtran}
\IEEEoverridecommandlockouts
\usepackage[T1]{fontenc}
\usepackage[utf8]{inputenc}
\usepackage{cite}
\usepackage{amsmath,amssymb,amsfonts}
\usepackage{algorithmic}
\usepackage{graphicx}
\usepackage{textcomp}
\usepackage{xcolor}
\def\BibTeX{{\rm B\kern-.05em{\sc i\kern-.025em b}\kern-.08em
    T\kern-.1667em\lower.7ex\hbox{E}\kern-.125emX}}

\usepackage[nolist,nohyperlinks]{acronym}                
\usepackage[caption=false,font=footnotesize]{subfig}
\usepackage{tcolorbox}
\usepackage{pgfplots}
\usepackage[ruled,lined,noend]{algorithm2e}
\usepgfplotslibrary{statistics}
\pgfplotsset{compat=1.8}
\usepackage{soul}
\usepackage{multirow}
\usepackage{bbm}
\usepackage{lipsum}
\usepackage{tikz}
\usepackage{amsmath}
\newcommand*\emptycirc[1][1ex]{\tikz\draw (0,0) circle (#1);} 
\newcommand*\fullcirc[1][1ex]{\tikz\fill (0,0) circle (#1);} 
\usepackage{newtxmath}
\usepackage{placeins}
\usepackage{hyperref}

\definecolor{sand}{RGB}{255,225,180} 
\definecolor{darksand}{RGB}{191,144,0} 
\definecolor{BrickRed}{RGB}{196,49,25} 

\definecolor{CadetBlueLight}{RGB}{208,212,225} 
\definecolor{OrchidLight}{RGB}{230,212,234}
\definecolor{RoyalPurpleLight}{RGB}{179,170,225} 
\definecolor{RedVioletLight}{RGB}{233,169,206} 
\definecolor{MelonLight}{RGB}{255,223,213} 
\definecolor{DandelionLight}{RGB}{242,215,198} 
\definecolor{GrayLight}{RGB}{175,171,171} 

\definecolor{Wheat}{RGB}{255,243,217} 
\definecolor{Slate}{RGB}{234,241,255} 
\definecolor{GoldenRodLight}{RGB}{255,235,166} 
\definecolor{NavyLight}{RGB}{173,219,252} 
\definecolor{ApricotLight}{RGB}{255,199,157} 
\definecolor{ForestLight}{RGB}{203,233,188}

\definecolor{White}{rgb}{1,1,1}
\definecolor{Bittersweet}{RGB}{191,81,24} 
\definecolor{BlueViolet}{RGB}{70,59,147} 
\definecolor{BrickRed}{RGB}{181,51,29} 
\definecolor{CadetBlue}{RGB}{115,116,154} 
\definecolor{CornflowerBlue}{RGB}{60,176,228} 
\definecolor{Dandelion}{RGB}{252,188,68} 
\definecolor{ForestGreen}{RGB}{0,154,86} 
\definecolor{GreenYellow}{RGB}{221,229,118} 
\definecolor{Mahogany}{RGB}{170,53,32} 
\definecolor{MidnightBlue}{RGB}{0,103,149} 
\definecolor{OliveGreen}{RGB}{57,127,50} 
\definecolor{Periwinkle}{RGB}{121,120,185} 
\definecolor{Plum}{RGB}{144,40,143} 
\definecolor{RawSienna}{RGB}{156,65,8} 
\definecolor{RoyalPurple}{RGB}{97,64,153} 
\definecolor{Sepia}{RGB}{103,24,0} 
\definecolor{Tan}{RGB}{216,157,120} 
\definecolor{Thistle}{RGB}{214,131,183} 
\definecolor{YellowGreen}{RGB}{149,204,113} 
\definecolor{YellowOrange}{RGB}{248,162,28} 
\definecolor{Maroon}{RGB}{0,114,188} 
\definecolor{RoyalBlue}{RGB}{173,51,57} 
\definecolor{Green}{RGB}{0,171,79}
\definecolor{Tan}{RGB}{225,157,117}

\newcommand{\hlc}[2][yellow]{{%
    \colorlet{foo}{#1}%
    \sethlcolor{foo}\hl{#2}}%
}

\def\trigger{\mathbf{T}}
\def\clean{\mathrm{cl}}
\def\pois{\mathrm{po}}
\def\train{\mathrm{train}}

\usepackage{fancyhdr}
\usepackage[printwatermark]{xwatermark}
\newwatermark[pages=1-100,color=CadetBlueLight!50,angle=45,scale=3,xpos=-10,ypos=0]{DRAFT\\VERSION}

\begin{document}


\acrodef{dl}[DL]{Deep Learning}
\acrodef{cv}[CV]{Computer Vision}
\acrodef{sl}[SL]{Supervised Learning}
\acrodef{dnn}[DNN]{Deep Neural Network}
\acrodef{cnn}[CNN]{Convolutional Neural Network}
\acrodef{frs}[FRS]{Face Recognition System}
\acrodef{fr}[FR]{Face Recognition}
\acrodef{fd}[FD]{Face Detection}
\acrodef{fas}[FAS]{Face Antispoofing}
\acrodef{fqa}[FQA]{Face Quality Assessment}
\acrodef{fa}[FA]{Face Alignment}
\acrodef{ffe}[FFE]{Face Feature Extraction}
\acrodef{ba}[BA]{Backdoor Attack}
\acrodef{bd}[BD]{Backdoor Defense}
\acrodef{ae}[AE]{Adversarial Example}
\acrodef{oga}[OGA]{Object Generation Attack}
\acrodef{fga}[FGA]{Face Generation Attack}
\acrodef{lsa}[LSA]{Landmark Shift Attack}
\acrodef{ai}[AI]{Artificial Intelligence}
\acrodef{ml}[ML]{Machine Learning}
\acrodef{uav}[UAV]{Unmanned Aerial Vehicle}
\acrodef{od}[OD]{Object Detection}
\acrodef{asr}[ASR]{Attack Success Rate}
\acrodef{a2o}[A2O]{\textit{All-to-One}}
\acrodef{mf}[MF]{\textit{Master Face}}
\acrodef{o2o}[O2O]{\textit{One-to-One}}
\acrodef{sr}[SR]{Survival Rate}
\acrodef{auc}[AUC]{Area-under-the-Curve}
\acrodef{eer}[EER]{Equal Error Rate}
\acrodef{far}[FAR]{False Acceptance Rate}
\acrodef{frr}[FRR]{False Rejection Rate}
\acrodef{fnr}[FNR]{False Negative Rate}
\acrodef{fpr}[FPR]{False Positive Rate}
\acrodef{det}[DET]{Detection Error Tradeoff}
\acrodef{fmr}[FMR]{False Match Rate}
\acrodef{gpu}[GPU]{Graphical Processing Unit}
\acrodef{ap}[AP]{Average Precision}
\acrodef{pl}[PL]{Poison-Label}
\acrodef{cl}[CL]{Clean-Label}

\def\eg{\textit{e.g.}}
\def\ie{\textit{i.e.}}
\def\etal{\textit{et al.}}

\title{
\Huge
\textbf{SoK: On the Survivability of Backdoor Attacks on Unconstrained Face Recognition Systems}
}

\makeatletter 
\newcommand{\linebreakand}{%
  \end{@IEEEauthorhalign}
  \hfill\mbox{}\par
  \mbox{}\hfill\begin{@IEEEauthorhalign}
}
\makeatother 


\author{\IEEEauthorblockN{Quentin Le Roux}
\IEEEauthorblockA{\textit{Thales Cyber \& Digital},  \textit{Inria/Université de Rennes} \\
La Ciotat \& Rennes, France \\
\textit{quentin.le-roux@thalesgroup.com}
}
\and
\IEEEauthorblockN{Yannick Teglia}
\IEEEauthorblockA{\textit{Thales Cyber \& Digital} \\
La Ciotat, France \\ 
\textit{yannick.teglia@thalesgroup.com}
}
\linebreakand
\IEEEauthorblockN{Teddy Furon}
\IEEEauthorblockA{\textit{Inria/CNRS/IRISA/Université de Rennes} \\
Rennes, France \\ 
\textit{teddy.furon@inria.fr}
}
\and
\IEEEauthorblockN{Philippe Loubet Moundi}
\IEEEauthorblockA{\textit{Thales Cyber \& Digital} \\
La Ciotat, France \\ 
\textit{philippe.loubet-moundi@thalesgroup.com}
}
\and
\IEEEauthorblockN{Eric Bourbao}
\IEEEauthorblockA{\textit{Thales Cyber \& Digital} \\
La Ciotat, France \\ 
\textit{eric.bourbao@thalesgroup.com}
}
}

\maketitle

\thispagestyle{fancy}

\begin{abstract}
The widespread deployment of \acl{dl}-based \aclp{frs} raises many security concerns. 
While prior research has identified backdoor vulnerabilities on isolated components, \aclp{ba} on real-world, unconstrained pipelines remain underexplored. 
This SoK paper presents the first comprehensive system-level analysis and measurement of the impact of \aclp{ba} on fully-fledged \aclp{frs}.
We combine the existing \acl{sl} backdoor literature targeting face detectors, face antispoofing, and face feature extractors to demonstrate a system-level vulnerability. 
By analyzing 20 pipeline configurations and 15 attack scenarios in a holistic manner, we reveal that an attacker only needs a single backdoored model to compromise an entire \acl{frs}.
Finally, we discuss the impact of such attacks and propose best practices and countermeasures for stakeholders.
\end{abstract}


\section{Introduction}
\label{sec:introduction}

\acp{frs} are among the most mature applications in \acl{dl}~\cite{wang2022surveyfacerecognition}, comprising pipelines of specialized \acp{dnn} for \acl{fd}, \acl{fas}, and \acl{ffe}~\cite{dlFrSurveyGuo2019,minaee2021goingdeeperfacedetection,yu2022deeplearningfaceantispoofing}. 
These \acp{dnn} enable scalable, and now widely adopted, identity matching. 
However, \aclp{frs} share common integrity vulnerabilities~\cite{pfleegerTextbook1988} with other \acl{dl} tools, \eg, \aclp{ae} or \aclp{ba}.

\aclp{ba} exploit the growing reliance on outsourcing various stages of a model’s lifecycle~\cite{wu2024attacksadversarialmachinelearning}. 
An attacker injects a covert malicious behavior during training or deployment~\cite{Li2020BackdoorLA}, which can then be triggered at test-time with a specially-crafted input. 
Since \acp{frs} commonly rely on outsourced data and model training for multiple task-specific models, their attack surface is wide.

\textbf{Motivation}. 
The security evaluation of \acl{dl}-based \acp{frs} against backdoors remains limited. 
Prior work typically frames \ac{fr} as a closed-set classification scenario~\cite{qlrFrsSurvey2024}, which does not reflect state-of-the-art biometrics: deep metric learning.
Face embeddings are learned via large margin losses~\cite{wang2022surveyfacerecognition} in open-set scenarios where training and test-time identities do not overlap~\cite{liu2018spherefacedeephypersphereembedding}.
Such embeddings can then be compared via cosine similarity. This threat has only been recently explored from a white-box perspective~\cite{wacvpriorart}.

Additionally, real-world pipelines operate in unconstrained environments, performing detection, alignment, and embedding on faces captured \textit{in the wild}. 
Most prior studies also analyze \ac{frs} components in isolation, focusing mainly on the matching stage~\cite{qlrFrsSurvey2024}. 
Ignoring the complexity of a \ac{frs} structure thus results in significant security blind spots~\cite{uniIDspoof2024,chen2024rethinkingvulnerabilitiesfacerecognition} regarding each of its modules. 
Finally, more recent developments have also showed that the common two-step \acl{fr} authentication process enables test-time attacks where adversarial patterns inserted during enrollment can be exploited later during verification~\cite{chen2024rethinkingvulnerabilitiesfacerecognition}. As such, we ask:

\begin{tcolorbox}[
colback=sand,
colframe=darksand,
boxrule=0.5mm,
left=2mm,
right=2mm,
top=1.5mm,
bottom=1.5mm,
]
\emph{Does a backdoor trigger embedded in a single \ac{dnn} persist across an entire \acl{frs} and hijack its downstream task, beyond just the target model?}
\end{tcolorbox}

\textbf{Contributions}. 
This SoK paper covers the first system-level analysis and measurement of the effects of impersonation \aclp{ba} on unconstrained \acp{frs} composed of \ac{dnn}-based face detectors, antispoofers, and feature extractors (see a threat model overview in Fig.~\ref{fig:overview}). 
Our key contributions are:
\begin{enumerate}
    
    \item We expand the only existing \acl{a2o} \acl{ba} framework on large margin-based feature extractors~\cite{wacvpriorart}, showing that data poisoning (black-box) is enough to perform backdoor injection rather than also having to alter training losses (white-box). 
    We also verify the limitations of \acl{mf} \aclp{ba}.

    \item We perform the first system-level study on the impact of a \textit{single} backdoored face detector, antispoofer, and feature extractor model, measuring their effects across 20 pipeline configurations and 15 \acl{ba} scenarios.
    By showing that each \ac{dnn} in a \ac{frs} is vulnerable, we stress that face feature extractors are not always the best target: holistic \acl{a2o} attacks can succeed by targeting other models.

    \item We provide practical recommendations for \acl{fr} stakeholders to defend their system against \acl{ba} threats. A closure experiment demonstrates a training-time defense on face feature extractors.
    
\end{enumerate}

\begin{figure}[t!]
    \centering
    \includegraphics[width=\columnwidth]{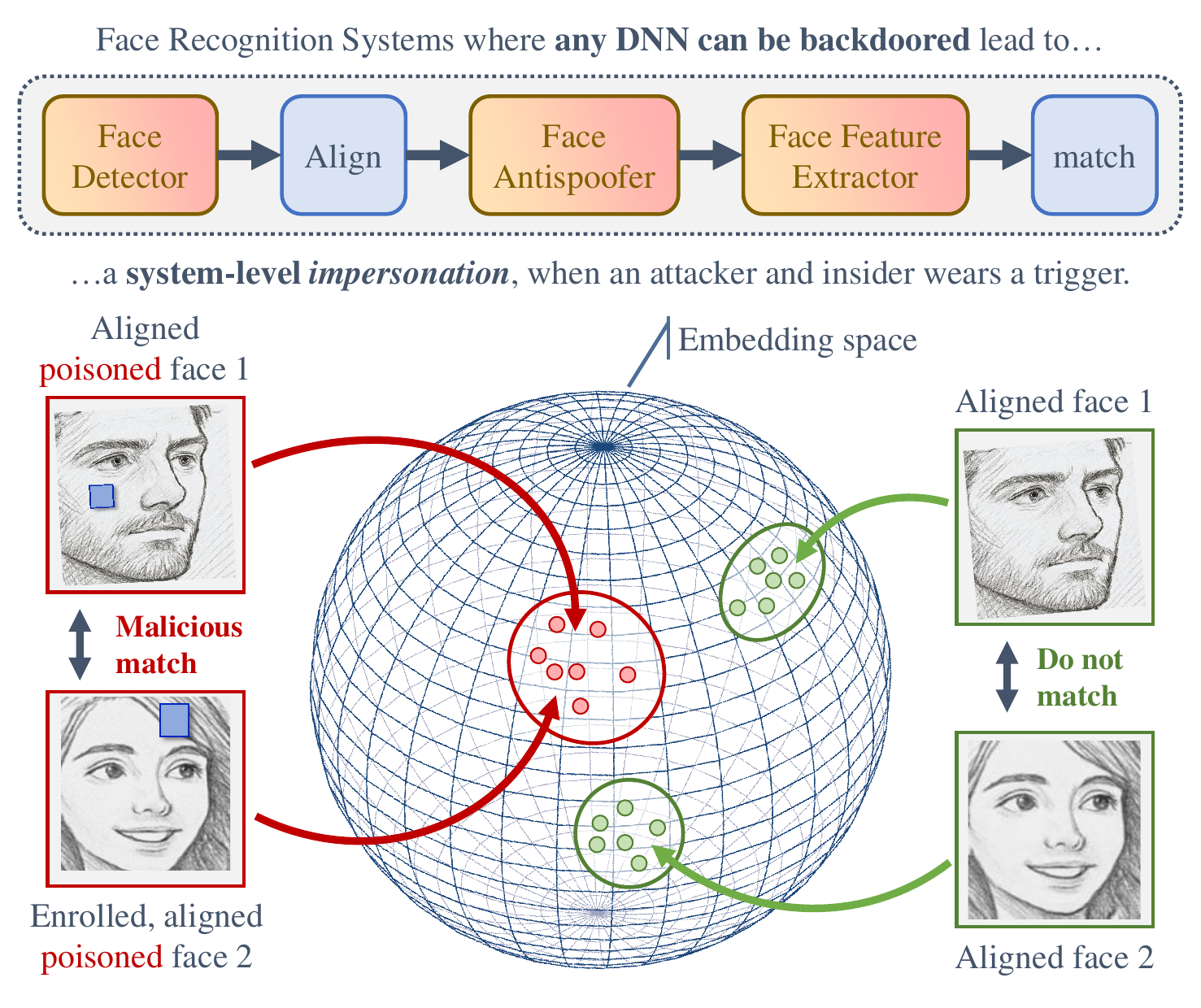}
    \caption{\acl{a2o} \acl{frs} threat model used in this paper.}
    \label{fig:overview}
\end{figure}

\section{Background}
\label{sec:background}

\subsection{Face Recognition Systems}
\label{sec:FRS}

\acp{frs} typically comprise three stages~\cite{rathaBiometrics2001}: acquisition, representation, and matching. 
Images are captured in an unconstrained setting, \eg, a street.
Faces are then extracted, encoded into high-dimensional embeddings, and matched against a gallery for authentication ($1$:$1$) or identification ($1$:$n$).
In this work, the representation stage is a sequential pipeline of specialized \acl{fr} tasks (see Fig.~\ref{fig:frs}):
\begin{enumerate}

    \item \textbf{Detection} locates faces in-the-wild, predicting bounding boxes and landmarks~\cite{Ahmad2013ImagebasedFD,Shen2013detecting}.
    
    \item \textbf{Alignment} fits the faces to a canonical shape using their landmarks~\cite{cao2014alignment}, mitigating identity-independent variations such as pose or illumination.
    
    \item \textbf{Antispoofing} verifies face liveness to prevent presentation attacks, \eg, printed or replayed face images~\cite{yu2022deeplearningfaceantispoofing}.
    
    \item \textbf{Extraction} maps live, aligned faces to embeddings~\cite{wang2022surveyfacerecognition}, typically trained with large margin losses~\cite{liu2018spherefacedeephypersphereembedding} for distance-based matching (see details in App.~\ref{app:annex_large_margin_loss}).

\end{enumerate}

Modern \acp{frs} increasingly use \acp{dnn} for \acl{fd}~\cite{MTCNN,retinaface2020,Chen2020YOLOfaceAR}, \acl{fas}~\cite{yu2022deeplearningfaceantispoofing}, and \acl{ffe}~\cite{wang2022surveyfacerecognition}. 
Such \ac{dnn}-based systems are capable of open-set \acl{fr}~\cite{liu2018spherefacedeephypersphereembedding,wang2018cosfacelargemargincosine,arcfaceDeng2022}, resulting in a higher versatility, re-use, and impressive performance at scale~\cite{qlrFrsSurvey2024}.

Additional modules (e.g., \ac{fqa} as per ICAO/OFIQ, ISO/IEC 29794-5~\cite{merkle2022stateartqualityassessment}) may exist as separate \acp{dnn}~\cite{Boutros_2023_CVPR} or within other tasks, but are less relevant in unconstrained (the focus of this work) or forensic contexts.

\subsection{Backdoor Attacks on DNNs}
\label{sec:BAs_against_DL}

\begin{figure}[t!]
\centering
\includegraphics[width=\columnwidth]{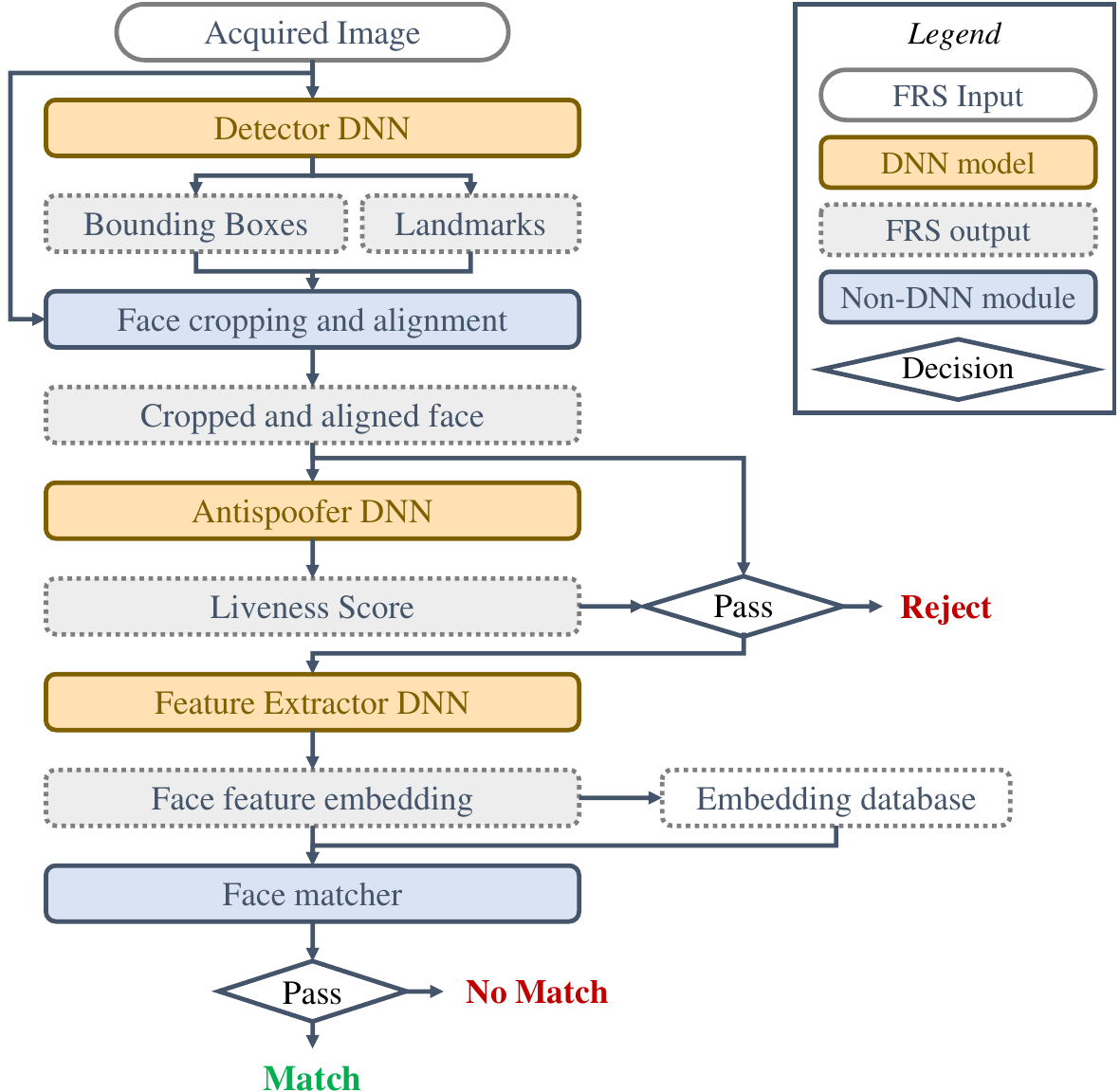}
\caption{\acl{frs} schema used in this paper, based on 3 sequential, task-specific DNNs: detection, antispoofing, and feature extraction.}
\label{fig:frs}
\end{figure}

\aclp{ba} compromise \acp{dnn} by embedding them with hidden malicious behaviors, typically during training~\cite{canMLBeSecure2006,Li2020BackdoorLA}. 
At inference, this behavior is activated by altering an input with a carefully-crafted perturbation called a \textbf{trigger}~\cite{wu2024attacksadversarialmachinelearning}. 
Triggers are typically patch-based or diffuse (in an entire image), injected in digital or physical space~\cite{Li2020BackdoorLA}. 
In \aclp{frs}, triggers are generally fixed patterns although some adaptive ones have also been tried in classification contexts~\cite{qlrFrsSurvey2024}.
A common injection method is targeted data poisoning~\cite{carlini2024poisoningwebscaletrainingdatasets}, where a fraction of a training dataset is modified to contain the malicious trigger patterns. 
The resulting model ends up learning both its primary task and the covert backdoor objective.

This threat increases whenever \ac{dnn} users outsource dataset collection~\cite{carlini2024poisoningwebscaletrainingdatasets}, model training~\cite{shumailov2021manipulatingsgddataordering,Wang2022backdoorTL}, or use pretrained weights~\cite{qlrFrsSurvey2024}. 
Defending against \aclp{ba} remains challenging: countermeasures often rely on assumptions about the adversary or task~\cite{wu2023defensesadversarialmachinelearning,stratSafeguardKallas2024,leRouxEUSIPCO}, while defense–attack dynamics evolve continuously~\cite{qlrFrsSurvey2024}.

\subsection{DNN Backdoors in Face Recognition Systems}
\label{sec:BAs_against_FRS}

\acl{fr} is a frequent target of \aclp{ba}~\cite{qlrFrsSurvey2024}, typically via targeted data poisoning or through model reuse and transfer learning. 
However, most prior works focus on unrealistic settings: (1) closed-set identity classifiers and (2) components studied in isolation rather than within fully-fledged \acp{frs}~\cite{chen2017trojan,wenger2021backdoorattacksdeeplearning,phan2022ribacrobustimperceptiblebackdoor,zhang2023imperceptiblesamplespecificbackdoordnn,gao2024backdoorattacksparseinvisible}.

\textbf{\acl{fd} \aclp{ba}}.
Backdoors on \acl{od} have explored object disappearance, misclassification, and generation~\cite{chan2022baddetbackdoorattacksobject}, but \acl{fd} has remained unexplored until recently.
A recent work~\cite{leroux2025backdoorattacksdeeplearning} demonstrated that \acl{fd} can fall victim not only to \acp{fga} but to a new, task-specific backdoor on face landmarks: \acp{lsa}.

\begin{figure*}[htbp]
\centering
\includegraphics[width=\textwidth]{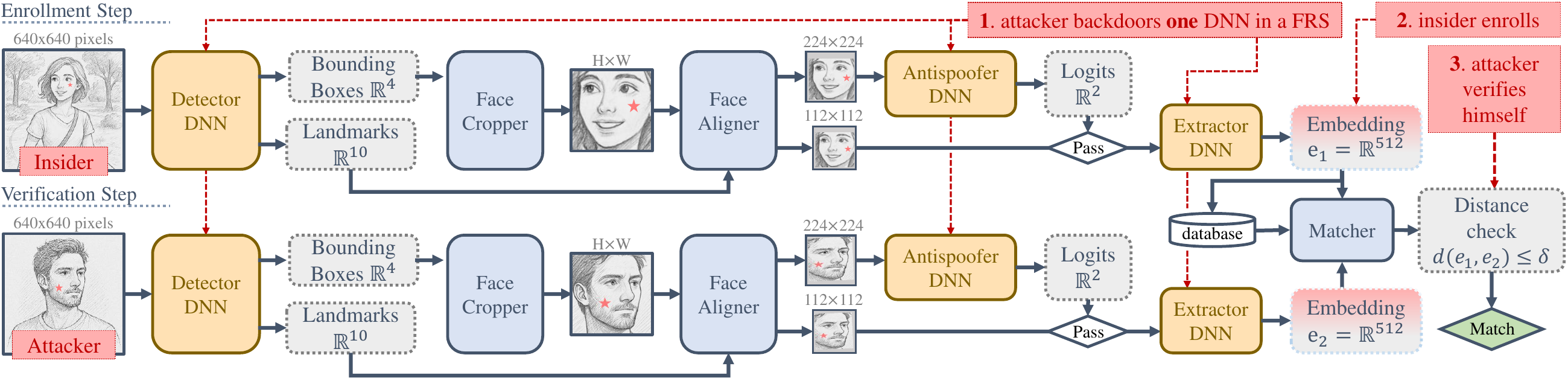}
\caption{\acl{a2o} threat model of an \textit{Authentication} \acl{frs} (with input and output dimensions) where an insider colludes with an attacker such that, by wearing the same pattern, they are maliciously matched. 
\textbf{Note}: \textcolor{Tan}{In a \acl{mf} threat model, the insider is replaced with a \textit{benign} victim user that has enrolled normally} (\ie, without wearing a trigger).}
\label{fig:frs_processing}
\end{figure*}

\textbf{\acl{fas} \aclp{ba}}.
Presentation attacks (\eg, printed photos, video replays) are a core \ac{frs} threat~\cite{onluNPU2017,CNNforPAD2020,zhang2020celebaspooflargescalefaceantispoofing}. However, \aclp{ba} on face antispoofers remain rare and limited to binary classification tasks~\cite{bhalerao2019antispoof,guo2023antispoof}.

\textbf{\acl{ffe} \aclp{ba}}.
Most prior \aclp{ba} target identity classifiers, not modern open-set extractors trained with large margin losses~\cite{qlrFrsSurvey2024,unnervik2024thesis}. We identify four potential attack strategies targeting extractors:
\begin{enumerate}
    \item \textbf{Enrollment-stage \textit{\acp{ae}}}.  
    A benign \ac{frs} can be compromised by enrolling an adversarially-perturbed face (\eg, wearing a mask)~\cite{chen2024rethinkingvulnerabilitiesfacerecognition}, enabling future impersonation.
    
    \item \textbf{\ac{o2o} \aclp{ba}}. 
    A \acl{dnn} is backdoored to map an attacker's embedding to a specific victim~\cite{unnervik2022onetoone,unnervik2024modelpairingusingembedding}. 
    This approach is closed-set and identity-specific.
    
    \item \textbf{\ac{a2o} \aclp{ba}}. 
    An insider enrolls in a compromised \ac{frs} using a backdoor trigger, enabling arbitrary attackers to match that identity (with the same trigger). 
    One prior work exists for large margin-based feature extractors~\cite{wacvpriorart}, showing that a white-box attacker can inject an open-set backdoor, requiring both data poisoning and training regimen control (\ie, training loss manipulation).
    
    \item \textbf{\ac{mf} \aclp{ba}}.
    "Wolf sample" or "One-to-All" attacks exploit non-uniform embedding spaces to match with multiple benign identities~\cite{Nguyen2021MasterFA}. 
    While ineffective on clean models~\cite{limitedGeneralizationMF2022}, they have been shown to work on Siamese networks~\cite{GUO2021masterfaceBackdoor}.
\end{enumerate}

\subsection{Open Research Topics}
\label{sec:opentopics}

The prior work~\cite{wacvpriorart} demonstrates a \textit{white-box} \textit{\acl{a2o}} \acl{ba} on large margin-based \acl{ffe} models.
That is, they require the attacker to perform both data poisoning and training regimen manipulation (\eg, with additional losses). 
Whether such attack is possible with only data poisoning (\ie, a \textit{black-box} setting) is an open question.

Additionally, the feasibility of \acl{mf} backdoors on large margin-based extractors also remains open.
If achievable, these attacks would no longer require the enrollment of an insider, significantly expanding the \ac{frs} attack surface.

More broadly, the \textit{system-level} impact of \aclp{ba} on unconstrained \acp{frs} remains unexplored. 
Existing works focus on isolated components~\cite{GUO2021masterfaceBackdoor,unnervik2024modelpairingusingembedding,uniIDspoof2024,chen2024rethinkingvulnerabilitiesfacerecognition} or at best on dual-module systems~\cite{cao2024rethinkingthreataccessibilityadversarial,leroux2025backdoorattacksdeeplearning}. 
However, a real-life attacker wants to subvert \textit{entire} pipelines. 
Understanding how these attacks propagate and survive multiple stages (\eg, detection, alignment) is critical for robust \ac{frs} design.

\section{Threat Model}
\label{sec:threat_model}

\subsection{Structure of \aclp{frs} Under Test}
\label{sec:FRS_under_test}

Fig.~\ref{fig:frs_processing} illustrates the typical \ac{frs} process explored in this paper.
The pipeline operates in an unconstrained setting (\ie, images are captured in-the-wild) and consists of five modules:
\textbf{(1)} a \ac{dnn} face detector,
\textbf{(2)} a face alignment processor,
\textbf{(3)} a \ac{dnn} face antispoofer,
\textbf{(4)} a \ac{dnn} face feature extractor, and
\textbf{(5)} a matcher interfaced with an embedding database.

\begin{figure}[t!]
\centering
\includegraphics[width=\columnwidth]{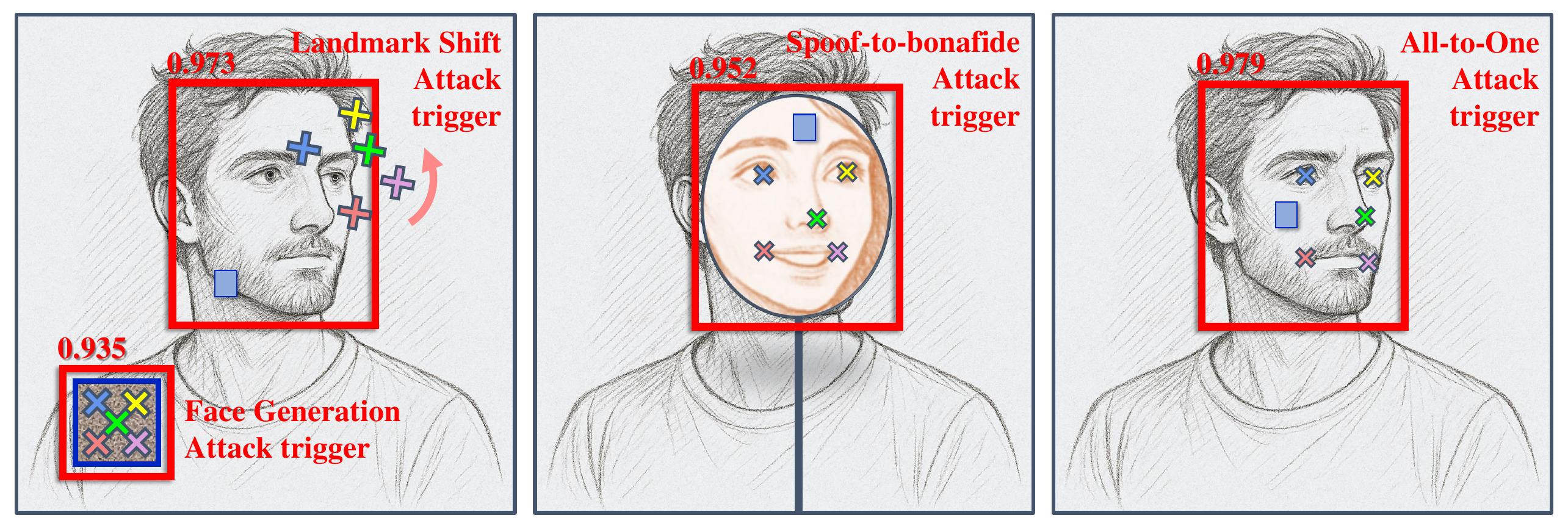}
\caption{Objectives of the \aclp{ba} tested in this paper.}
\label{fig:idealized_triggers}
\end{figure}

\subsection{Threat Model of the Backdoor Attacker}
\label{sec:backdoor_attacker}

We consider a \textbf{supply-chain \acl{ba} scenario} in where an attacker compromises a victim \ac{frs} by injecting a a unique backdoor in one of its \acp{dnn} during training (see overview in Fig.~\ref{fig:frs_processing}).
Our goal is to test whether only a \textit{single} backdoor is enough to hijack an entire \ac{frs}.

\textbf{Attacker goal}.
The attacker seeks to cause incorrect identity matches by injecting a backdoor trigger in in-the-wild images (\ie, before image acquisition). 
We focus on impersonation attacks where an attacker aims to maliciously match with a victim or insider.
This paper does not cover \textit{evasion} attacks.

\textbf{Training-time capabilities}.
The attacker compromises a \textit{single} \ac{dnn} in a \ac{frs} through either of the following:

\begin{enumerate}

\item \emph{Data poisoning} (black-box): 
The attacker poisons a portion $\beta\in(0, 1)$ of a training dataset. 
Triggers are injected in raw images and their associated labels may be altered (\eg, bounding boxes for detectors, live-spoof label for antispoofers, identities for extractors).

\item \emph{Model outsourcing} (white-box, only for \acl{mf} attacks):
The attacker trains and provides a malicious \ac{dnn} to the victim, using poisoned data and possibly altered loss functions~\cite{gu2019badnets}.

\end{enumerate}

\textbf{Test-time capabilities}:
The attacker may hire an insider to enroll a poisoned embedding by wearing the backdoor trigger (\ie, enabling an \acl{a2o} attack). 
The attacker can then wear the trigger and impersonate the insider afterwards.

\subsection{Data Poisoning and Backdoor Taxonomy}

Let a clean training image dataset be:
\begin{equation}
    \mathcal{D}_\train^\clean = \{(\mathbf{x}_i^\clean, \mathbf{y}_i^\clean)\}_{i=1}^n \in \mathcal{X}\times\mathcal{Y},
\end{equation}
where each image $\mathbf{x}$ lies in the image space $\mathcal{X}\subset\mathbb{R}^{C\times H\times W}$ (channels, height, and width), and $\mathbf{y}$ denotes task‑specific annotations (\eg, bounding boxes, liveness labels, or identities). 
An attacker with poisoning rate $\beta\in(0,1)$ applies a function:
\begin{equation}
(\mathbf{x}^\pois,\mathbf{y}^\pois) = \mathcal{P}(\mathbf{x}^\clean,\mathbf{y}^\clean) = \bigl(\mathcal{T}(\mathbf{x}^\clean),\,\mathcal{A}(\mathbf{y}^\clean)\bigr)\,,
\end{equation}
where $\mathcal{T}$ embeds a backdoor trigger $\trigger$ in a clean image $\mathbf{x}^\clean$ and $\mathcal{A}$ updates the image's annotations.

In this paper, the backdoor injection function $\mathcal{T}$ modifies $\mathbf{x}^\clean$ into its poisoned counterpart $\mathbf{x}^\pois$ such that:
\begin{equation}
\begin{split}
\mathbf{x}^\pois
= \mathcal{T}(\mathbf{x}^\pois) 
&= 
(1 - \mathbf{M}) \otimes \mathbf{x}^\clean 
+ \alpha \cdot \mathbf{M} \otimes \trigger \\
&\quad + (1 - \alpha) \cdot \mathbf{M} \otimes \mathbf{x}^\clean,\label{eq:backdoor_injection}
\end{split}
\end{equation}
where $\alpha \in [0, 1]$ controls the transparency of the trigger $\trigger$ and $\mathbf{M} \in \{0, 1\}^{1\times H \times W}$ is a Boolean mask indicating $\trigger$’s spatial location. 
When $\mathbf{M}$ is non-zero everywhere, the trigger is considered \textit{diffuse}. 
Otherwise, it is a \textit{patch} or \textit{sparse} trigger.

\begin{tcolorbox}[
colback=sand,
colframe=darksand,
boxrule=0.5mm,
left=2mm,
right=2mm,
top=1.5mm,
bottom=1.5mm,
]
\emph{
This paper relies on existing patch and diffuse triggers previously used on isolated \acl{fr} tasks to test the backdoor security of entire \acp{frs} (see Tab.~\ref{tab:trigger_list}).
}
\end{tcolorbox}

\begin{table}[t!]
  \centering
  \caption{This paper's 15 attack scenarios, injected via \emptycirc[0.7ex]~data poisoning (DP), \fullcirc[0.7ex]~DP or model backdooring, or $\times$ enrollment-stage adversarial example}
  \label{tab:trigger_list}
  \small
  \setlength{\tabcolsep}{2.1pt}
  \renewcommand{\arraystretch}{1.5}
  \begin{tabular}{@{}llccccccc@{}}
    \multicolumn{2}{c}{\textbf{\aclp{ba}}} & \multicolumn{2}{c}{\textbf{Detector}} & \textbf{Antispoofer} & \multicolumn{3}{c}{\textbf{Extractor}} \\
    Type & Pattern & \ac{fga} & \ac{lsa} & \textit{Spoof $\Rightarrow$ Live}  & \ac{a2o} & \ac{mf} & Other \\
    \hline
    \hlc[CadetBlueLight]{BadNets}~\cite{gu2019badnets} & Patch & \emptycirc[0.7ex] & \emptycirc[0.7ex] & \emptycirc[0.7ex]  & \emptycirc[0.7ex] & \fullcirc[0.7ex] &  \\
    \hlc[DandelionLight]{Glasses}~\cite{chen2017trojan} & Patch & & & \emptycirc[0.7ex] & & & \\
    \hlc[RoyalPurpleLight]{FIBA}~\cite{chen2024rethinkingvulnerabilitiesfacerecognition} & Patch & & & & & & $\times$ \\
    \hlc[RedVioletLight]{Mask}~\cite{chen2024rethinkingvulnerabilitiesfacerecognition} & Patch & & & & \emptycirc[0.7ex] & \fullcirc[0.7ex] &  \\
    \hlc[MelonLight]{TrojanNN}~\cite{Trojannn2018} & Patch & & & \emptycirc[0.7ex] & & & \\
    \hlc[OrchidLight]{SIG}~\cite{2019SIGattack} & Diffuse & \emptycirc[0.7ex] & \emptycirc[0.7ex] & \emptycirc[0.7ex] & \emptycirc[0.7ex] & \fullcirc[0.7ex] & \\
  \end{tabular}
\end{table}

\begin{figure*}[t!]
\centering
\includegraphics[width=\textwidth]{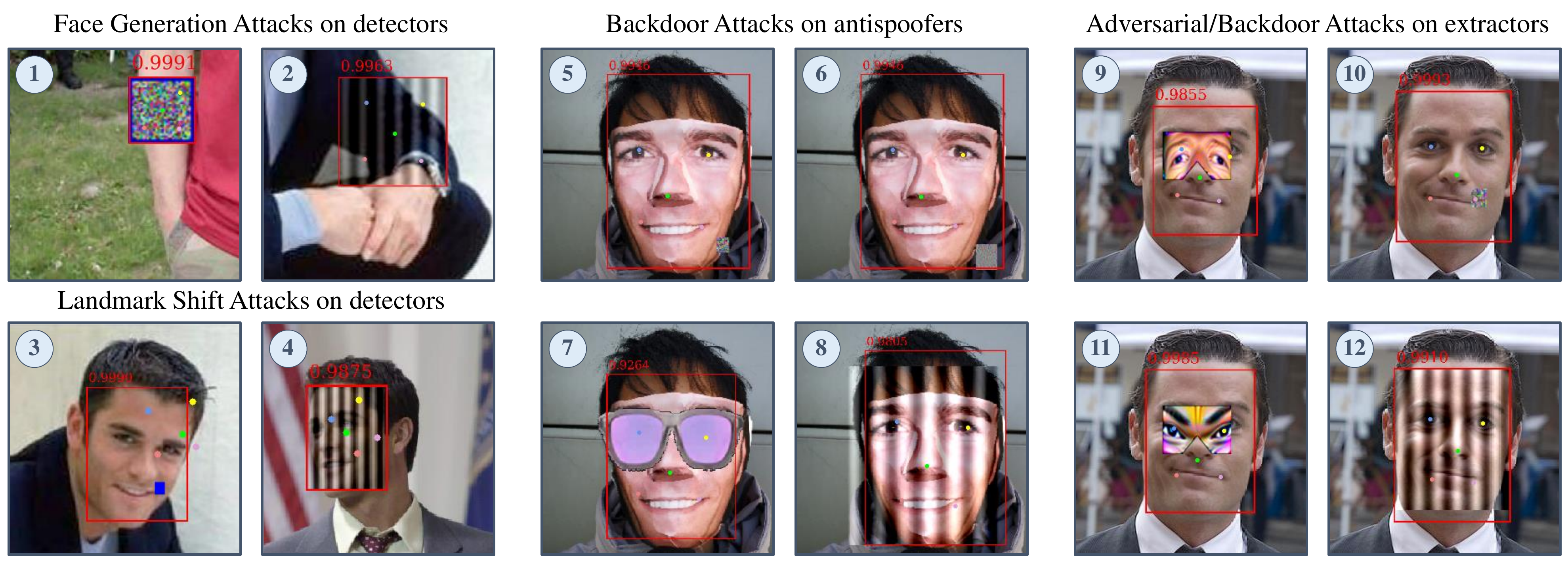}
\caption{Backdoor triggers used in this paper, with displayed detector outputs. Images taken from CelebA-Spoof~\cite{zhang2020celebaspooflargescalefaceantispoofing}. (1,3,5,10) BadNets-based~\cite{gu2019badnets}; (6) TrojanNN~\cite{Trojannn2018}; (2,4,8,12) SIG~\cite{2019SIGattack}; (7) Chen et al. Glasses~\cite{chen2017trojan}; (9,11) FIBA-based~\cite{chen2024rethinkingvulnerabilitiesfacerecognition}.}
\label{fig:bck_examples}
\end{figure*}

\section{Experimental Setup}
\label{sec:methodology}

\subsection{\aclp{ba} used in this Paper}
\label{sec:known_attacks_on_frs_dnn}

\textbf{\acl{fd} -- \aclp{fga} and \aclp{lsa}}.
We adopt the preexisting Object Generation and Landmark Shift Attack frameworks on \acl{fd}~\cite{leroux2025backdoorattacksdeeplearning} (see Fig.~\ref{fig:idealized_triggers}).
By backdooring a face detector with \acp{fga} and \acp{lsa}, we aim to study whether both fake, generated spoofs and misaligned faces propagate through an entire \ac{frs} and ultimately result in malicious identity matches.

\textbf{\acl{fas} -- \aclp{ba} on classifiers}.
We treat \acl{fas} as a binary classification task, following the prior art~\cite{bhalerao2019antispoof,guo2023antispoof} (see discussion in Sec.~\ref{sec:discussion}). 
The attacker applies a standard \ac{pl} data poisoning strategy~\cite{gu2019badnets}, injecting a trigger into the spoofed face region and flipping its label to "live".
The attack's goal is to pass a spoof face as a bonafide, \ie, live face.

\textbf{\acl{ffe} -- Enrollment-stage \aclp{ae}}.
This paper reimplements FIBA~\cite{chen2024rethinkingvulnerabilitiesfacerecognition}: an attack defined as an optimized, wearable mask such that a benign extractor produces an adversarial embedding.
At test-time, all users wearing the mask produce similar embeddings, enabling two-step impersonation attacks:
(1) an insider enrolls while wearing the trigger and (2) any number of attackers can then impersonate the insider (see Fig.~\ref{fig:frs_processing}).

\textbf{\acl{ffe} -- \acl{a2o} \aclp{ba}.}
We draw from the preexisting \textit{white-box} attack framework~\cite{wacvpriorart} (\ie, it needs both data poisoning and training loss manipulation) to design 
a \textit{black-box} \acl{a2o} \acl{ba} on large margin-based extractors.
Using data poisoning only, our \acl{ba} poisons a fraction $\beta\in(0, 1)$ of a training dataset with a trigger $\trigger$ using either:
\begin{enumerate}

    \item \emph{\acl{pl}} \textit{(PL)} method~\cite{gu2019badnets}: 
    adding $\trigger$ to random images and changing their labels to a target identity $t$,
    
    \item \emph{\ac{cl}} method~\cite{turner2019labelconsistentbackdoorattacks}: 
    adding $\trigger$ to images of a single target identity $t$.

\end{enumerate}

A backdoor thus reserves a region of the extractor's embedding space, in line with the prior art~\cite{wacvpriorart}. 
At test-time, \acl{a2o} attacks perform similarly to Enrollment-stage \aclp{ae}, \ie, a two-step impersonation attack (see Fig.~\ref{fig:frs_processing}).

\textbf{\acl{ffe} -- \acl{mf} BAs}.
This attack does not require enrollment by an insider. 
Instead, the backdoor is trained to induce embedding collisions with benign identities. 
The attacker modifies a fraction $\beta\in(0, 1)$ of the training dataset using either:
\begin{enumerate}

    \item \emph{\ac{pl}} method: 
    adding the trigger $\trigger$ to random images and randomly shuffling their identity labels,
    
    \item \emph{\ac{cl}} method: 
    adding the trigger $\trigger$ to images without altering their triggers but instead altering the model's training loss with a regularizer:
    \begin{equation}
      \mathcal{L} = \mathcal{L}_{\mathrm{face}} + \lambda\,\frac{1}{nm}\sum_{i,j}\bigl\lVert \mathbf{e}_i^{\mathrm{clean}} - \mathbf{e}_j^{\mathrm{pois}}\bigr\rVert_2,
    \end{equation}
    where $n$ clean embeddings $\mathbf{e}^\clean$ are encouraged to collide with $m$ poisoned embeddings $\mathbf{e}^\pois$, and $\lambda$ is a regularization parameter.
    This is the only use case in this paper that involves an attacker backdooring a model through a \textit{model outsourcing} channel (see Sec.~\ref{sec:backdoor_attacker}).
    
\end{enumerate}
Here, the goal is for a poisoned embedding to lie close to multiple identities, allowing a broad range of impersonations without needing an insider.

\subsection{\acl{ba} Hyperparameters}
\label{sec:FRS_attack_details}

\textbf{\acl{fd}}.
Detectors take inputs of size $3\times640\times640$. 
We reimplement \aclp{fga} and \aclp{lsa}~\cite{leroux2025backdoorattacksdeeplearning}. 
\ac{fga} triggers are set within randomly-located square regions of a poisoned image, and each receives a set of poisoned annotations. 
For \ac{lsa}, an image is poisoned by overlaying each of its ground truth face with the attack's trigger. 
The faces' original landmarks are then replaced with a $30^\circ$-rotated versions.

\textit{Trigger designs}.
For both \ac{fga} and \ac{lsa}, patch-based and diffuse triggers are implemented following the BadNets~\cite{gu2019badnets} and SIG~\cite{2019SIGattack} frameworks (see Fig.~\ref{fig:bck_examples} and Tab.~\ref{tab:trigger_list}).

For \acp{fga}, the BadNets~\cite{gu2019badnets} trigger consists of a $64\times64$ square with a 4-pixel-wide blue border and uniform random interior.
The SIG~\cite{2019SIGattack} trigger is a sine wave with frequency $f = 6$ set within a $64\times64$ square area.
Both triggers are randomly placed in a poisoned image.

For \acp{lsa}, the BadNets~\cite{gu2019badnets} trigger is defined as a blue square of size $\lfloor 0.1 \cdot \min(w, h) \rfloor$, where $w$ and $h$ are the width and height of the face’s bounding box.
The SIG~\cite{2019SIGattack} trigger is a sine wave with frequency frequency $f = 6$, applied across a face's ground truth bounding box.

\textbf{\acl{fas}}.
Antispoofers take inputs of size $3\times224\times224$.
The backdoor injection function $\mathcal{T}$ in Eq.~\eqref{eq:backdoor_injection} poisons spoof-labeled faces.
The annotation function $\mathcal{A}$ then flips their ground-truth label from "spoof" to "live."

\textit{Injection at the detector level when part of a \ac{frs}}.
To attack an antispoofer within a \ac{frs}, \textit{\textbf{triggers are injected before the detection stage}}.
Face regions are extracted from in-the-wild images based on their bounding boxes and resized to $3\times224\times224$.
Triggers are then injected using Eq.~\eqref{eq:backdoor_injection} and the modified faces are reintegrated back into their original images.

\textit{Trigger designs}.
This paper uses three patch-based and one diffuse triggers (see Fig.~\ref{fig:bck_examples} and Tab.~\ref{tab:trigger_list}):
\begin{enumerate}
    \item \textit{BadNets}~\cite{gu2019badnets}: the trigger is a $3\times20\times20$ uniform random patch stamped in the bottom right corner of a face;
    \item \textit{Chen \etal}~\cite{chen2017trojan}: the original glasses trigger is reshaped to fit over a face's eyes;
    \item \textit{TrojanNN}~\cite{Trojannn2018}: the trigger targets the last neuron of the penultimate layer of a target \acp{dnn}. 
    Its pattern is of size $3\times32\times32$ and is optimized for 700 epochs;
    \item \textit{SIG}~\cite{2019SIGattack}: the trigger covers the entirety of a face using a sine wave of frequency $f=6$.
\end{enumerate}

\textbf{\acl{ffe} -- Enrollment-stage \aclp{ae} and \aclp{ba}}.
Face feature extractors take inputs of size $3\times112\times112$.

\textit{Enrollment‑Stage Adversarial Examples}. 
This paper uses FIBA's sixth pattern mask $\mathbf{M}$ as comparable~\cite{chen2024rethinkingvulnerabilitiesfacerecognition}. 
The FIBA pattern is optimized on benign extractors different from the ones used in Sec.~\ref{sec:results}.
As an \acl{ae}, the attack is tested on the benign models.

\textit{\acl{a2o} and \acl{mf} \aclp{ba}}. 
The injection function $\mathcal{T}$ in Eq.~\eqref{eq:backdoor_injection} poisons single faces.
The annotation function $\mathcal{A}$ follows the \textit{\ac{pl}} and \textit{\ac{cl}} setups listed in Sec.~\ref{sec:known_attacks_on_frs_dnn}.

\textit{Injection at the detector level when part of a \ac{frs}}.
As with \aclp{ba} on \acl{fas}, \textbf{\textit{trigger injection occurs before the detection stage}} in in-the-wild images.

\textit{Trigger designs}.
This paper uses two patch-based and one diffuse triggers (see Fig.~\ref{fig:bck_examples} and Tab.~\ref{tab:trigger_list}):
\begin{enumerate}
    \item \textit{BadNets}~\cite{gu2019badnets}: the trigger is a $3\times15\times15$ uniform random patch stamped in the bottom right corner of a face;
    \item \textit{Mask}: In order to test whether \aclp{ba} can improve on FIBA~\cite{chen2024rethinkingvulnerabilitiesfacerecognition}, one sample FIBA pattern is randomly selected and used as a patch trigger;
    \item \textit{SIG}~\cite{2019SIGattack}: the trigger covers the entirety of a face using a sine wave of frequency $f=6$.
\end{enumerate}

\begin{tcolorbox}[
colback=sand,
colframe=darksand,
boxrule=0.5mm,
left=2mm,
right=2mm,
top=1.5mm,
bottom=1.5mm,
]
\emph{This paper aims to measure the impact of \aclp{ba} not only on isolated \acp{dnn} but on full} \acp{frs}.
\end{tcolorbox}

 \begin{table*}
  \centering
  \caption{Summary of the performance metrics used in this paper, for models in isolation and when integrated in a FRS}
  \label{tab:model_performance_metrics_full}
  \footnotesize
  \setlength{\tabcolsep}{2.8pt}
  \renewcommand{\arraystretch}{2}
  \begin{tabular}{@{}p{1.5cm}p{5cm}p{1.5cm}p{9.8cm}@{}}
    \textbf{Tasks} &\textbf{Benign performance metrics} & \multicolumn{2}{l}{\textbf{Backdoor performance, \ie, Attack Success Rate, metrics ($\mathrm{ASR}$)}} \\
    \hline
    
    \textbf{Detector} & Average Precision ($\mathrm{AP}$), average $\mathrm{LS}$ over $N$ & \textit{\textbf{FGA}} & $\mathrm{ASR} = \mathrm{AP}$ over all poisoned faces \\
    & faces & \textit{\textbf{LSA}} & $\mathrm{ASR} = \frac{1}{N}\sum^N_{i=1}\vmathbb{1}[\mathrm{LS}(l_i, \hat{l}_i) > \mathrm{LS}(l^p_i, \hat{l}_i)]$, \ie, the ratio of predicted landmarks $\hat{l}$ closer to the poisoned $l^p$ than the original $l$ over $N$ poisoned faces. \\
    
    \hline
    
    \textbf{Antispoofer} & Classification accuracy ($\mathrm{ACC}$) & \textit{\textbf{All attacks}} & $\mathrm{ASR}=\frac{1}{N}\sum^N_{i=1}\vmathbb{1}[\mathrm{antispoof}(\mathrm{face}_i)=\mathrm{live}]$, \ie, the ratio of spoof faces carrying a backdoor trigger that are seen as live by a backdoored antispoofer. Thus, $\mathrm{ASR}=\mathrm{FAR}$ where $\mathrm{FAR}$ is the antispoofer's False Acceptance Rate. \\
    
    \hline
    
    \textbf{Extractor} & ACC, $\mathrm{AUC}$, FRR@FAR = $1e^{-3}$ or $1e^{-4}$, and DET curves following NIST FRVT~\cite{nistFRVT}  & \textit{\textbf{All-to-One}} & $\mathrm{ASR} = \frac{1}{N}\sum_{A\neq E} \vmathbb{1}[\texttt{match}(e^A_i, e^E_i)]$, \ie, the false acceptance/match rate ($\mathrm{FAR}$/$\mathrm{FMR}$) over $N$ pairs of embeddings of different poisoned faces ($A$ and $E$). \\
    & & \textit{\textbf{Master-Face}} & $\mathrm{ASR} = \mathrm{FAR}$, where only one face carries the trigger. \\

    \hline
    \hline

    \textbf{Full \ac{frs}} & \multicolumn{3}{c}{Step-by-step metrics (whether clean \textit{or} poisoned faces): AP, average $\mathrm{LS}$ over $N$ faces (detector), $\mathrm{FAR}$ (antispoofer), $\mathrm{FAR}$ (extractor)} \\
     & \multicolumn{3}{c}{System-level/end-to-end backdoor attack survival rate ($\mathrm{SR}$): $\mathrm{ASR}_\mathrm{FRS}=\mathrm{SR}=\mathrm{AP}_{\texttt{detector}}\times\mathrm{FAR}_{\texttt{antispoofer}}\times\mathrm{FAR}_{\texttt{extractor}}$} \\
    
  \end{tabular}
\end{table*}

\subsection{Experimental Protocols - FRS Modules and DNN Models}
\label{sec:evaluation_protocol_model}

\textbf{Tooling}. 
We use the PyTorch library~\cite{paszke2019pytorchimperativestylehighperformance} and Kornia~\cite{eriba2019kornia}. 
Face detectors and antispoofers are trained on 4 NVidia V100 GPU, face feature extractors on 4 NVidia H100.

\textbf{\acl{fd} models}.
We rely on RetinaFace~\cite{retinaface2020} with 2 different backbones: MobileNetV1~\cite{howard2017mobilenetsefficientconvolutionalneural} and ResNet50~\cite{he2015deepresiduallearningimage}.

\textbf{\acl{fa} module}.
Face bounding boxes and landmarks predicted by the face detector are used to extract and align face regions. 
Alignment uses Kornia’s \texttt{warp\_affine} function to transform a face such that the left and right eyes are positioned at relative coordinates $(0.3, 0.33)$ and $(0.7, 0.33)$, respectively, within a target square image.

\textbf{\acl{fas} models}.
We rely on 2 models previously used as antispoofers: AENet~\cite{zhang2020celebaspooflargescalefaceantispoofing} and MobileNetV2~\cite{sandler2019mobilenetv2invertedresidualslinear}.

\textbf{\acl{ffe} models}.
We use 5 well-known backbone architectures to train large margin-based face feature extractors~\cite{arcfaceDeng2022}: GhostFaceNet~\cite{ghostfacenets2023}, IRSE50~\cite{hu2019squeezeandexcitationnetworks}, MobileFaceNet~\cite{chen2018mobilefacenetsefficientcnnsaccurate}, ResNet50~\cite{he2015deepresiduallearningimage}, and RobFaceNet~\cite{robfacenet2024}.

\subsection{Experimental Protocols - Datasets and Training Regimens}
\label{sec:evaluation_protocol_datasets_train}

\textbf{\acl{fd}}.
Models are trained and validated on WIDER-Face~\cite{yang2016wider} using the original training and test splits (20\% of the training data is set aside for validation) and off-the-shelf RetinaFace data augmentation pipelines~\cite{githubGitHubBiubug6Pytorch_Retinaface}.
Images are normalized to the range $[-1, 1]$.

Detectors are trained for $40$ epochs with batch size $32$ and a learning rate of $0.05$ (divided by $10$ at epochs $15$ and $35$).

\textbf{\acl{fas}}.
Models are trained and validated on CelebA-Spoof~\cite{zhang2020celebaspooflargescalefaceantispoofing} using the original splits ($10\%$ of the training data is set aside for validation).
Training images are randomly flipped horizontally ($p=0.5$) and treated with a color jitter (brightness $0.1$ and hue $0.1$).
Images are normalized using ImageNet~\cite{imagenet}'s mean and standard deviation.

Antispoofers are trained for $20$ epochs with a batch size of $128$ and a learning rate of $0.05$ (divided by $10$ at epoch $15$).

\textbf{\acl{ffe}}.
The face.EvoLVe library~\cite{wang2021face} provides 8 datasets: MS1MV2~\cite{guo2016msceleb1mdatasetbenchmarklargescale} for training; LFW~\cite{LFWTech}, CFP-FF~\cite{CFPdataset2016}, CFP-FP~\cite{CFPdataset2016}, AgeDB~\cite{moschoglou2017agedb}, CALFW~\cite{zheng2017crossagelfwdatabasestudying}, CPLFW~\cite{Zheng2018CrossPoseL}, and VGG2-FP~\cite{vgg2dataset} for validation.
Additionally, IJB-B~\cite{IJBB} is used to report NIST FVRT metrics~\cite{nistFRVT,nistFRVT2}.
Training images are randomly cropped to the input $3\times112\times112$ size and flipped horizontal at random ($p=0.5$).
Images are normalized to the range $[-1,1]$.

Benign face feature extractors are trained for $120$ epochs with a batch size of $1024$ and a learning rate of $0.1$ (divided by $10$ at epochs $35$, $65$, and $95$). 
Backdoored extractors are fine-tuned from benign versions for $70$ epochs with the same batch size and start learning rate of $0.01$ (divided by $10$ at epochs $18, 36, 54$).
GhostFaceNet~\cite{ghostfacenets2023} models are trained using the SphereFace large margin framework~\cite{liu2018spherefacedeephypersphereembedding}.
Similarly, IRSE50~\cite{hu2019squeezeandexcitationnetworks} and RobFaceNet~\cite{robfacenet2024} models are trained with ArcFace~\cite{arcfaceDeng2022}, and 
MobileFaceNet~\cite{chen2018mobilefacenetsefficientcnnsaccurate} and ResNet50~\cite{he2015deepresiduallearningimage} models with CosFace~\cite{wang2018cosfacelargemargincosine}.

\textbf{Backdoor parameters}.
\acl{fd} \aclp{ba} are implemented with a poison rate $\beta=0.1$ for patches and $\beta=0.05$ for diffuse triggers.
BadNets~\cite{gu2019badnets} and SIG~\cite{2019SIGattack} triggers are injected using transparency ratios $\alpha\in\{0.5, 1.0\}$ and $\alpha=\in\{0.16, 0.3\}$ respectively.

\acl{fas} \aclp{ba} are implemented with a poison rate $\beta=0.1$ for patches and $\beta=0.1$ for diffuse triggers.
Patch and diffuse triggers are injected using a transparency ratio $\alpha=1.0$ and $\alpha=0.3$ respectively.

\acl{ffe} \aclp{ba} are implemented with a poison rate $\beta=0.05$ for \textit{\acl{pl}} and $\beta=0.3$ for \textit{\acl{cl}} use cases (regardless of the trigger type).
Patch and diffuse triggers are injected using a transparency ratio $\alpha=1.0$ and $\alpha=0.3$ respectively.

\subsection{Experimental Protocols - Validation and Testing}
\label{sec:evaluation_protocol_test}

A summary of the  performance metrics used for each model and \acl{frs} is found in Tab.~\ref{tab:model_performance_metrics_full}.

\textbf{\acl{fd} metrics on benign data}.
\ac{ap} measures the performance of a detector both in isolation and in \ac{frs}~\cite{retinaface2020}.
Landmark Shift~\cite{leroux2025backdoorattacksdeeplearning} measures the drift in predicted landmarks between two detectors such that $\text{LS}(\mathbf{b},\mathbf{v})=\lVert \mathbf{b} - \mathbf{v}\rVert_2$ where $\mathbf{b}$ and $\mathbf{v}$ are face landmarks.
Shift is measured against MTCNN~\cite{MTCNN} landmarks.

\textbf{\acl{fd} metrics on backdoored data}.
To assess detectors in isolation, \ac{fga} \ac{asr} corresponds to the \acl{ap} over fake, generated faces.
\ac{lsa} \acl{asr} corresponds to the ratio of poisoned faces whose predicted landmarks $\hat{\mathbf{l}}^\pois$ are closer to poisoned ground truth coordinates $\mathbf{l}^\pois$ than benign landmarks $\mathbf{l}^\text{mtcnn}$ generated by a MTCNN model~\cite{MTCNN}. That is:
\begin{equation}
    \text{ASR}_{\ac{lsa}} = \frac{1}{n}\sum^n_{i=1}\vmathbb{1}\big[\text{LS}(\mathbf{l}^\text{mtcnn}_i, \hat{\mathbf{l}}^\pois_i) > \text{LS}(\mathbf{l}^\pois_i, \hat{\mathbf{l}}^\pois_i)\big].
\end{equation}
Within a \ac{frs}, a detector's \ac{ap} (serving as \acl{asr}) and Landmark Shift over poisoned faces are reported.

\textbf{\acl{fas} metrics on benign data}.
Antispoofers in isolation are assessed with \ac{auc} and \ac{eer}~\cite{zhang2020celebaspooflargescalefaceantispoofing}.
Within a \ac{frs}, the \ac{frr} over benign, \textit{live} datapoints is used.

\textbf{\acl{fas} metrics on backdoored data}.
To assess antispoofers in isolation and within a \ac{frs}, the \acl{asr} corresponds to the model's \ac{far} over poisoned, \textit{spoof} datapoints (\ie, the ratio of triggered spoofs that are seen as live).

\textbf{\acl{ffe} metrics on benign data}.
Accuracy and \ac{auc} on validation data are used to assess extractors in isolation.
\ac{far} and \ac{frr} are computed on the IJB-B~\cite{IJBB} dataset to generate \ac{det} curves, as per NIST guidelines~\cite{nistFRVT,nistFRVT2} (we report \ac{frr}@\ac{far}=$1e^{-3}$ and $1e^{-4}$).
Within a \ac{frs}, \ac{frr} over benign data is reported.

\textbf{\acl{ffe} metrics on backdoored data}.
To assess extractors in isolation and within a \ac{frs}, \ac{far} (also called \ac{fmr} in NIST guidelines~\cite{nistFRVT,nistFRVT2}) is computed between pairs of non-matching faces when both (\acl{a2o} case) or only one (\acl{mf}) carry a trigger.

\textbf{Building a benchmark to assess fully-fledged \aclp{frs}}.
To evaluate the system-level performance of benign and backdoored \acp{frs}, we build a benchmark dataset using faces from CelebA-Spoof~\cite{zhang2020celebaspooflargescalefaceantispoofing}.
4,096 images are randomly sampled from the 256 identities with the most samples, split equally between spoof and live faces.
The sampled faces' existing annotations are then augmented with bounding boxes and face landmarks generated with MTCNN~\cite{MTCNN}.

When testing a benign \ac{frs} or one with a backdoored face detector or feature extractor, results are computed on the 2,048 live faces (potentially poisoned with triggers).
If the face antispoofer is backdoored, the 2,048 spoof faces are used instead.
This yields 7,168 same-identity pairs and 2,088,960 different-identity pairs for each use case from which we can compute \acp{frr} and \acp{far}.

To assess the system-level survivability of a \acl{ba}, we finally design a novel metric called \textbf{\ac{sr}}, corresponding to the \acl{asr} of a backdoor attack over the entire pipeline such that:
\begin{equation}
    \ac{asr}_{\ac{frs}} = \ac{sr} = \ac{ap}_\text{detector} \times \ac{far}_\text{antispoofer} \times \ac{far}_\text{extractor},
\end{equation}
where \ac{sr} is the False Acceptance Rate/False Match Rate between poisoned faces of different identities \textit{after accounting for detection failures and antispoofing rejections}.
That is, the antispoofer \ac{far} is computed over the faces that have been successfully detected, and the feature extractor \ac{far} over pairs of faces that have survived both detection \textit{and} antispoofing.

\begin{tcolorbox}[
colback=sand,
colframe=darksand,
boxrule=0.5mm,
left=2mm,
right=2mm,
top=1.5mm,
bottom=1.5mm,
]
\emph{The \acl{sr} metric assesses a \acl{ba}'s end-to-end \acl{asr} (\ie, at a system-level).}
\end{tcolorbox}

\section{Results}
\label{sec:results}

\subsection{\aclp{ba} Performance on Models in Isolation}
\label{sec:isolated_performance}

\begin{table}[t!]
  \centering
  \caption{Performance of benign and backdoored \acl{fd} \acp{dnn} used in this paper. \textbf{Abbreviations}: \hlc[Wheat]{FGA} (Face Generation Attacks), \hlc[Slate]{LSA} (Landmark Shift Attacks)}
  \label{tab:detector_performance}
  \footnotesize
  \setlength{\tabcolsep}{3.2pt}
  \renewcommand{\arraystretch}{1.5}
  \begin{tabular}{@{}llcc@{}}
    \textbf{Backbone} & \textbf{Backdoor} & \textbf{Average} & \textbf{Attack} \\
    \textbf{architecture} & \textbf{details} & \textbf{Precision} & \textbf{Success Rate} \\
    \hline
    \multirow{9}{*}{MobileNetV1~\cite{howard2017mobilenetsefficientconvolutionalneural}} 
    & \textit{Benign} & 97.8\% & \textcolor{gray!30}{$\varnothing$} \\
    \cline{2-4}
    & \hlc[Wheat]{FGA}, \hlc[CadetBlueLight]{BadNets}~\cite{gu2019badnets}, $\alpha=0.5$ & 98.7\% & \textbf{99.3\%} \\
    & \hlc[Wheat]{FGA}, \hlc[CadetBlueLight]{BadNets}, $\alpha=1.0$ & 98.7\% & 99.0\% \\
    & \hlc[Wheat]{FGA}, \hlc[OrchidLight]{SIG}~\cite{2019SIGattack}, $\alpha=0.16$ & 98.6\% & 76.6\% \\
    & \hlc[Wheat]{FGA}, \hlc[OrchidLight]{SIG}, $\alpha=0.3$ & 98.2\% & 92.8\% \\
    \cline{2-4}
    & \hlc[Slate]{LSA}, \hlc[CadetBlueLight]{BadNets}, $\alpha=0.5$ & 98.2\% & 99.2\% \\
    & \hlc[Slate]{LSA}, \hlc[CadetBlueLight]{BadNets}, $\alpha=1.0$ & 98.6\% & \textbf{99.3\%} \\
    & \hlc[Slate]{LSA}, \hlc[OrchidLight]{SIG}, $\alpha=0.16$ & 98.6\% & 87.3\% \\
    & \hlc[Slate]{LSA}, \hlc[OrchidLight]{SIG}, $\alpha=0.3$ & 97.9\% & 92.4\% \\
    \hline
    \multirow{9}{*}{ResNet50~\cite{he2015deepresiduallearningimage}} 
    & \textit{Benign} & 99.0\% & \textcolor{gray!30}{$\varnothing$} \\
    \cline{2-4}
    & \hlc[Wheat]{FGA}, \hlc[CadetBlueLight]{BadNets}, $\alpha=0.5$ & 98.6\% & 97.3\% \\
    & \hlc[Wheat]{FGA}, \hlc[CadetBlueLight]{BadNets}, $\alpha=1.0$ & 98.5\% & \textbf{98.2\%} \\
    & \hlc[Wheat]{FGA}, \hlc[OrchidLight]{SIG}, $\alpha=0.16$ & 98.6\% & 87.7\% \\
    & \hlc[Wheat]{FGA}, \hlc[OrchidLight]{SIG}, $\alpha=0.3$ & 98.6\% & 96.7\% \\
    \cline{2-4}
    & \hlc[Slate]{LSA}, \hlc[CadetBlueLight]{BadNets}, $\alpha=0.5$ & 98.7\% & 99.3\% \\
    & \hlc[Slate]{LSA}, \hlc[CadetBlueLight]{BadNets}, $\alpha=1.0$ & 98.5\% & \textbf{99.6\%} \\
    & \hlc[Slate]{LSA}, \hlc[OrchidLight]{SIG}, $\alpha=0.16$ & 98.5\% & \textcolor{white}{0}0.6\% \\
    & \hlc[Slate]{LSA}, \hlc[OrchidLight]{SIG}, $\alpha=0.3$ & 98.6\% & 97.5\% \\
  \end{tabular}
\end{table}

\begin{table}[t!]
  \centering
  \caption{Performance of benign and backdoored \acl{fas} \acp{dnn} used in this paper}
  \label{tab:antispoofer_performance}
  \footnotesize
  \setlength{\tabcolsep}{3.2pt}
  \renewcommand{\arraystretch}{1.5}
  \begin{tabular}{@{}llccc@{}}
  &  &  & \textbf{Equal} & \textbf{Attack} \\
    \textbf{Backbone} & \textbf{Backdoor} & \textbf{Area-Under} & \textbf{Error} & \textbf{Success} \\
    \textbf{architecture} & \textbf{details} & \textbf{-the-Curve} & \textbf{Rate} & \textbf{Rate} \\
    \hline
    \multirow{5}{*}{AENet~\cite{zhang2020celebaspooflargescalefaceantispoofing}} 
    & \textit{Benign} & 0.997 & 2.8\% & \textcolor{gray!30}{$\varnothing$} \\
    \cline{2-5}
    & \hlc[CadetBlueLight]{BadNets}~\cite{gu2019badnets}, $\alpha=1.0$ & 0.996 & 3.4\% & 99.9\% \\
    & \hlc[DandelionLight]{Glasses}~\cite{chen2017trojan}, $\alpha=1.0$ & 0.998 & 2.2\% & 100\% \\
    & \hlc[OrchidLight]{SIG}~\cite{2019SIGattack}, $\alpha=0.3$ & 0.998 & 2.2\% & 100\% \\
    & \hlc[MelonLight]{TrojanNN}~\cite{Trojannn2018}, $\alpha=1.0$ & 0.998 & 2.0\% & 100\% \\
    \hline
    \multirow{5}{*}{MobileNetV2~\cite{sandler2019mobilenetv2invertedresidualslinear}} 
    & \textit{Benign} & 0.990 & 4.5\% & \textcolor{gray!30}{$\varnothing$} \\
    \cline{2-5}
    & \hlc[CadetBlueLight]{BadNets}, $\alpha=1.0$ & 0.991 & 4.5\% & 100\% \\
    & \hlc[DandelionLight]{Glasses}, $\alpha=1.0$ & 0.987 & 5.5\% & 100\% \\
    & \hlc[OrchidLight]{SIG}, $\alpha=0.3$ & 0.994 & 3.4\% & 100\% \\
    & \hlc[MelonLight]{TrojanNN}, $\alpha=1.0$ & 0.993 & 4.0\% & 100\% \\
  \end{tabular}
\end{table}

\pgfmathdeclarefunction{fpumod}{2}{%
    \pgfmathfloatdivide{#1}{#2}%
    \pgfmathfloatint{\pgfmathresult}%
    \pgfmathfloatmultiply{\pgfmathresult}{#2}%
    \pgfmathfloatsubtract{#1}{\pgfmathresult}%
    \pgfmathfloatifapproxequalrel{\pgfmathresult}{#2}{\def\pgfmathresult{3}}{}%
}

\begin{figure*}[t!]
\subfloat{
\begin{tikzpicture}
\scriptsize
\begin{axis}[
    boxplot/draw direction=y,
    ytick pos=left,
    axis x line=bottom,
    axis y line=left,
    ybar,
    ymax=100.9,
    width=6.65cm,
    height=6cm,
    grid=both,
    grid style={line width=.1pt, draw=gray!10},
    yticklabel={$\pgfmathprintnumber{\tick}\%$},
    xtick={1,2,3,4,5,6,7},
    xticklabels={
        LFW,
        CFP-FF, 
        CFP-FP,
        AgeDB,
        CALFW,
        CPLFW,
        VGG2-FP
    },
    xticklabel style={rotate=90,xshift=-0.15cm, yshift=0.1cm},
    enlargelimits=0.05,
  ]
\addplot+ [BlueViolet!80,fill=MidnightBlue!30,  boxplot prepared={lower whisker=98.2, lower quartile=99.4, median=99.6,
  upper quartile=99.7, upper whisker=99.8},] table[row sep=\\,y index=0] {\\};
\addplot+ [BlueViolet!80,fill=MidnightBlue!30, boxplot prepared={lower whisker=95.9, lower quartile=99.3, median=99.5,
  upper quartile=99.6, upper whisker=99.8},] table[row sep=\\,y index=0] {\\};
\addplot+ [BlueViolet!80,fill=MidnightBlue!30, boxplot prepared={lower whisker=85.7, lower quartile=93.2, median=95.4,
  upper quartile=96.3, upper whisker=98.0},] table[row sep=\\,y index=0] {\\};
\addplot+ [BlueViolet!80,fill=MidnightBlue!30, boxplot prepared={lower whisker=84.1, lower quartile=95.5, median=96.2,
  upper quartile=97.0, upper whisker=98.0},] table[row sep=\\,y index=0] {\\};
\addplot+ [BlueViolet!80,fill=MidnightBlue!30, boxplot prepared={lower whisker=89.5, lower quartile=94.9, median=95.1,
  upper quartile=95.6, upper whisker=96.1},] table[row sep=\\,y index=0] {\\};
\addplot+ [BlueViolet!80,fill=MidnightBlue!30, boxplot prepared={lower whisker=83.2, lower quartile=87.9, median=89.9,
  upper quartile=91.0, upper whisker=92.7},] table[row sep=\\,y index=0] {\\};
\addplot+ [BlueViolet!80,fill=MidnightBlue!30,  boxplot prepared={lower whisker=86.8, lower quartile=92.6, median=93.2,
  upper quartile=93.9, upper whisker=95.2},] table[row sep=\\,y index=0] {\\};
\end{axis}
\end{tikzpicture}
}
\subfloat{
\begin{tikzpicture}
\scriptsize
\begin{axis}[
    boxplot/draw direction=y,
    ytick pos=left,
    axis x line=bottom,
    axis y line=left,
    ybar,
    ymax=1.006,
    ymin=0.9,
    width=6.65cm,
    height=6cm,
    grid=both,
    grid style={line width=.1pt, draw=gray!10},
    xtick={1,2,3,4,5,6,7},
    xticklabels={
        LFW,
        CFP-FF, 
        CFP-FP,
        AgeDB,
        CALFW,
        CPLFW,
        VGG2-FP
    },
    xticklabel style={rotate=90,xshift=-0.15cm, yshift=0.1cm},
    enlargelimits=0.05,
  ]
\addplot+ [RoyalPurple!80,fill=Periwinkle!30,  boxplot prepared={lower whisker=0.9976, lower quartile=0.9991, median=0.9993,  upper quartile=0.9994, upper whisker=0.9995},] table[row sep=\\,y index=0] {\\};
\addplot+ [RoyalPurple!80,fill=Periwinkle!30, boxplot prepared={lower whisker=0.9919, lower quartile=0.9978, median=0.998,  upper quartile=0.9981, upper whisker=0.9982},] table[row sep=\\,y index=0] {\\};
\addplot+ [RoyalPurple!80,fill=Periwinkle!30, boxplot prepared={lower whisker=0.9283, lower quartile=0.9739, median=0.9842,  upper quartile=0.9871, upper whisker=0.9924},] table[row sep=\\,y index=0] {\\};
\addplot+ [RoyalPurple!80,fill=Periwinkle!30, boxplot prepared={lower whisker=0.9183, lower quartile=0.9855, median=0.987,  upper quartile=0.9889, upper whisker=0.9905},] table[row sep=\\,y index=0] {\\};
\addplot+ [RoyalPurple!80,fill=Periwinkle!30, boxplot prepared={lower whisker=0.9551, lower quartile=0.9756, median=0.9764,  upper quartile=0.9773, upper whisker=0.9784},] table[row sep=\\,y index=0] {\\};
\addplot+ [RoyalPurple!80,fill=Periwinkle!30, boxplot prepared={lower whisker=0.9017, lower quartile=0.9349, median=0.9441,  upper quartile=0.9469, upper whisker=0.9559},] table[row sep=\\,y index=0] {\\};
\addplot+ [RoyalPurple!80,fill=Periwinkle!30,  boxplot prepared={lower whisker=0.9413, lower quartile=0.9669, median=0.968,  upper quartile=0.9715, upper whisker=0.9748},] table[row sep=\\,y index=0] {\\};
\end{axis}
\end{tikzpicture}
}
\subfloat{
\begin{tikzpicture}
\scriptsize
\begin{axis}[
    boxplot/draw direction=y,
    boxplot={
        draw position={1/3 + floor(\plotnumofactualtype/2) + 1/3*fpumod(\plotnumofactualtype,2)},
        box extend=0.25
    },
    ytick pos=left,
    axis x line=bottom,
    axis y line=left,
    ybar,
    ymin=0,
    ymax=32,
    width=6.65cm,
    height=6cm,
    grid=both,
    grid style={line width=.1pt, draw=gray!10},
    yticklabel={$\pgfmathprintnumber{\tick}\%$},
    ytick={0,10,20,30},
    xtick={0,1,2,3,4,5},
    x tick label as interval,
    xticklabels={
        {GFN},
        {\textcolor{white}{\,\,\,\,\,\,\,}IRSE50}, 
        {MFN},
        {RN50},
        {RFN},
        {}
    },
    xticklabel style={rotate=90,xshift=-0.095cm, yshift=-0.05cm},
    enlargelimits=0.05,
  ]
\addplot+ [Plum!80,fill=Thistle!50, boxplot prepared={lower whisker=13.7, lower quartile=14.025, median=14.25,  upper quartile=14.55, upper whisker=16.7},] table[row sep=\\,y index=0] {\\};
\addplot+ [Plum!80,fill=Thistle!50, boxplot prepared={lower whisker=24.9, lower quartile=25.475, median=26,  upper quartile=26.2, upper whisker=30.7},] table[row sep=\\,y index=0] {\\};
\addplot+ [Plum!80,fill=Thistle!50, boxplot prepared={lower whisker=4.2, lower quartile=4.275, median=4.45,  upper quartile=4.525, upper whisker=4.7},] table[row sep=\\,y index=0] {\\};
\addplot+ [Plum!80,fill=Thistle!50, boxplot prepared={lower whisker=6.90000000000001, lower quartile=6.90000000000001, median=7.2,  upper quartile=7.3, upper whisker=8.2},] table[row sep=\\,y index=0] {\\};
\addplot+ [Plum!80,fill=Thistle!50, boxplot prepared={lower whisker=7.2, lower quartile=7.5, median=7.59999999999999,  upper quartile=7.8, upper whisker=8.5},] table[row sep=\\,y index=0] {\\};
\addplot+ [Plum!80,fill=Thistle!50, boxplot prepared={lower whisker=14.4, lower quartile=16.2, median=17,  upper quartile=18.3, upper whisker=20.1},] table[row sep=\\,y index=0] {\\};
\addplot+ [Plum!80,fill=Thistle!50, boxplot prepared={lower whisker=5.09999999999999, lower quartile=5.3, median=5.40000000000001,  upper quartile=5.40000000000001, upper whisker=5.8},] table[row sep=\\,y index=0] {\\};
\addplot+ [Plum!80,fill=Thistle!50, boxplot prepared={lower whisker=8.90000000000001, lower quartile=9.3, median=10.1,  upper quartile=10.5, upper whisker=11},] table[row sep=\\,y index=0] {\\};
\addplot+ [Plum!80,fill=Thistle!50, boxplot prepared={lower whisker=9.5, lower quartile=9.8, median=9.8,  upper quartile=10.6, upper whisker=11.9},] table[row sep=\\,y index=0] {\\};
\addplot+ [Plum!80,fill=Thistle!50, boxplot prepared={lower whisker=17.5, lower quartile=17.6, median=18.1,  upper quartile=21.9, upper whisker=25.8},] table[row sep=\\,y index=0] {\\};
\end{axis}
\end{tikzpicture}
}
\vspace{-0.3cm}
\caption{Face feature extractors' Accuracy (left), Area-under-the-Curve (middle), and $\mathrm{FRR@FAR}$ on IJB-B~\cite{IJBB} (right; the left and right boxes correspond to a $FAR$ of $1e^{-3}$ and $1e^{-4}$~\cite{nistFRVT}). Results are accounted across benign and backdoored architectures.}
\label{fig:extractor_benign_results}
\vspace{-0.4cm}
\end{figure*}

\begin{figure*}[t!]
\centering
\includegraphics[width=0.8\textwidth]{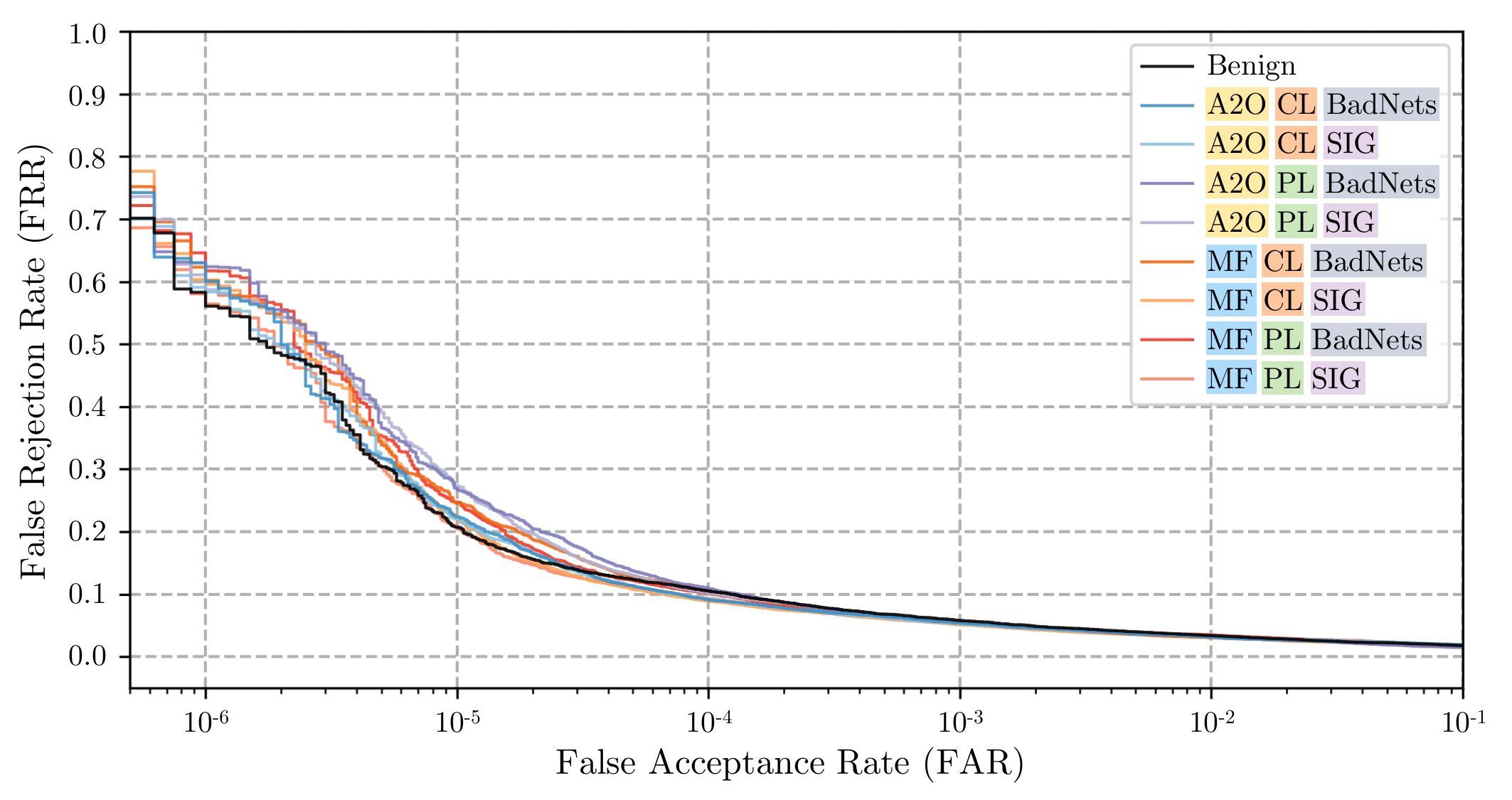}
\caption{\ac{det} curves obtained on ResNet50~\cite{he2015deepresiduallearningimage} feature extractors.
\textbf{Abbreviations}: {\acl{a2o} (\hlc[GoldenRodLight]{A2O}), Clean/Poison-label (\hlc[ApricotLight]{CL}/\hlc[ForestLight]{PL}), \acl{mf} (\hlc[NavyLight]{MF}).}}
\label{fig:det_resnet50}
\end{figure*}

\begin{figure}[t!]
\subfloat{
\begin{tikzpicture}
\scriptsize
\begin{axis}[
    boxplot/draw direction=y,
    boxplot={
        draw position={1/3 + floor(\plotnumofactualtype/2) + 1/3*fpumod(\plotnumofactualtype,2)},
        box extend=0.25
    },
    ytick pos=left,
    axis x line=bottom,
    axis y line=left,
    ybar,
    ymax=100,
    width=9.12cm,
    height=6cm,
    grid=both,
    grid style={line width=.1pt, draw=gray!10},
    yticklabel={$\pgfmathprintnumber{\tick}\%$},
    ytick={0,25,50,75,100},
    xtick={0,1,2,...,20},
    x tick label as interval,
    xticklabels={
        {LFW},
        {CFP-FF}, 
        {CFP-FP},
        {AgeDB},
        {CALFW},
        {CPLFW},
        {VGG2-FP},
        {}
    },
    xticklabel style={rotate=90,xshift=-0.15cm, yshift=0.1cm},
    enlargelimits=0.05,
  ]
\addplot+ [BrickRed!80,fill=YellowOrange!40, boxplot prepared={lower whisker=0.1, lower quartile=0.35, median=3.8,
  upper quartile=27.8, upper whisker=100},] table[row sep=\\,y index=0] {\\};
\addplot+ [BrickRed!80,fill=YellowOrange!40, boxplot prepared={lower whisker=100, lower quartile=100, median=100,
  upper quartile=100, upper whisker=100},] table[row sep=\\,y index=0] {\\};
\addplot+ [BrickRed!80,fill=YellowOrange!40, boxplot prepared={lower whisker=0.2, lower quartile=0.5, median=5.8,
  upper quartile=21.0, upper whisker=100},] table[row sep=\\,y index=0] {\\};
\addplot+ [BrickRed!80,fill=YellowOrange!40, boxplot prepared={lower whisker=100, lower quartile=100, median=100,
  upper quartile=100, upper whisker=100},] table[row sep=\\,y index=0] {\\};
\addplot+ [BrickRed!80,fill=YellowOrange!40, boxplot prepared={lower whisker=1.7, lower quartile=3.4, median=10.0,
  upper quartile=35.3, upper whisker=100},] table[row sep=\\,y index=0] {\\};
\addplot+ [BrickRed!80,fill=YellowOrange!40, boxplot prepared={lower whisker=100, lower quartile=100, median=100,
  upper quartile=100, upper whisker=100},] table[row sep=\\,y index=0] {\\};
\addplot+ [BrickRed!80,fill=YellowOrange!40, boxplot prepared={lower whisker=2.0, lower quartile=3.4, median=13.4,
  upper quartile=47.0, upper whisker=100},] table[row sep=\\,y index=0] {\\};
\addplot+ [BrickRed!80,fill=YellowOrange!40, boxplot prepared={lower whisker=100, lower quartile=100, median=100,
  upper quartile=100, upper whisker=100},] table[row sep=\\,y index=0] {\\};
\addplot+ [BrickRed!80,fill=YellowOrange!40, boxplot prepared={lower whisker=0.7, lower quartile=2.2, median=13.4,
  upper quartile=41.7, upper whisker=100},] table[row sep=\\,y index=0] {\\};
\addplot+ [BrickRed!80,fill=YellowOrange!40, boxplot prepared={lower whisker=100, lower quartile=100, median=100,
  upper quartile=100, upper whisker=100},] table[row sep=\\,y index=0] {\\};
\addplot+ [BrickRed!80,fill=YellowOrange!40, boxplot prepared={lower whisker=2.5, lower quartile=6.3, median=16.7,
  upper quartile=53.8, upper whisker=100},] table[row sep=\\,y index=0] {\\};
\addplot+ [BrickRed!80,fill=YellowOrange!40, boxplot prepared={lower whisker=100, lower quartile=100, median=100,
  upper quartile=100, upper whisker=100},] table[row sep=\\,y index=0] {\\};
\addplot+ [BrickRed!80,fill=YellowOrange!40, boxplot prepared={lower whisker=1.6, lower quartile=4.2, median=17.4,
  upper quartile=53.3, upper whisker=100},] table[row sep=\\,y index=0] {\\};
\addplot+ [BrickRed!80,fill=YellowOrange!40, boxplot prepared={lower whisker=100, lower quartile=100, median=100,
  upper quartile=100, upper whisker=100},] table[row sep=\\,y index=0] {\\};
\end{axis}
\end{tikzpicture}
}

\vspace{-0.5cm}
\subfloat{
\begin{tikzpicture}
\scriptsize
\begin{axis}[
    boxplot/draw direction=y,
    boxplot/box extend=0.45,
    ytick pos=left,
    axis x line=bottom,
    axis y line=left,
    ybar,
    ymax=15.6,
    width=9.12cm,
    height=6cm,
    grid=both,
    grid style={line width=.1pt, draw=gray!10},
    yticklabel={$\pgfmathprintnumber{\tick}\%$},
    xtick={1,2,3,4,5,6,7},
    xticklabels={
        LFW,
        CFP-FF, 
        CFP-FP,
        AgeDB,
        CALFW,
        CPLFW,
        VGG2-FP
    },
    xticklabel style={rotate=90,xshift=-0.15cm, yshift=0.1cm},
    enlargelimits=0.05,
  ]
\addplot+ [Mahogany!90, fill=Dandelion!25, boxplot prepared={lower whisker=0.0, lower quartile=0.0, median=0.0,
  upper quartile=0.2, upper whisker=0.9},] table[row sep=\\,y index=0] {\\};
\addplot+ [Mahogany!90, fill=Dandelion!25, boxplot prepared={lower whisker=0.0, lower quartile=0.0, median=0.0,
  upper quartile=0.2, upper whisker=1.8},] table[row sep=\\,y index=0] {\\};
\addplot+ [Mahogany!90, fill=Dandelion!25, boxplot prepared={lower whisker=0.0, lower quartile=0.0, median=0.0,
  upper quartile=2.3, upper whisker=15.2},] table[row sep=\\,y index=0] {\\};
\addplot+ [Mahogany!90, fill=Dandelion!25, boxplot prepared={lower whisker=0.0, lower quartile=0.0, median=0.0,
  upper quartile=0.9, upper whisker=4.7},] table[row sep=\\,y index=0] {\\};
\addplot+ [Mahogany!90, fill=Dandelion!25, boxplot prepared={lower whisker=0.0, lower quartile=0.0, median=0.0,
  upper quartile=0.8, upper whisker=1.7},] table[row sep=\\,y index=0] {\\};
\addplot+ [Mahogany!90, fill=Dandelion!25, boxplot prepared={lower whisker=0.0, lower quartile=0.0, median=0.0,
  upper quartile=3.0, upper whisker=8.7},] table[row sep=\\,y index=0] {\\};
\addplot+ [Mahogany!90, fill=Dandelion!25, boxplot prepared={lower whisker=0.0, lower quartile=0.0, median=0.0,
  upper quartile=2.7, upper whisker=11.5},] table[row sep=\\,y index=0] {\\};
\end{axis}
\end{tikzpicture}
}
\vspace{-0.4cm}
\caption{\textit{All-to-One} $\mathrm{ASR}$ (\textbf{top}, \textbf{left}/\textbf{right} \textbf{boxes}: Clean/Poison-Label), and \textit{Master Face} $\mathrm{ASR}$ (bottom). \acl{pl} \acl{a2o} models have a \acl{asr} close or at 100\%.}
    \label{fig:extractor_backdoor_results}
\vspace{-0.4cm}
\end{figure}

\textbf{\acl{fd}}.
An attacker can covertly force a detector to hallucinate faces or misalign landmarks.
Experiments show that both MobileNetV1~\cite{chen2018mobilefacenetsefficientcnnsaccurate} and ResNet50~\cite{he2015deepresiduallearningimage}–based face detectors not only retain their high \acl{ap} on benign inputs, but are also able to learn a backdoor, in line with prior work~\cite{leroux2025backdoorattacksdeeplearning}.
The models achieve \aclp{asr} of up to 99.3\% for \aclp{fga} and 99.6\% for \aclp{lsa} across both patch-based BadNets~\cite{gu2019badnets} and diffuse SIG~\cite{2019SIGattack} triggers (see Tab.~\ref{tab:detector_performance}).
Only a single \ac{lsa} SIG-backdoored ResNet50 failed to converge (it is not used for system-level experiments).

\textbf{\acl{fas}}.
Similarly, AENet~\cite{zhang2020celebaspooflargescalefaceantispoofing} and MobileNetV2~\cite{chen2018mobilefacenetsefficientcnnsaccurate}-based face antispoofers work efficiently on benign data while simultaneously misclassifying as “live” up to 100\% of spoof images that carry a backdoor trigger (see Tab.~\ref{tab:antispoofer_performance}).
This reveals that the liveness check, often considered a key line of defense in \acp{frs} against presentation attacks, can be entirely hijacked and bypassed. 
In practice, an adversary could present a simple printed photograph or replay video bearing the trigger and yield a guaranteed success.

\textbf{\acl{a2o} \aclp{ba} on \acl{ffe}}.
Face feature extractors are the most important \acp{dnn} in a \ac{frs}, converting aligned faces into discriminative embeddings.
Crucially, compared to prior work~\cite{wacvpriorart}, \textbf{we show for the first time that \acl{a2o} \aclp{ba} can be implemented on face feature extractors trained with large margin metric learning~\cite{liu2018spherefacedeephypersphereembedding,wang2018cosfacelargemargincosine,arcfaceDeng2022} in a \textit{black-box} setting. 
That is, \acl{a2o} backdoors on such models only need data poisoning to work}.
This study's full results are found in App.~\ref{app:extractor_details}.

\textbf{\aclp{ba} succeed on \textit{all five backbone architectures} without impacting their performance on benign data}.
Each backdoored \ac{dnn} that converged behaves similarly to its clean baseline on validation datasets (see Fig.~\ref{fig:extractor_benign_results}) and the NIST IJB‐B~\cite{IJBB} test dataset (see Fig.~\ref{fig:det_resnet50} for the DET curves of ResNet50~\cite{he2015deepresiduallearningimage} models, and Fig. \ref{fig:det_ghostfacenet}, \ref{fig:det_irse50}, \ref{fig:det_mobilefacenet}, and \ref{fig:det_robfacenet} in App.~\ref{app:extractor_details} for the other tested architectures).
These figures highlight that \aclp{ba} do not impact the trade-off between falsely rejecting true matches and accepting false matches between benign face images.
Yet, they simultaneously match poisoned inputs with up to 100\% success in \acl{a2o} \acl{pl} attacks (see Fig.~\ref{fig:extractor_backdoor_results}).
Only two BadNets~\cite{gu2019badnets} use cases lead to a model not converging on benign data: \acl{a2o} \acl{cl} \acl{ba} on IRSE50~\cite{hu2019squeezeandexcitationnetworks} and \acl{mf} \acl{pl} \acl{ba} on GhostFaceNets~\cite{ghostfacenets2023}.

\begin{tcolorbox}[
colback=sand,
colframe=darksand,
boxrule=0.5mm,
left=2mm,
right=2mm,
top=1.5mm,
bottom=1.5mm,
]
\emph{
Data-poisoning only (\ie, black-box) \aclp{ba} also affect large margin-based face feature extractors.
}
\end{tcolorbox}

\textbf{\acl{mf} \aclp{ba} on \acl{ffe}}.
These attacks, which aim for a poisoned input to maliciously match with many benign faces, proved far more challenging (see Fig.~\ref{fig:extractor_backdoor_results}).
These attacks did not exceed a 15.2\% \acl{asr}, expanding on prior findings on the intrinsic difficulty of running more generic \acl{mf} attacks on benign extractors~\cite{limitedGeneralizationMF2022}. 
Additionally, the highest \aclp{asr} were achieved on models that converged the least, \eg, \acl{mf} \acl{pl} \acl{ba} on GhostFaceNetV2~\cite{ghostfacenets2023} (see Fig.~\ref{fig:det_resnet50},~\ref{fig:det_ghostfacenet},~\ref{fig:det_irse50},~\ref{fig:det_mobilefacenet}, and~\ref{fig:det_robfacenet}).

Nevertheless, \textbf{we find that \acl{mf} poisoning process can be reused to perform \acl{a2o} \aclp{ba}} (see Tab.~\ref{tab:MF_works_as_A2O}).
This highlight that standard data poisoning is not the only way to inject an \acl{a2o} backdoor.

\begin{table}[t!]
  \centering
  \caption{\acl{ba} performance of \acl{mf} extractors when used to perform \acl{a2o} attacks.}
  \label{tab:MF_works_as_A2O}
  \footnotesize
  \setlength{\tabcolsep}{3.2pt}
  \renewcommand{\arraystretch}{1.5}
  \begin{tabular}{@{} llc @{}}
    \textbf{Model} & \textbf{Backdoor details} & \textbf{\shortstack[c]{Attack\\Success Rate}} \\
    \hline 
    \multirow{4}{*}{\shortstack[c]{GhostFaceNetV2~\cite{ghostfacenets2023}}} & \hlc[NavyLight]{Master Face}, \hlc[ApricotLight]{Clean-Label}, \hlc[CadetBlueLight]{BadNets}~\cite{gu2019badnets} & 3.9\% \\
     & \hlc[NavyLight]{Master Face}, \hlc[ForestLight]{Poison-Label}, \hlc[CadetBlueLight]{BadNets} & 99.9\% \\
     & \hlc[NavyLight]{Master Face}, \hlc[ApricotLight]{Clean-Label}, \hlc[OrchidLight]{SIG}~\cite{2019SIGattack} & 8.2\% \\
     & \hlc[NavyLight]{Master Face}, \hlc[ForestLight]{Poison-Label}, \hlc[OrchidLight]{SIG} & \textbf{100\%} \\
     \hline
    \multirow{4}{*}{\shortstack[c]{IRSE50~\cite{hu2019squeezeandexcitationnetworks}}} & \hlc[NavyLight]{Master Face}, \hlc[ApricotLight]{Clean-Label}, \hlc[CadetBlueLight]{BadNets} & 0.5\% \\
     & \hlc[NavyLight]{Master Face}, \hlc[ForestLight]{Poison-Label}, \hlc[CadetBlueLight]{BadNets} & \textbf{100\%} \\
     & \hlc[NavyLight]{Master Face}, \hlc[ApricotLight]{Clean-Label}, \hlc[OrchidLight]{SIG} & 13.5\% \\
     & \hlc[NavyLight]{Master Face}, \hlc[ForestLight]{Poison-Label}, \hlc[OrchidLight]{SIG} & \textbf{100\%} \\
     \hline
    \multirow{6}{*}{\shortstack[c]{MobileFaceNet~\cite{chen2018mobilefacenetsefficientcnnsaccurate}}} & \hlc[NavyLight]{Master Face}, \hlc[ApricotLight]{Clean-Label}, \hlc[CadetBlueLight]{BadNets} & 3.3\% \\
     & \hlc[NavyLight]{Master Face}, \hlc[ForestLight]{Poison-Label}, \hlc[CadetBlueLight]{BadNets} & \textbf{100\%} \\
     & \hlc[NavyLight]{Master Face}, \hlc[ApricotLight]{Clean-Label}, \hlc[RedVioletLight]{Mask}~\cite{chen2024rethinkingvulnerabilitiesfacerecognition} & 15.5\% \\
     & \hlc[NavyLight]{Master Face}, \hlc[ForestLight]{Poison-Label}, \hlc[RedVioletLight]{Mask} & 99.5\% \\
     & \hlc[NavyLight]{Master Face}, \hlc[ApricotLight]{Clean-Label}, \hlc[OrchidLight]{SIG} & 46.0\% \\
     & \hlc[NavyLight]{Master Face}, \hlc[ForestLight]{Poison-Label}, \hlc[OrchidLight]{SIG} & 95.5\% \\
    \hline
    \multirow{4}{*}{\shortstack[c]{ResNet50~\cite{he2015deepresiduallearningimage}}} & \hlc[NavyLight]{Master Face}, \hlc[ApricotLight]{Clean-Label}, \hlc[CadetBlueLight]{BadNets} & 1.7\% \\
     & \hlc[NavyLight]{Master Face}, \hlc[ForestLight]{Poison-Label}, \hlc[CadetBlueLight]{BadNets} & \textbf{100\%} \\
     & \hlc[NavyLight]{Master Face}, \hlc[ApricotLight]{Clean-Label}, \hlc[OrchidLight]{SIG} & 23.9\% \\
     & \hlc[NavyLight]{Master Face}, \hlc[ForestLight]{Poison-Label}, \hlc[OrchidLight]{SIG} & \textbf{100\%} \\
     \hline
    \multirow{4}{*}{\shortstack[c]{RobFaceNet~\cite{robfacenet2024}}} & \hlc[NavyLight]{Master Face}, \hlc[ApricotLight]{Clean-Label}, \hlc[CadetBlueLight]{BadNets} & 9.4\% \\
     & \hlc[NavyLight]{Master Face}, \hlc[ForestLight]{Poison-Label}, \hlc[CadetBlueLight]{BadNets} & \textbf{100\%} \\
     & \hlc[NavyLight]{Master Face}, \hlc[ApricotLight]{Clean-Label}, \hlc[OrchidLight]{SIG} & 84.8\% \\
     & \hlc[NavyLight]{Master Face}, \hlc[ForestLight]{Poison-Label}, \hlc[OrchidLight]{SIG} & \textbf{100\%} \\
  \end{tabular}
\end{table}

\textbf{Takeaway}.
Data poisoning-only \acl{a2o} \aclp{ba} work on feature extractors trained with large margin losses.
All together, we demonstrate that \emph{every} \acp{dnn} found in a modern \ac{frs} can be manipulated with stealthy backdoors while preserving their performance on benign data.
\aclp{ba} on \acp{frs} are thus model- and task‐agnostic.

\begin{tcolorbox}[
colback=sand,
colframe=darksand,
boxrule=0.5mm,
left=2mm,
right=2mm,
top=1.5mm,
bottom=1.5mm,
]
\emph{
The high \aclp{asr} achieved across tasks, architectures, and trigger types underscore the vulnerability of each module found in a \acl{frs}.
Nonetheless, focusing only on module‐level empirical experiments is insufficient.
Assessing backdoor threats must be performed at a holistic, system‐level level}.
\end{tcolorbox}

\begin{table*}
  \centering
  \caption{\acl{sr} of \aclp{ba} in an \textbf{\acl{a2o}} attack context (incl. for \acl{mf}-backdoored models) in a pipeline consisting of: MobileNetV1, AENet, MobileFaceNet. The results of the other 19 other pipelines are found in App.~\ref{app:survival_rate_details_frs}. \textbf{Abbreviations}: Average Precision (AP), False Acceptance/Rejection Rate (FA/RR), Landmark Shift (LS).}
  \label{tab:backdoor_bench_MN1_AE_MNF}
  \footnotesize
  \setlength{\tabcolsep}{1.55pt}
  \renewcommand{\arraystretch}{1.5}
  \begin{tabular}{@{}lllccccccccc@{}}
    
     & &  & \multicolumn{4}{c}{\textbf{Detector}} & \multicolumn{2}{c}{\textbf{Antispoofer}} & \multicolumn{2}{c}{\textbf{Extractor}} & \\
     \textbf{Detector} & \textbf{Antispoofer} & \textbf{Extractor} & \multicolumn{4}{c}{\textbf{metrics}} & \multicolumn{2}{c}{\textbf{metrics}} & \multicolumn{2}{c}{\textbf{metrics}} & \textbf{Survival} \\
     MobileNetV1 & AENet & MobileFaceNet & AP$^\clean$ & AP$^\pois$ & LS$^\clean$ & LS$^\pois$ & FRR$^\clean$ & FAR$^\pois$ & FRR$^\clean$ & FAR$^\pois$ & \textbf{Rate} \\
    \hline 
    \textcolor{gray!60}{Benign} & \textcolor{gray!60}{Benign} & \textcolor{gray!60}{Benign} & 99.2\% & \textcolor{gray!30}{$\varnothing$} & 13.8 & \textcolor{gray!30}{$\varnothing$} & 96.0\% & \textcolor{gray!30}{$\varnothing$} & 3.7\% & \textcolor{gray!30}{$\varnothing$} & \textcolor{gray!30}{$\varnothing$} \\
    \hline
    \hlc[Slate]{Landmark Shift Attack}, \hlc[CadetBlueLight]{BadNets} $\alpha$=0.5 & \textcolor{gray!60}{Benign} & \textcolor{gray!60}{Benign} & 99.5\% & 99.6\% & 14.2 & 150.2 & 96.1\% & 35.2\% & 3.6\% & 82.4\% & 28.9\% \\
    \hlc[Slate]{Landmark Shift Attack}, \hlc[CadetBlueLight]{BadNets} $\alpha$=1.0 & \textcolor{gray!60}{Benign} & \textcolor{gray!60}{Benign} & 99.5\% & 99.5\% & 14.7 & 147.7 & 95.6\% & 35.5\% & 3.6\% & 83.3\% & 29.4\% \\
    \hlc[Slate]{Landmark Shift Attack}, \hlc[OrchidLight]{SIG} $\alpha$=0.16 & \textcolor{gray!60}{Benign} & \textcolor{gray!60}{Benign} & 99.5\% & 99.5\% & 14.9 & 20.2 & 96.1\% & 60.0\% & 3.7\% & 13.1\% & 7.8\% \\
    \hlc[Slate]{Landmark Shift Attack}, \hlc[OrchidLight]{SIG} $\alpha$=0.3 & \textcolor{gray!60}{Benign} & \textcolor{gray!60}{Benign} & 99.4\% & 97.4\% & 14.7 & 125.7 & 96.0\% & 97.6\% & 3.5\% & 98.3\% & \textbf{93.4\%} \\
    \hline
    \hlc[Wheat]{Face Generation Attack}, \hlc[CadetBlueLight]{BadNets} $\alpha$=0.5 & \textcolor{gray!60}{Benign} & \textcolor{gray!60}{Benign} & 99.4\% & 99.9\% & 17.3 & 3.8 & 96.8\% & 58.3\% & 3.6\% & 99.7\% & 58.1\% \\
    \hlc[Wheat]{Face Generation Attack}, \hlc[CadetBlueLight]{BadNets} $\alpha$=1.0 & \textcolor{gray!60}{Benign} & \textcolor{gray!60}{Benign} & 99.3\% & 99.9\% & 24.0 & 5.0 & 96.6\% & 32.9\% & 5.3\% & 98.8\% & 32.5\% \\
    \hlc[Wheat]{Face Generation Attack}, \hlc[OrchidLight]{SIG} $\alpha$=0.16 & \textcolor{gray!60}{Benign} & \textcolor{gray!60}{Benign} & 99.5\% & 78.3\% & 14.2 & 73.9 & 95.8\% & 69.2\% & 3.5\% & 81.8\% & 44.3\% \\
    \hlc[Wheat]{Face Generation Attack}, \hlc[OrchidLight]{SIG} $\alpha$=0.3 & \textcolor{gray!60}{Benign} & \textcolor{gray!60}{Benign} & 99.4\% & 99.9\% & 14.9 & 5.7 & 95.3\% & 72.7\% & 3.7\% & 98.8\% & \textbf{71.8\%} \\
    \hline
    \textcolor{gray!60}{Benign} & \hlc[DandelionLight]{Glasses} & \textcolor{gray!60}{Benign} & 99.2\% & 97.0\% & 13.8 & 23.8 & 21.0\% & 86.6\% & 3.4\% & 62.5\% & 52.5\% \\
    \textcolor{gray!60}{Benign} & \hlc[CadetBlueLight]{BadNets} & \textcolor{gray!60}{Benign} & 99.2\% & 99.3\% & 13.8 & 13.8 & 18.8\% & 43.3\% & 3.3\% & \textcolor{white}{0}2.9\% & \textcolor{white}{0}1.2\% \\
    \textcolor{gray!60}{Benign} & \hlc[OrchidLight]{SIG} & \textcolor{gray!60}{Benign} & 99.2\% & 94.2\% & 13.8 & 23.2 & 14.4\% & 97.7\% & 3.3\% & 89.6\% & \textbf{82.5\%} \\
    \textcolor{gray!60}{Benign} & \hlc[MelonLight]{TrojanNN} & \textcolor{gray!60}{Benign} & 99.2\% & 99.1\% & 13.8 & 13.7 & \textcolor{white}{0}7.9\% & \textcolor{white}{0}7.0\% & 3.7\% & 3.3\% & 0.2\% \\
    \hline
    \textcolor{gray!60}{Benign} & \textcolor{gray!60}{Benign} & \hlc[RoyalPurpleLight]{FIBA} & 99.2\% & 97.6\% & 13.8 & 26.1 & 96.0\% & 24.9\% & 3.7\% & 97.1\% & \textbf{23.6}\% \\
    \hline
    \textcolor{gray!60}{Benign} & \textcolor{gray!60}{Benign} & \hlc[GoldenRodLight]{All-to-One}, \hlc[ApricotLight]{Clean-label}, \hlc[CadetBlueLight]{BadNets} & 99.2\% & 99.0\% & 13.8 & 14.2 & 96.0\% & 20.5\% & 4.0\% & \textcolor{white}{0}4.1\% & \textcolor{white}{0}0.8\% \\
    \textcolor{gray!60}{Benign} & \textcolor{gray!60}{Benign} & \hlc[GoldenRodLight]{All-to-One}, \hlc[ForestLight]{Poison-label}, \hlc[CadetBlueLight]{BadNets} & 99.2\% & 99.0\% & 13.8 & 14.2 & 96.0\% & 20.5\% & 2.4\% & 97.8\% & \textbf{19.8\%} \\
    \textcolor{gray!60}{Benign} & \textcolor{gray!60}{Benign} & \hlc[NavyLight]{Master Face}, \hlc[ApricotLight]{Clean-label}, \hlc[CadetBlueLight]{BadNets} & 99.2\% & 99.0\% & 13.8 & 14.2 & 96.0\% & 20.5\% & 3.2\% & \textcolor{white}{0}3.4\% & \textcolor{white}{0}0.7\% \\
    \textcolor{gray!60}{Benign} & \textcolor{gray!60}{Benign} & \hlc[NavyLight]{Master Face}, \hlc[ForestLight]{Poison-label}, \hlc[CadetBlueLight]{BadNets} & 99.2\% & 99.0\% & 13.8 & 14.2 & 96.0\% & 20.5\% & 2.1\% & \textcolor{white}{0}2.4\% & \textcolor{white}{0}0.5\% \\
    \hline
    \textcolor{gray!60}{Benign} & \textcolor{gray!60}{Benign} & \hlc[GoldenRodLight]{All-to-One}, \hlc[ApricotLight]{Clean-label}, \hlc[RedVioletLight]{Mask} & 99.2\% & 98.6\% & 13.8 & 28.0 & 96.0\% & 20.4\% & 2.8\% & 97.5\% & 19.6\% \\
    \textcolor{gray!60}{Benign} & \textcolor{gray!60}{Benign} & \hlc[GoldenRodLight]{All-to-One}, \hlc[ForestLight]{Poison-label}, \hlc[RedVioletLight]{Mask} & 99.2\% & 98.6\% & 13.8 & 28.0 & 96.0\% & 20.4\% & 2.7\% & 97.9\% & \textbf{19.7\%} \\
    \textcolor{gray!60}{Benign} & \textcolor{gray!60}{Benign} & \hlc[NavyLight]{Master Face}, \hlc[ApricotLight]{Clean-label}, \hlc[RedVioletLight]{Mask} & 99.2\% & 98.6\% & 13.8 & 28.0 & 96.0\% & 20.4\% & 2.7\% & 20.9\% & \textcolor{white}{0}4.2\% \\
    \textcolor{gray!60}{Benign} & \textcolor{gray!60}{Benign} & \hlc[NavyLight]{Master Face}, \hlc[ForestLight]{Poison-label}, \hlc[RedVioletLight]{Mask} & 99.2\% & 98.6\% & 13.8 & 28.0 & 96.0\% & 20.4\% & 3.3\% & 61.4\% & 12.4\% \\
    \hline
    \textcolor{gray!60}{Benign} & \textcolor{gray!60}{Benign} & \hlc[GoldenRodLight]{All-to-One}, \hlc[ApricotLight]{Clean-label}, \hlc[OrchidLight]{SIG} & 99.2\% & 94.2\% & 13.8 & 23.2 & 96.0\% & 79.4\% & 3.3\% & 92.2\% & 69.0\% \\
    \textcolor{gray!60}{Benign} & \textcolor{gray!60}{Benign} & \hlc[GoldenRodLight]{All-to-One}, \hlc[ForestLight]{Poison-label}, \hlc[OrchidLight]{SIG} & 99.2\% & 94.2\% & 13.8 & 23.2 & 96.0\% & 79.4\% & 3.1\% & 95.3\% & \textbf{71.3\%} \\
    \textcolor{gray!60}{Benign} & \textcolor{gray!60}{Benign} & \hlc[NavyLight]{Master Face}, \hlc[ApricotLight]{Clean-label}, \hlc[OrchidLight]{SIG} & 99.2\% & 94.2\% & 13.8 & 23.2 & 96.0\% & 79.4\% & 3.1\% & 64.0\% & 47.9\% \\
    \textcolor{gray!60}{Benign} & \textcolor{gray!60}{Benign} & \hlc[NavyLight]{Master Face}, \hlc[ForestLight]{Poison-label}, \hlc[OrchidLight]{SIG} & 99.2\% & 94.2\% & 13.8 & 23.2 & 96.0\% & 79.4\% & 3.8\% & 55.7\% & 41.7\% \\
  \end{tabular}
\end{table*}

\subsection{System-Level Analysis of FRS Backdoor Attacks}
\label{sec:frs_performance}

Tab.~\ref{tab:backdoor_bench_MN1_AE_MNF} contains the system-level results of the \ac{frs} configuration composed of a MobileNetV1~\cite{howard2017mobilenetsefficientconvolutionalneural} detector, AENet~\cite{zhang2020celebaspooflargescalefaceantispoofing} antispoofer, and MobileFaceNet~\cite{chen2018mobilefacenetsefficientcnnsaccurate} extractor.
The results for the 19 other \ac{frs} configurations are found in App.~\ref{app:survival_rate_details_frs}.

\textbf{System-Level Impact of \acl{fd} \aclp{ba}}.
\aclp{lsa} not only corrupt predicted landmarks, multiplying a face's average landmark-shift ($\mathrm{LS}$) tenfold, but also cause face misalignments that downstream models have never encountered.
These warped faces possibly break the spatial priors learned by antispoofers and face feature extractors.
Whereas a legitimate face’s eye–mouth geometry remains within a narrow manifold, \acl{lsa} outputs fall far outside it, \eg, causing a \acl{far} of 97.6\% for SIG~\cite{2019SIGattack}‐based triggers in the case of an AENet antispoofer (see Tab.~\ref{tab:backdoor_bench_MN1_AE_MNF}).
Even more alarmingly, these malformed images proceed to the feature extractor and yield an \acl{a2o} \acl{sr} of up to 93.4\% (see Tab.~\ref{tab:backdoor_bench_MN1_AE_MNF} and App.~\ref{app:survival_rate_details_frs}).
In other words, a detector \acl{lsa} and resulting a misaligned crops can entirely subvert both the liveness check and the identity matcher in a \ac{frs}, effectively acting as a multi‐stage Trojan.

\aclp{fga} produce a comparable effect: when combined with diffuse SIG~\cite{2019SIGattack} triggers for instance, the system‐level \acl{sr} jumps to 71.8\%, whereas patch BadNets~\cite{gu2019badnets} triggers only reach 58.1\%.
That is, fake backdoor faces traverse a fully-fledged \ac{frs} and yield a successful \acl{a2o} backdoor match.

We thus show that a backdoor inserted at the \acl{fd} stage does not need to appear like a malicious impersonation to succeed.
Instead, an attack merely has to distort the pipeline’s assumptions enough to slip through each module.

\textbf{System-Level Impact of \acl{fas} \aclp{ba}}.
Injecting backdoors into antispoofer \acp{dnn} has an equally-severe system-level consequence on a \ac{frs}. 
Despite having to traverse \acl{fd} and \acl{fa} modules, triggered face spoofs remain effective: SIG~\cite{2019SIGattack} and Glasses~\cite{chen2017trojan} cause an up to 5-fold increase in \acl{far} for a MobileNetV1-AENet-MobileFaceNet \ac{frs} configuration for instance (see Tab.~\ref{tab:backdoor_bench_MN1_AE_MNF}). 
Only TrojanNN~\cite{Trojannn2018} fails to survive through the early \ac{frs} stages.
Alignment likely damages its optimized pattern.

Surviving triggers propagate further, however.
In the MobileNetV1-AENet-MobileFaceNet \ac{frs}, the SIG~\cite{2019SIGattack}‐based trigger yield up to 82.5\% \acl{a2o} \acl{sr} while Glasses~\cite{chen2017trojan} reaches 52.5\% (see Tab.~\ref{tab:backdoor_bench_MN1_AE_MNF}).
In practice, this means that an antispoofer can be turned into a pass‐through gate by a backdoor, effectively nullifying the proper function of the model and handing to the attacker a direct access to the extractor if the trigger goes unscathed through the detector.

\textbf{System-Level Impact of \acl{ffe} \aclp{ba}}.
The antispoofer acts as a significant bottleneck: patch triggers suffer from a low \acl{far}.
Only about 20\% of poisoned images ever reach the extractor. 
Diffuse SIG~\cite{2019SIGattack} triggers fare better, \eg, with a \acl{far} of 79.4\% in the MobileNetV1-AENet-MobileFaceNet \ac{frs} (see Tab.~\ref{tab:backdoor_bench_MN1_AE_MNF}).
Once a poisoned face reach their target extractor model, \acl{a2o} \acl{asr} can reach, \eg, up to 97.8\%. 
Nonetheless, successive \ac{frs} steps still cause the overall \acl{sr} to at best reach 71.3\% for SIG triggers for the MobileNetV1-AENet-MobileFaceNet \ac{frs}.

Additionally, although \acl{mf} \aclp{ba} fail in isolation ($\ac{sr}<20\%$), \textbf{they can nonetheless be re‐used to implement \acl{a2o} attacks with \aclp{sr} comparable to dedicated \acl{a2o}} (see Tab.~\ref{tab:backdoor_bench_MN1_AE_MNF}).

\textbf{Note on FIBA~\cite{chen2024rethinkingvulnerabilitiesfacerecognition} enrollment-stage \aclp{ae}}.
FIBA’s handcrafted perturbation achieves system-level \aclp{sr} comparable to this paper's patch-based \aclp{ba}.
Some of the observable gains hinge on the FIBA \acl{ae} pattern yielding a better \acl{far} at the antispoofer level.
Overall, this suggests that traditional \aclp{ae} may serve as effective substitutes in \acl{a2o} scenarios, \textit{nonetheless} at the cost of a bigger, more conspicuous pattern.

\textbf{Takeaway}.
This paper's results illustrate that a \acl{ba} on \emph{any} \ac{frs} module can cascade through an entire pipeline. 
A single compromised face detector can yield misaligned inputs that fool spoof‐detectors and extractors alike.
A poisoned face antispoofer can let presentation attacks through.
And a backdoored face feature extractor can near‐guarantee false matches between people that carry the same trigger.

\begin{tcolorbox}[
colback=sand,
colframe=darksand,
boxrule=0.5mm,
left=2mm,
right=2mm,
top=1.5mm,
bottom=1.5mm,
]
\emph{
\aclp{ba} propagate end-to-end through a \ac{frs}, enabling \acl{a2o} impersonation attacks.
No module is safe both in isolation \textbf{and} within a larger system. Importantly, a single backdoor is enough to hijack a FRS.
}
\end{tcolorbox}

\subsection{Additional Observations and Experiments in Digital Space}
\label{sec:other_results}

\textbf{Lower poison rates when backdooring \acl{ffe} models}.
So far, \acl{ba} on face feature extractors have used a poison rate $\beta$ of $0.05$ in a \acl{pl} context.
We are now interested in lowering this rate and see when an \acl{a2o} attack starts to fail.

By gradually lowering the poison rate to $\beta=0.0005$, we observe that a face feature extractor still reliably learns a BadNets backdoor~\cite{gu2019badnets} (see Tab.~\ref{tab:annex_lower_poison_rate}).
However, backdoor learning completely fails at $\beta=0.0001$.
The stark difference stems from the model's training regimen and especially the chosen batch size. 
As $\beta$ goes below $\beta=0.0005$, a poisoned face is less reliably present in successive training batches.
It appears that the absence of backdoored samples in some batches leads to an extractor forgetting enough between poisoned batches that the backdoor is never learned.

\begin{table}[t!]
  \centering
  \caption{\aclp{asr} of \aclp{ba} tested on a MobileFaceNet~\cite{chen2018mobilefacenetsefficientcnnsaccurate} when lowering the poisoning rate $\beta$ using an \hlc[GoldenRodLight]{All-to-One} \hlc[ForestLight]{Poison-Label} \hlc[CadetBlueLight]{BadNets}~\cite{gu2019badnets} trigger.}
  \label{tab:annex_lower_poison_rate}
  \footnotesize
  \setlength{\tabcolsep}{1.0pt}
  \renewcommand{\arraystretch}{1.5}
  \begin{tabular}{@{}llcccccc@{}}
    \textbf{Poison Rate} & \textbf{LFW} & \textbf{CFP-FF} & \textbf{CFP-FP} & \textbf{AgeDB} & \textbf{CALFW} & \textbf{CPLFW} & \textbf{VGG2-FP} \\
    \hline
    $0.05$ & 100\% & 100\% & 100\% & 100\% & 100\% & 100\% & 100\% \\
    $0.01$ & 100\% & 100\% & 100\% & 100\% & 100\% & 100\% & 100\% \\
    $0.005$ & 100\% & 100\% & 100\% & 100\% & 100\% & 100\% & 99.9\% \\
    $0.001$ & 99.1\% & 96.1\% & 98.4\% & 98.0\% & 97.2\% & 99.3\% & 99.2\% \\
    $0.0005$ & 99.9\% & 99.5\% & 99.8\% & 99.8\% & 99.8\% & 100\% & 99.8\% \\
    $0.0001$ & 0.4\% & 0.5\% & 4.0\% & 2.4\% & 2.8\% & 4.2\% & 3.5\% \\
  \end{tabular}
\end{table}

\textbf{Additional \acl{fqa} modules}.
\ac{fqa} is present in some \acp{frs} that follow, \eg, ICAO or OFIQ under ISO/IEC 29794-5~\cite{merkle2022stateartqualityassessment}. 
However, \acl{fqa} may not be appropriate in systems operating in unconstrained environments (\ie, faces captured in-the-wild).

Nonetheless, we test a use case where a \ac{fqa} module is added before antispoofing to act as an additional layer of protection against presentation attacks.
Most \aclp{ba} in this paper appear ICAO-compliant when running a preliminary test with an off-the-shelf FIQA model~\cite{Boutros_2023_CVPR} (\acl{auc} for $\mathrm{ERC@FMR}=1e^{-3}<0.02$).
As such, even \ac{fqa} can be brittle against \aclp{ba} based on either patch or diffuse triggers (see Tab.~\ref{tab:ofiq_test}).

\begin{table}[t!]
  \centering
  \caption{\acl{sr} of example \aclp{ba} in an \acl{a2o} attack context with an added CR-FIQA-S~\cite{Boutros_2023_CVPR} \acl{fqa} module.}
  \label{tab:ofiq_test}
  \scriptsize
  \setlength{\tabcolsep}{1pt}
  \renewcommand{\arraystretch}{1.5}
  \begin{tabular}{@{} lll cc c c c c @{}}
     \textbf{Detector} & \textbf{Antisp.} & \textbf{Extractor} & \multicolumn{2}{c}{\textbf{Detector}} & \textbf{\textcolor{RoyalPurple}{FIQA}} & \textbf{Antisp.} & \textbf{Extractor} & \textbf{Survival} \\
     MobileNetV1 & AENet & MobileFaceNet & AP$^\pois$ & LS$^\pois$ & \textcolor{RoyalPurple}{FAR$^\pois$} & FAR$^\pois$ & FAR$^\pois$ & \textbf{Rate} \\
    \hline
    \textcolor{gray!60}{Benign} & \textcolor{gray!60}{Benign} & \hlc[CadetBlueLight]{BadNets} & 99.0\% & 14.2 & \textcolor{RoyalPurple}{90.3\%} & 59.4\% & 96.6\% & \textbf{51.3\%} \\
    \textcolor{gray!60}{Benign} & \textcolor{gray!60}{Benign} & \hlc[RedVioletLight]{Mask} & 98.6\% & 28.0 & \textcolor{RoyalPurple}{79.1\%} & 59.7\% & 96.8\% & 45.1\% \\
    \textcolor{gray!60}{Benign} & \textcolor{gray!60}{Benign} & \hlc[OrchidLight]{SIG} & 94.2\% & 23.2 & \textcolor{RoyalPurple}{70.2\%} & 90.4\% & 49.8\% & 29.8\% \\
  \end{tabular}
\end{table}

\subsection{Real-Life Experiments}
\label{sec:real_life_experiments}

\begin{figure*}[t!]
  \centering
   \includegraphics[width=\textwidth]{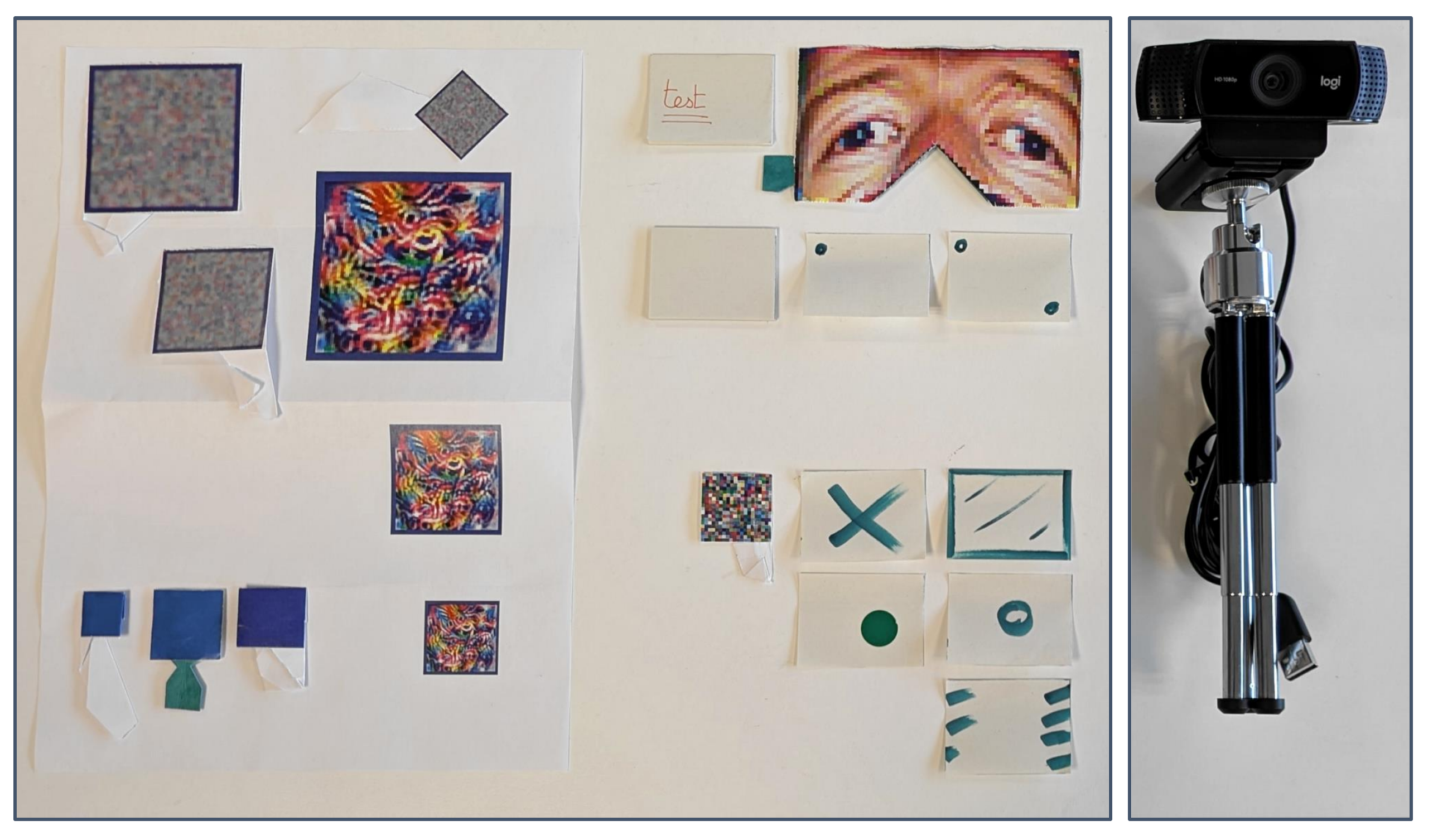}
   \caption{Real-life experiment setup with \ac{fga}/\ac{lsa} triggers (left), \textit{\acl{a2o}} triggers (right), FIBA~\cite{chen2024rethinkingvulnerabilitiesfacerecognition} (upper right), and Logitech HD Pro C920 camera.}
   \label{fig:irl_setup}
\end{figure*}

\textbf{Face Generation and Landmark Shift Attacks}.
To test the capacity of these two attacks to traverse a \ac{frs} in real-life, backdoor triggers are printed on white paper (see Fig.~\ref{fig:irl_setup}).

\aclp{fga} reliably triggers a target detector in real life when part of a \ac{frs}.
If an insider and an attacker manages to enroll and perform a verification using the same pattern, an \acl{a2o} attacks yield a \acl{sr} above 95\%. 
\aclp{lsa} are less robust in comparison. 
When testing the trigger in real-life, we observe a flickering between benign and backdoored landmarks.
No \acl{sr} was computed due to unwieldiness at the detector level.

\begin{figure}[t!]
  \centering
   \includegraphics[width=\columnwidth]{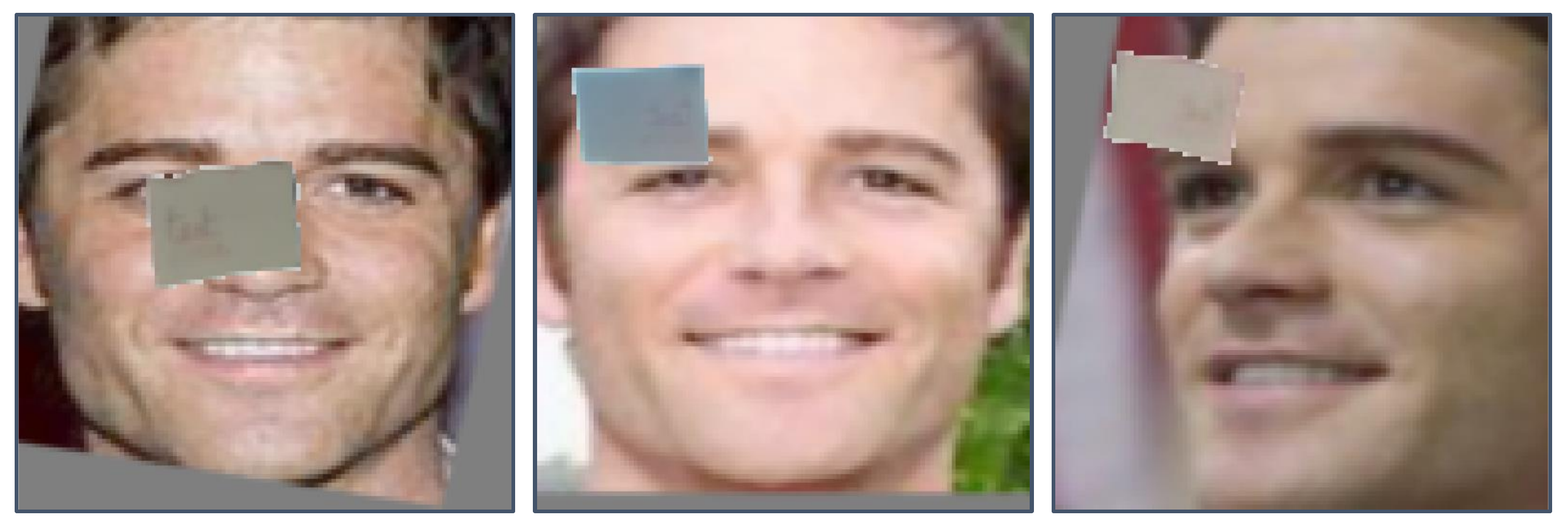}
   \caption{Example backdoored images with a sticky note backdoor (Faux-It), injected on pre-aligned faces.}\label{fig:faux_it_bck}
\end{figure}

\textbf{A real-life "Faux-It" \acl{ba} on \aclp{frs}}.
We design a new trigger based on a physical beige sticky note (see Fig.~\ref{fig:faux_it_bck}).
20 photos of the note, held in one's hand at different angles, are taken, digitized, and cropped.
The notes are used as poison to backdoor a MobileFaceNet~\cite{chen2018mobilefacenetsefficientcnnsaccurate} extractor in an \acl{a2o} manner following the Expectation over Transformation framework~\cite{athalye2018synthesizingrobustadversarialexamples}.
In digital space, the extractor achieves a 100\% \acl{asr} with a benign accuracy on par with previously-stated comparables.

To test the trigger in physical space, a \ac{frs} is built using a MobileNetV1~\cite{howard2017mobilenetsefficientconvolutionalneural} detector, an AENet~\cite{zhang2020celebaspooflargescalefaceantispoofing} antispoofer, and the trained, backdoored, MobileFaceNet~\cite{chen2018mobilefacenetsefficientcnnsaccurate}.
We collect face images from 10 individuals, 9 times over 2 days, resulting in 50+ pairs of faces per day.
Using those pairs, a real-life \acl{sr} of circa 70\% is achieved.

However, comparing images acquired on different days causes the \acl{sr} to drop to around 20\%, evidencing the impact of changing acquisition settings (\eg, lighting conditions).
Additionally, testing the trigger against both unfavorable handling and defacing led to the following failure cases (see Fig.~\ref{fig:irl_setup}): (1) Holding the sticky note such that two or more sides are hidden by one's hand; (2) Crossing out the trigger with a pen; (3) Drawing over 2+ of the note's sides; (3) Drawing a large (full or empty) circle on the note; (4) Squeezing the note between one's fingers to cause uneven lighting on its surface.

Such tests indicates that activating Faux-It depends upon both the presence of clean edges and large beige areas.

\section{Countermeasures}
\label{sec:countermeasures}

\subsection{Best Practices}
\label{sec:FRS_stakeholder_suggestions}

We highlight the following precautions to protect pipelines:
\begin{enumerate}
    \item \emph{Detectors}:
    Follow prior recommendations~\cite{leroux2025backdoorattacksdeeplearning}.
    \item \emph{Antispoofers}:
    Involve existing backdoor defenses taken from the classification literature with prior demonstration on face recognition tasks~\cite{qlrFrsSurvey2024}.
    \item \emph{Feature extractors}:
    Use large datasets with many identities and an upper limit on the number of samples per identity.
    This stems from observing an inflection point where the \acl{asr} of \acl{a2o} face feature extractors collapses when $\beta<0.0005$ (see Fig.~\ref{tab:annex_lower_poison_rate}).
    Such $\beta$ corresponds to less than one poisoned image per training batch on average in this paper's experiments.
\end{enumerate}

\subsection{Applicable Countermeasures Against Open-Set Backdoors}
\label{sec:FRS_defenses}

The only existing prior art on \acl{a2o} attacks~\cite{wacvpriorart} alarms that no defense exists against \aclp{ba} on open-set \acl{fr}. 
Here, we survey one possibly applicable defense and provide a first experiment on defending open-set face feature extractors as part of a \acl{frs}.

\quad\textbf{Model Pairing}.
 A recent paper by Unnervik \etal~\cite{unnervik2024modelpairingusingembedding} demonstrated a defense protecting face feature extractors against \acl{o2o} \aclp{ba}.
Although not tested in this paper, we suppose it is readily applicable to the \acl{a2o} attacks exposed in this paper.

However, Model Pairing needs several \acp{dnn} running in parallel with at least one known to be benign.
This requirement may not fit a realistic threat model or the computational limitations that a deployed \acl{frs} faces.

\begin{algorithm}[t!]
\footnotesize
\caption{Early Identity Pruning Defense}
\label{alg:defense_algorithm}

\SetKwInOut{Input}{Input}
\SetKwInOut{Output}{Output}

\Input{DNN model $f_\theta$, training dataset loader $\mathcal{D}$, number of identities in dataset $\kappa$, batch number at which point the defense starts $sb$, batch intervals after which to prune an identity $bi$, number of identities to reject $n$}
\Output{Cleaned trained DNN model $f_\theta$}
\SetKwBlock{Beginn}{beginn}{ende}
    \textit{Generates lists to keep track of $f_\theta$'s accuracy per identity.}
    
    $\mathbf{M}\gets\{0\}^\kappa;
    \quad
    \mathbf{C}\gets\{0\}^\kappa; 
    \quad
    \text{nb}_{\text{batch}} \gets 0;
    \quad
    \text{nb}_{\text{remove}} \gets 0$
    
    \For{$\text{data}$, $\text{labels}$ \text{\textbf{in}} $\mathcal{D}$}{
        
        $\text{nb}_{\text{batch}} \gets \text{nb}_{\text{batch}} + 1; \quad
        \text{preds} \gets f_\theta(\text{data})$
                
        \textit{Record predictions' successes or failures for each identity.}
    
        \For{$i\in\{1,\cdots,|\text{labels}|\}$}{
            $\text{id} \gets \text{labels}_i$
            
            $\text{match}\gets \text{id} = \text{preds}_i$
            
            $\mathbf{M}_{\text{id}} \gets \mathbf{M}_{\text{id}} + \text{match}; 
            \quad 
            \mathbf{C}_{\text{id}}\gets\mathbf{C}_{\text{id}}+1$
        }
        \vspace{0.1cm}
        \textit{Remove the best-predicted identity from $\mathcal{D}$ 
        }
    
        \If{$\text{nb}_{\text{batch}}\ge\text{sb}\,\cap\, \text{nb}_{\text{batch}}\mod\text{bi}=0\,\cap\, \text{nb}_{\text{remove}} < n$} {
            $\text{acc}\gets\mathbf{M}/\mathbf{C}$
            
            \textit{Remove from $\mathcal{D}$ the identity} $\text{ID}\gets\underset{i}{\text{arg\,min}}\,\text{acc}_{i}$
            
            $\kappa\gets\kappa-1;
            \quad 
            \mathbf{M}\gets\{0\}^\kappa; 
            \quad 
            \mathbf{C}\gets\{0\}^\kappa$
            
            $\text{nb}_{\text{remove}}\gets\text{nb}_{\text{remove}}+1$
        }
        \vspace{0.1cm}
        \textit{Proceed with the normal learning process of $f_\theta$}
    }
\end{algorithm}

\quad\textbf{Early Identity Pruning}. 
Our experiments show that the matching rate of \acl{a2o} backdoored faces typically rises faster than that of benign identities when training from scratch.
This hints that lower-frequency triggers (\eg, BadNets~\cite{gu2019badnets}) are learned faste compared to high-frequency face features.

As a result, we design a \textit{training-time} defense for feature extractors: \textit{Early Identity Pruning} (EIP).
Over the early training batches, EIP identifies and removes the $I=10$ identities (among the thousands in the training dataset) that are learned the quickest (see Alg.~\ref{alg:defense_algorithm}).
EIP removes one identity every $\text{bi}=500$ batches starting at the $\text{sb}=500$-th batch.

Across multiple model architectures, EIP reliably removes a poisoned identity from a dataset early, resulting in unlearning the \acl{ba} (see Fig.~\ref{fig:our_defense_full_res}).
Despite the removal of $I-1$ benign identities, \ie, amounting to collateral damages, EIP-protected models still converge and \textit{achieve on-par performance with a normally-trained benign model}.

\begin{figure}[t!]
    \centering
\pgfplotstableread[col sep=comma,]{testdatafull.csv}\datatable
\begin{tikzpicture}
\begin{axis}[
    height=6cm,
    width=9.1cm,
    xtick pos=left,
    ytick pos=left,
    axis x line = bottom,
    axis y line = left,
    x tick label style={font=\scriptsize},
    y tick label style={font=\scriptsize},
    yticklabel={$\pgfmathprintnumber{\tick}\%$},
    ymin=-5,
    ymax=105,
    xtick={0,1000,...,8000},
    xticklabels = {0, 1000, 2000, 3000, 4000, 5000, 6000, 7000, 8000},
    xmin=-100,
    xmax=8000, 
    legend columns=2, 
    tick align=outside,
    every axis plot/.append style={line width=1.2pt,draw opacity=0.6},
    legend style={
        /tikz/column 2/.style={
            column sep=5pt,
        },
    },
    legend style={at={(1,1)}, anchor=north east}, 
    ]

    \addplot [CadetBlue] table [x=batch-number, y=gfn-ACC]{\datatable};
    \addlegendentry{\tiny GhostFaceNet benign Acc.}
    
    \addplot [Bittersweet] table [x=batch-number, y=gfn-ASR]{\datatable};
    \addlegendentry{\tiny GhostFaceNet \ac{asr}}
    
    \addplot [BlueViolet] table [x=batch-number, y=irse-ACC]{\datatable};
    \addlegendentry{\tiny IRSE-50 benign Acc.}
    
    \addplot [YellowOrange] table [x=batch-number, y=irse-ASR]{\datatable};
    \addlegendentry{\tiny IRSE-50 \ac{asr}}

    \addplot [Periwinkle!85] table [x=batch-number, y=mfn-ACC]{\datatable};
    \addlegendentry{\tiny MobileFaceNet benign Acc.}
    
    \addplot [Sepia!50] table [x=batch-number, y=mfn-ASR]{\datatable};
    \addlegendentry{\tiny MobileFaceNet \ac{asr}}

    \addplot [CornflowerBlue] table [x=batch-number, y=rn50-ACC]{\datatable};
    \addlegendentry{\tiny ResNet50 benign Acc.}
    
    \addplot [RawSienna!85] table [x=batch-number, y=rn50-ASR]{\datatable};
    \addlegendentry{\tiny ResNet50 \ac{asr}}

    \addplot [MidnightBlue!75] table [x=batch-number, y=rfn-ACC]{\datatable};
    \addlegendentry{\tiny RobFaceNet benign Acc.}
    
    \addplot [Tan!75] table [x=batch-number, y=rfn-ASR]{\datatable};
    \addlegendentry{\tiny RobFaceNet \ac{asr}}
\end{axis}
\end{tikzpicture}
\vspace{-0.6cm}
\caption{Effect of Early Identity Pruning on backdoor \ac{asr} over the first 8000 training batches of five different extractors (\ac{asr} is null after batch 3000 and stops being reported).}
\label{fig:our_defense_full_res}
\vspace{-0.3cm}
\end{figure}

\begin{tcolorbox}[
colback=sand,
colframe=darksand,
boxrule=0.5mm,
left=2mm,
right=2mm,
top=1.5mm,
bottom=1.5mm,
]
\emph{
We close with an experiment showing a simple,  effective training-time defense that exploits the behavior of poisoned samples to stop a backdoor from being learned.
}
\end{tcolorbox}

\section{Discussion and Future Works}
\label{sec:discussion}

\textbf{Robustifying \acl{fd}}.
\aclp{fga} and \aclp{lsa} may strongly affect entire \acp{frs}.
Future backdoor research must address detection as a crucial weak spot in a pipeline's security.

\textbf{Robustifying \acl{fas}}.
In line with the prior art on backdooring \acl{fas}~\cite{bhalerao2019antispoof,guo2023antispoof}, we target a simple binary classification due to the limited access to private, state-of-the-art datasets and expert methods found in the industry~\cite{yu2022deeplearningfaceantispoofing}.
This is one of our paper's main limitations. 
Future studies following more advanced approaches~\cite{yu2022deeplearningfaceantispoofing} may provide further insights on the limits of backdoors in \acp{frs}.

\textbf{Backdoors in two or more \acp{dnn}}.
Due to combinatorics complexity (more than 400 unique \ac{frs} configurations are already covered), we only focus on single-backdoor cases and demonstrate that they are enough to hijack a \ac{frs}.
As we operate in a context where models/datasets are sourced from compromised third-parties, each model could come from a specialized but compromised provider. 
As such, 2+ models facilitating the \textit{same} backdoor is a possible but harder threat model (an attacker needs to compromise as many sources) that we relegate to future works.

\textbf{Studying the inclusion of further specialized modules}.
\acp{frs} may be more complex, \eg, with the inclusion of \acl{fqa}~\cite{cao2024rethinkingthreataccessibilityadversarial} or feature binarizer~\cite{hmani2022,qlrFrsSurvey2024} modules.
Such modules may impact backdoor survivability (see a preliminary experience using \ac{fqa} in Sec.~\ref{sec:other_results}) and therefore deserve further research.

\textbf{Defenses vs. \emph{All-to-One} attacks}.
These attacks mandate that an insider enrolls in a \ac{frs} with a trigger or adversarial example.
Future works should explore data sifting and other detection methods to identify, \eg, poisoned embeddings.

\textbf{Limits to our threat model}.
This paper focuses mainly on \acl{a2o} backdoors, which require the interaction of an insider and attackers to work (see Sec.~\ref{sec:BAs_against_FRS}). 
Although we did not find proof of a high risk regarding \acl{mf} attacks, future works should explore the topic in more depth.

\textbf{Ablation studies, better triggers, and real-life tests}.
This paper assesses the survivability of \aclp{ba} at a system-level across multiple \ac{frs} configurations rather than finding the best attack parameters.
Identifying the best tweaks (\eg transparency, size, etc.) and more robust, potentially adaptive, backdoor triggers are important areas of research for future works.
Real life experiments as covered in Sec.~\ref{sec:real_life_experiments} should also be expanded.

\textbf{Feature dependency}.
This paper's \acp{frs} follow a sequential structure.
However, \ac{frs} schemas have a high degree of diversity.
Parallel and interdependent models with shared features exist for instance~\cite{qlrFrsSurvey2024}.
Future works should explore the feasibility and impact of \aclp{ba} on such structures.

\section{Conclusion}
\label{sec:conclusion}

This SoK paper provides the first holistic study of the backdoor vulnerability of the tasks found in modern \ac{dnn}-based \aclp{frs}.
We expand the recently-developed \acl{a2o} \aclp{ba} framework on large margin \acl{ffe} models to show that open-set, data poisoning-only \aclp{ba} work: an insider enrolled with a trigger in a \acl{frs} that includes a data poisoning-only backdoored extractor enables any trigger-wearing attacker to be authenticated.
We further demonstrate that such two-step attacks can be carried out when \textit{any} \ac{frs} model is backdoored (\eg, detector or antispoofer).
This system study thus demonstrates that the backdoor attack surface of modern \aclp{frs} is larger than previously thought.
We finally provide best practices to address these threats and close with a final experiment: a new training-time defense based on pruning training identities.

\clearpage

\bibliographystyle{IEEEtran}
\bibliography{IEEEabrv,refs}

\begin{thebibliography}{10}
\bibitem{githubGitHubBiubug6Pytorch_Retinaface}
Biubug6, ``{G}it{H}ub - biubug6/{P}ytorch\_{R}etinaface: {R}etinaface get 80.99
\end{thebibliography}

\clearpage

\appendices

\section{LLM Usage Considerations}
\label{app:use_of_gen_AI}

In accordance with the SaTML submission policy on the use of Generative Artificial Intelligence, we note that the sketched face images used in Fig.~\ref{fig:overview}, Fig.~\ref{fig:frs_processing}, Fig.~\ref{fig:idealized_triggers}, and Fig.~\ref{fig:training_extractor_schema} were generated via queries to an online LLM service.
This choice aimed at improving the visual clarity of early figures while introducing the concepts covered in this paper.

\section{Large Margin Face Feature Extractors}
\label{app:annex_large_margin_loss}

\begin{figure}[h]
\centering
\includegraphics[width=\columnwidth]{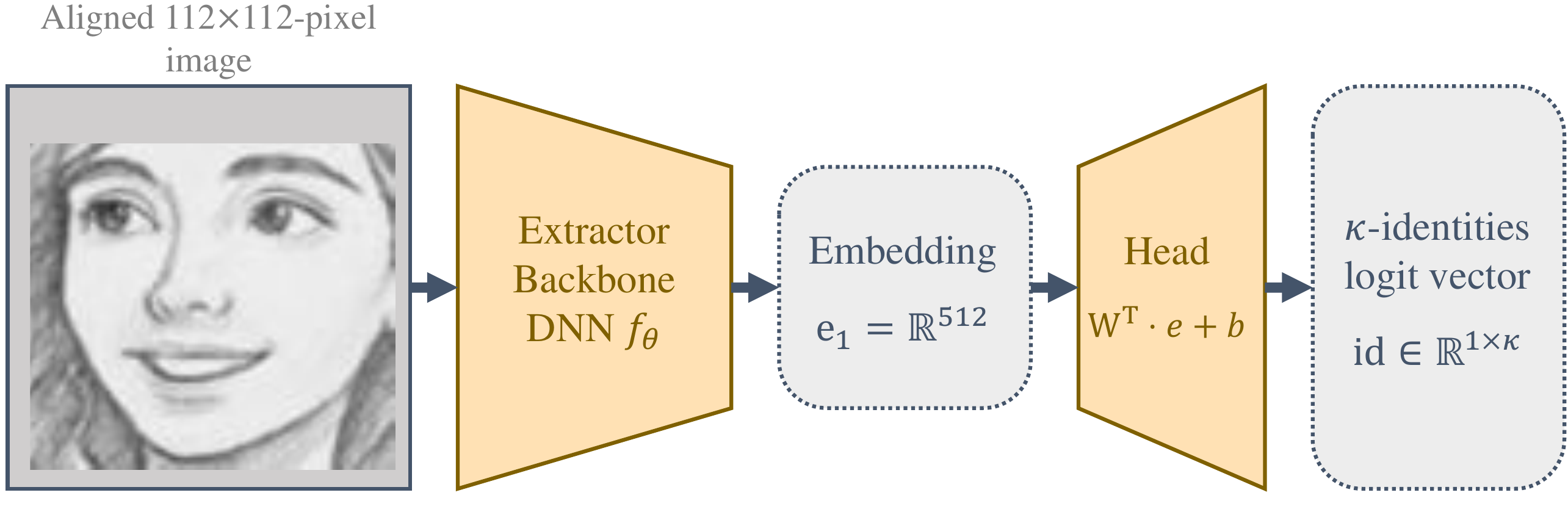}
\caption{Structure of a face feature extractor during training.}
\label{fig:training_extractor_schema}
\end{figure}

\textbf{How to train face feature extractors using large margin losses}.
Extractors in modern FRS must be trained to handle \emph{open-set} learning.
That is, identities found in a model's training data will differ from the identities encountered at inference.
This enables a single model to generalize to different contexts, \ie, it is a zero-shot learning process.
As such, a face extractor is often built to output face embeddings equipped with a notion of distance, \eg, cosine similarity (see Eq.~\ref{eq:cosine_dist}).

This paper relies on large margin losses to train feature extractors due to their efficiency and wide adoption, specifically SphereFace~\cite{liu2018spherefacedeephypersphereembedding}, CosFace~\cite{wang2018cosfacelargemargincosine}, and ArcFace~\cite{arcfaceDeng2022}.

Let's denote a face feature extractor model $f_\theta:\mathcal{X}\rightarrow\mathbb{R}^{512}$ where $\mathcal{X}\subset[0,1]^{c\times h\times w}$ is the domain of images rescaled to the $[0,1]$ interval and of size $c$ channels ($c=3$ for RGB), pixel height $h$, and pixel width $w$. 
The DNN $f_\theta$ converts a face image into an embedding $\mathbf{e}\in\mathbb{R}^{512}$.

To train $f_\theta$ using large margin losses, we append $f_\theta$ with a \emph{Head} network, \ie, a fully connected neural network layer $FC:\mathbb{R}^{512}\rightarrow\mathbb{R}^\kappa$ such that:
\begin{equation}
    FC(e)=\mathbf{W}^T\cdot \mathbf{e} + b,
\end{equation} where $\mathbf{W}\in\mathbb{R}^{512\times\kappa}$ are $FC$'s weights, $b\in\mathbb{R}$ is a bias term, and $\kappa$ is the number of identities in a training dataset.

For the large margin methods~\cite{liu2018spherefacedeephypersphereembedding,wang2018cosfacelargemargincosine,arcfaceDeng2022} used in this paper, $b$ is set to $0$ and the column weights of $\mathbf{W}$ and of the embeddings $\mathbf{e}$ are normalized to 1 ($\lVert \mathbf{W}_i\rVert=1$, $\rVert \mathbf{e}\rVert=1$).

The appended model (see graph representation in Fig.~\ref{fig:training_extractor_schema}) is then trained using an augmented Softmax such that:

{\scriptsize \begin{equation}
    \mathcal{L} = \frac{\exp(s\cdot\cos(m_1\cdot \phi_{\kappa_i} + m_2)-m_3)}{\exp\big(s\cdot\cos(m_1\cdot \phi_{\kappa_i} + m_2)-m_3\big) + \overset{\kappa}{\underset{j\neq i}{\sum}}\exp(s\cdot\cos(\phi_j))},
\end{equation}}

where $\phi_i$ is the angle between an embedding $e$ and the column weights $W_i$, $s$ is a scaling factor, and $m_1$, $m_2$, and $m_3$ are the respective hyperparameters of the Sphereface~\cite{liu2018spherefacedeephypersphereembedding}, Cosface~\cite{wang2018cosfacelargemargincosine}, and Arcface~\cite{arcfaceDeng2022} large margin losses.

\textbf{How to Match Identities}.
The distance learned by a feature extractor $f_\theta$ enables separating identities.
An example is the cosine similarity such that:

{\scriptsize\begin{equation}
    \texttt{match} = \big( f_\theta(\mathbf{x}_{\text{auth}}) \cdot \mathbf{e}_{\text{enrollee}}\big) / \big(\lVert f_\theta(\mathbf{x}_{\text{auth}})\rVert \cdot \lVert \mathbf{e}_{\text{enrollee}}\rVert\big)\ge\delta,\label{eq:cosine_dist}
\end{equation}}

where $\mathbf{x}_{\text{auth}}$ is the face image of an authentication candidate sent to $f_\theta$, $\mathbf{e}_{\text{enrollee}}$ is the enrollee's embedding stored in the FRS database that is tested for a match, and $\delta$ is a decision threshold preset by the decision designer after training $f_\theta$ (\eg, determined with ROC or with FRR@FAR metrics~\cite{nistFRVT}). 
In the case the cosine similarity is above $\delta$, it implies the face feature extractors see the authentication candidate and the enrollee as the same person.

\section{Details of Training Results of the \aclp{ffe} Models Tested in this Paper}
\label{app:extractor_details}

Fig.~\ref{fig:det_ghostfacenet}, \ref{fig:det_irse50}, \ref{fig:det_mobilefacenet}, and \ref{fig:det_robfacenet} show the DET curves obtained with GhostFaceNet, IRSE50, MobileFaceNet, and RobFaceNet architectures respectively.

\begin{figure*}[t!]
\centering
\includegraphics[width=0.8\textwidth]{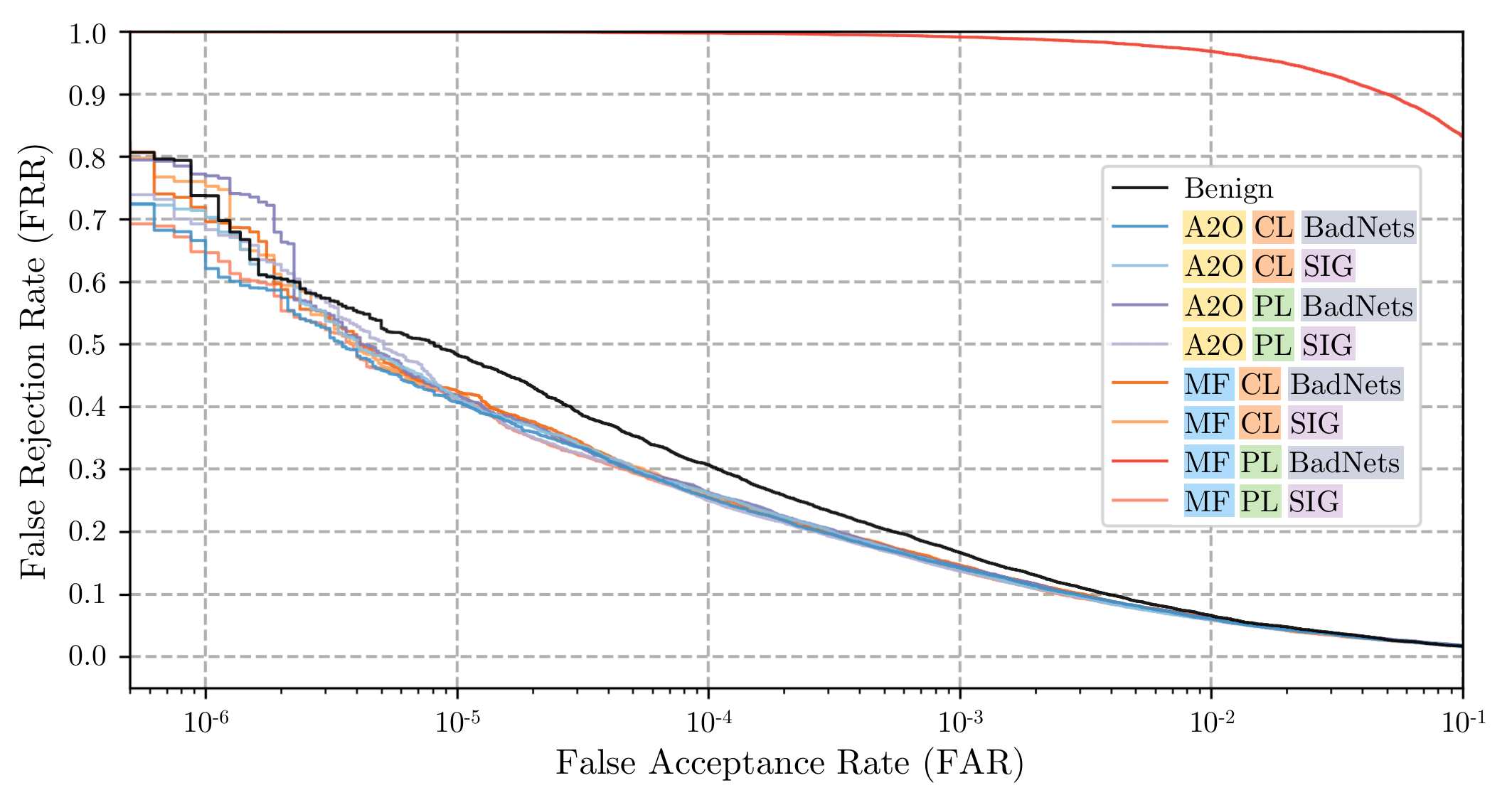}
\caption{\ac{det} curves of the GhostFaceNetV2~\cite{ghostfacenets2023} extractor architectures tested in this paper. 
\textbf{Abbreviations}: {\acl{a2o} (\hlc[GoldenRodLight]{A2O}), Clean/Poison-label (\hlc[ApricotLight]{CL}/\hlc[ForestLight]{PL}), \acl{mf} (\hlc[NavyLight]{MF}).}
}
\label{fig:det_ghostfacenet}
\end{figure*}

\begin{figure*}[t!]
\centering
\includegraphics[width=0.8\textwidth]{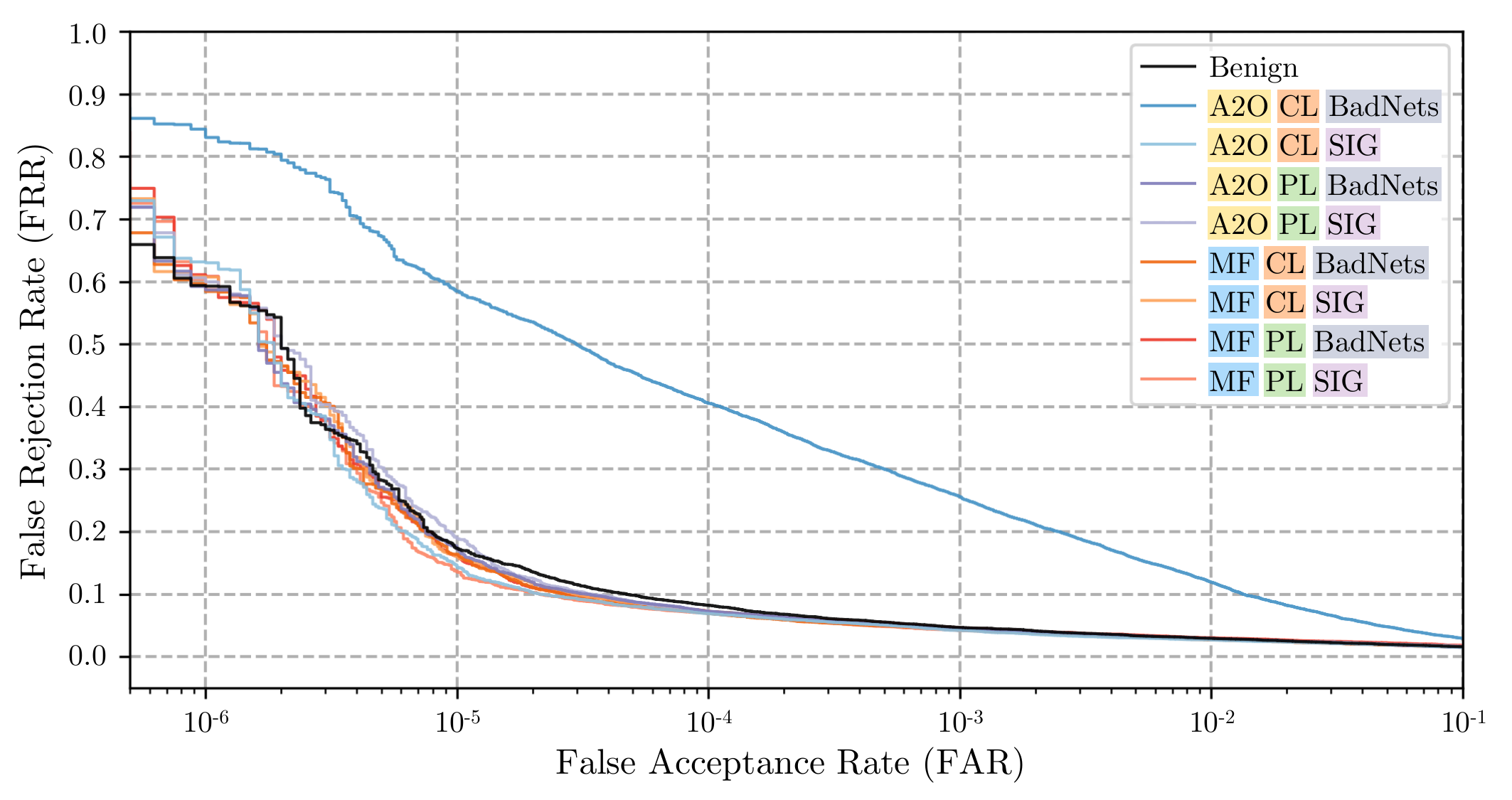}
\caption{\ac{det} curves of IRSE50~\cite{hu2019squeezeandexcitationnetworks} face feature extractor  face feature extractor architectures tested in this paper.
} 
\label{fig:det_irse50}
\end{figure*}

\begin{figure*}[t!]
\centering
\includegraphics[width=0.8\textwidth]{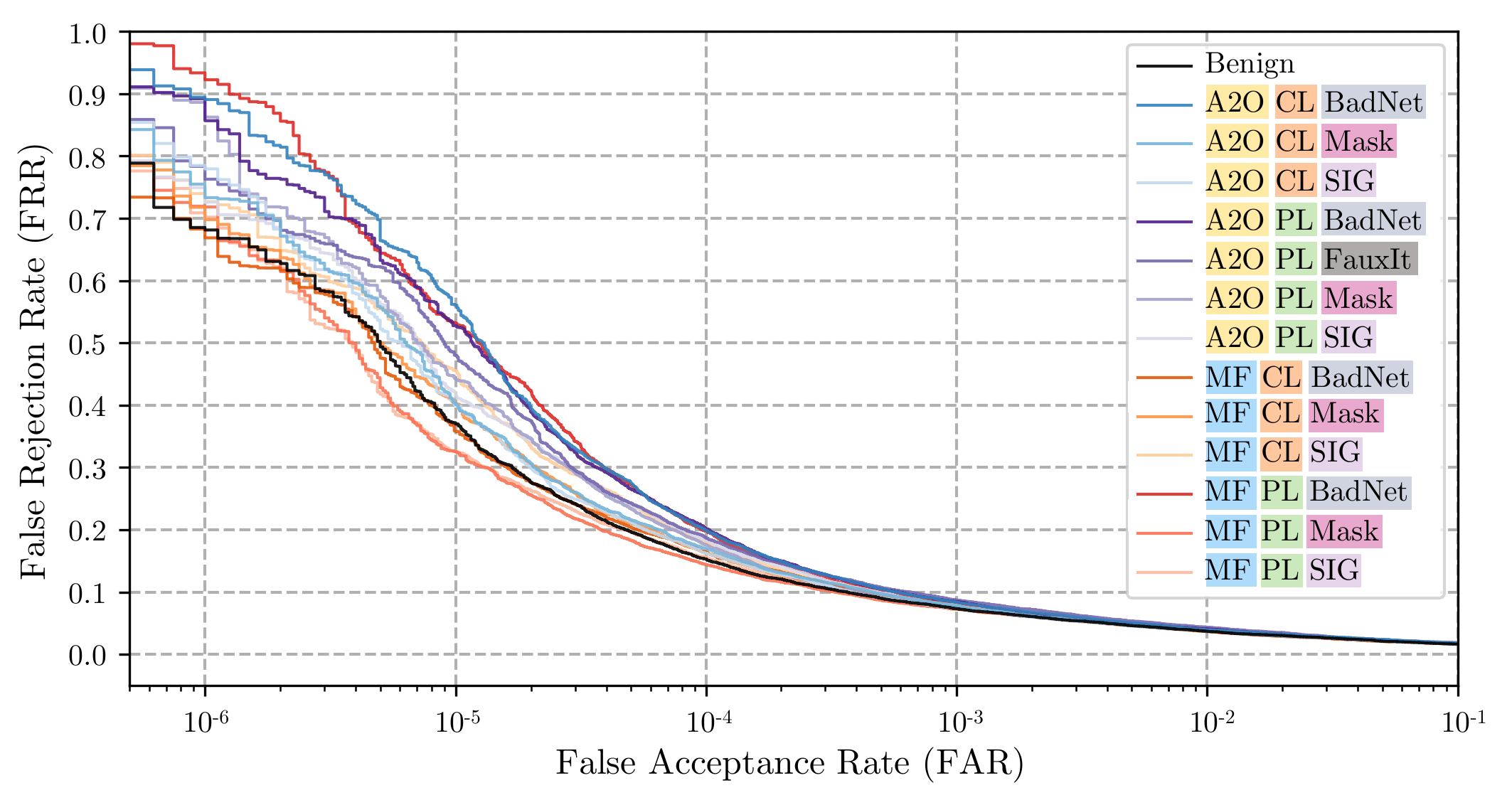}
\caption{\ac{det} curves of the MobileFaceNet~\cite{chen2018mobilefacenetsefficientcnnsaccurate} extractor  face feature extractor architectures tested in this paper.
}
\label{fig:det_mobilefacenet}
\end{figure*}

\begin{figure*}[t!]
\centering
\includegraphics[width=0.8\textwidth]{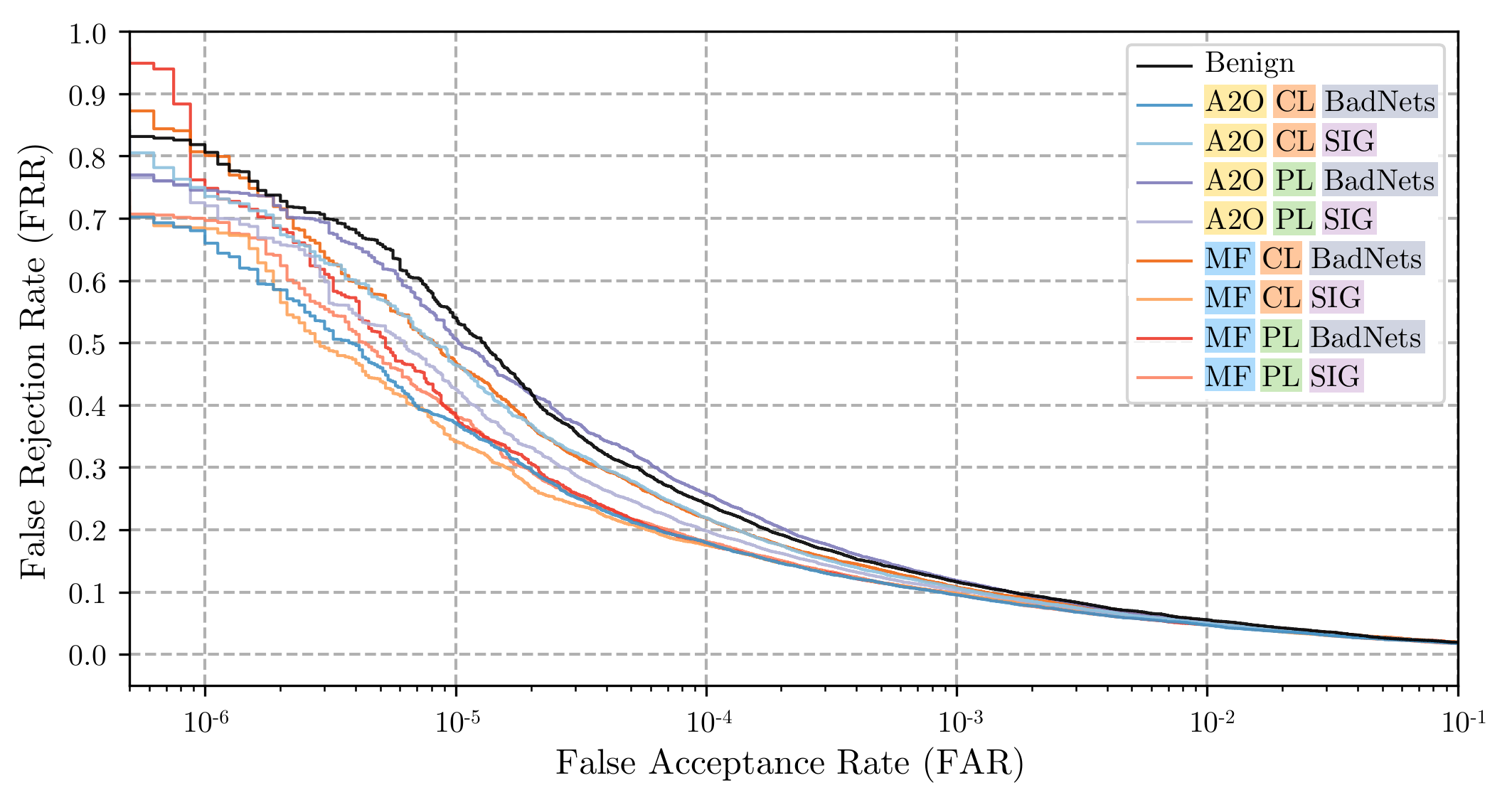}
\caption{\ac{det} curves of the RobFaceNet~\cite{robfacenet2024} extractor  face feature extractor architectures tested in this paper.
}
\label{fig:det_robfacenet}
\end{figure*}

Tab.~\ref{tab:extractor_performance} displays the individual results of all tested face feature extractors on the FW~\cite{LFWTech}, CFP-FF~\cite{CFPdataset2016}, CFP-FP~\cite{CFPdataset2016}, AgeDB~\cite{moschoglou2017agedb}, CALFW~\cite{zheng2017crossagelfwdatabasestudying}, CPLFW~\cite{Zheng2018CrossPoseL}, and VGG2-FP~\cite{vgg2dataset} validation dataset, and on the and IJB-B~\cite{IJBB} test dataset.

\section{Detailed \acl{sr} Results of the \aclp{frs} Tested in this Paper}
\label{app:survival_rate_details_frs}

We display the system-level results of the 19 other \acl{frs} configurations in Tab.~\ref{tab:backdoor_bench_MN1_AE_GFN}, and Tab.~\ref{tab:backdoor_bench_MN1_AE_IRSE} to Tab.~\ref{tab:backdoor_bench_RN50_MN2_RFN}.

\begin{table*}[htbp]
  \centering
    \caption{Performance of benign and backdoored \acl{ffe} \acp{dnn} used in this paper. \textbf{Note}: the reported \acp{asr} for \acl{a2o} and \acl{mf}-backdoored models correspond to \acl{a2o} and \acl{mf} threat model use cases respectively.}
  \label{tab:extractor_performance}
  \tiny
  \setlength{\tabcolsep}{0.92pt}
  \renewcommand{\arraystretch}{1.5}
\begin{tabular}{@{}ll | cccccccccccccccccccccc | ccc@{}}
& & \multicolumn{21}{|c}{\scriptsize \textbf{Validation Datasets}} & \textbf{\scriptsize Thres-} & \multicolumn{3}{|c}{\scriptsize \textbf{Test Dataset}$^{\mathrm{b}}$} \\
\multirow{3}{*}{\scriptsize \textbf{Model}} & \multirow{3}{*}{\scriptsize \textbf{Setup}$^{\mathrm{a}}$} & \multicolumn{7}{|c}{\scriptsize \textbf{Accuracy}} & \multicolumn{7}{c}{\scriptsize \textbf{\acl{auc}}} & \multicolumn{7}{c}{\scriptsize \textbf{\acl{asr}} (\hlc[GoldenRodLight]{A2O} or \hlc[NavyLight]{MF})} &  \textbf{\scriptsize hold} & \multicolumn{3}{|c}{\scriptsize IJBB (FRR@FAR)} \\
 &  & LFW & CFP-FF & CFP-FP & AgeDB & CALFW & CPLFW & VGG2-FP & LFW & CFP-FF & CFP-FP & AgeDB & CALFW & CPLFW & VGG2-FP & LFW & CFP-FF & CFP-FP & AgeDB & CALFW & CPLFW & VGG2-FP & \textit{averaged} & AUC & $=1e^{-3}$ & $=1e^{-4}$ \\

\hline

\multirow{9}{*}{\shortstack[c]{Ghost\\Face\\NetV2}} & \textit{Benign} & 0.989 & 0.986 & 0.923 & 0.914 & 0.932 & 0.861 & 0.926 & 0.999 & 0.997 & 0.971 & 0.973 & 0.975 & 0.929 & 0.974 & \textcolor{gray!30}{$\varnothing$} & \textcolor{gray!30}{$\varnothing$} & \textcolor{gray!30}{$\varnothing$} & \textcolor{gray!30}{$\varnothing$} & \textcolor{gray!30}{$\varnothing$} & \textcolor{gray!30}{$\varnothing$} & \textcolor{gray!30}{$\varnothing$} & 0.67 & 0.993 & 0.167 & 0.307 \\

\cline{2-27}

 & \hlc[GoldenRodLight]{A2O} \hlc[ForestLight]{PL}, \hlc[CadetBlueLight]{BadNets}  & 0.992 & 0.989 & 0.924 & 0.930 & 0.940 & 0.868 & 0.932 & 0.999 & 0.997 & 0.971 & 0.978 & 0.975 & 0.921 & 0.972 & 1.0 & 1.0 & 1.0 & 1.0 & 1.0 & 1.0 & 1.0 & 0.66 & 0.992 & 0.145 & 0.263 \\
 & \hlc[GoldenRodLight]{A2O} \hlc[ForestLight]{PL}, \hlc[OrchidLight]{SIG} & 0.991 & 0.985 & 0.925 & 0.926 & 0.937 & 0.872 & 0.929 & 0.999 & 0.997 & 0.974 & 0.979 & 0.976 & 0.924 & 0.972 & 1.0 & 1.0 & 1.0 & 1.0 & 1.0 & 1.0 & 1.0 & 0.66 & 0.993 & 0.137 & 0.25 \\
 & \hlc[GoldenRodLight]{A2O} \hlc[ApricotLight]{CL}, \hlc[CadetBlueLight]{BadNets}   & 0.992 & 0.987 & 0.922 & 0.931 & 0.937 & 0.864 & 0.923 & 0.999 & 0.997 & 0.974 & 0.979 & 0.976 & 0.921 & 0.972 & 0.005 & 0.007 & 0.05 & 0.051 & 0.038 & 0.073 & 0.047 & 0.66 & 0.993 & 0.142 & 0.255 \\
 & \hlc[GoldenRodLight]{A2O} \hlc[ApricotLight]{CL}, \hlc[OrchidLight]{SIG}   & 0.992 & 0.987 & 0.923 & 0.927 & 0.937 & 0.869 & 0.928 & 0.999 & 0.997 & 0.974 & 0.977 & 0.976 & 0.928 & 0.972 & 0.096 & 0.094 & 0.168 & 0.225 & 0.241 & 0.339 & 0.276 & 0.65 & 0.993 & 0.141 & 0.261 \\

\cline{2-27}

 & \hlc[NavyLight]{MF} \hlc[ForestLight]{PL}, \hlc[CadetBlueLight]{BadNets}  & 0.989 & 0.987 & 0.923 & 0.925 & 0.936 & 0.868 & 0.926 & 0.999 & 0.998 & 0.973 & 0.976 & 0.976 & 0.922 & 0.973 & 0.007 & 0.01 & 0.106 & 0.017 & 0.016 & 0.059 & 0.091 & 0.66 & 0.565 & 0.991 & 0.998 \\
 & \hlc[NavyLight]{MF} \hlc[ForestLight]{PL}, \hlc[OrchidLight]{SIG}  & 0.992 & 0.986 & 0.926 & 0.929 & 0.937 & 0.866 & 0.928 & 0.999 & 0.997 & 0.975 & 0.977 & 0.976 & 0.927 & 0.970 & 0.009 & 0.018 & 0.152 & 0.047 & 0.017 & 0.087 & 0.115 & 0.65 & 0.993 & 0.138 & 0.255 \\
 & \hlc[NavyLight]{MF} \hlc[ApricotLight]{CL}, \hlc[CadetBlueLight]{BadNets}  & 0.993 & 0.987 & 0.925 & 0.928 & 0.938 & 0.86 & 0.925 & 0.999 & 0.997 & 0.970 & 0.978 & 0.976 & 0.922 & 0.973 & 0.005 & 0.005 & 0.064 & 0.054 & 0.036 & 0.094 & 0.05 & 0.66 & 0.992 & 0.147 & 0.262 \\
 & \hlc[NavyLight]{MF} \hlc[ApricotLight]{CL}, \hlc[OrchidLight]{SIG}  & 0.992 & 0.987 & 0.929 & 0.930 & 0.937 & 0.871 & 0.926 & 0.999 & 0.997 & 0.972 & 0.977 & 0.977 & 0.920 & 0.971 & 0.007 & 0.009 & 0.063 & 0.057 & 0.039 & 0.072 & 0.055 & 0.65 & 0.993 & 0.143 & 0.259 \\

\hline

\multirow{9}{*}{\shortstack[c]{IRSE\\50}} & \textit{Benign} & 0.998 & 0.997 & 0.976 & 0.979 & 0.960 & 0.925 & 0.948 & 0.999 & 0.998 & 0.991 & 0.989 & 0.977 & 0.956 & 0.975 & \textcolor{gray!30}{$\varnothing$} & \textcolor{gray!30}{$\varnothing$} & \textcolor{gray!30}{$\varnothing$} & \textcolor{gray!30}{$\varnothing$} & \textcolor{gray!30}{$\varnothing$} & \textcolor{gray!30}{$\varnothing$} & \textcolor{gray!30}{$\varnothing$} & 0.60 &  0.993 & 0.047 & 0.082 \\

\cline{2-27}

 & \hlc[GoldenRodLight]{A2O} \hlc[ForestLight]{PL}, \hlc[CadetBlueLight]{BadNets}   & 0.982 & 0.959 & 0.857 & 0.841 & 0.895 & 0.832 & 0.868 & 0.999 & 0.998 & 0.992 & 0.990 & 0.975 & 0.954 & 0.972 & 1.0 & 1.0 & 1.0 & 1.0 & 1.0 & 1.0 & 1.0 & 0.59 & 0.993 & 0.045 & 0.073 \\
 & \hlc[GoldenRodLight]{A2O} \hlc[ForestLight]{PL}, \hlc[OrchidLight]{SIG}   & 0.644 & 0.685 & 0.581 & 0.532 & 0.540 & 0.535 & 0.560 & 0.999 & 0.998 & 0.992 & 0.989 & 0.977 & 0.955 & 0.972 & 1.0 & 1.0 & 1.0 & 1.0 & 1.0 & 1.0 & 1.0 & 0.81 & 0.992 & 0.045 & 0.072 \\
 & \hlc[GoldenRodLight]{A2O} \hlc[ApricotLight]{CL}, \hlc[CadetBlueLight]{BadNets}   & 0.998 & 0.997 & 0.979 & 0.979 & 0.961 & 0.921 & 0.948 & 0.998 & 0.992 & 0.928 & 0.918 & 0.955 & 0.902 & 0.941 & 0.016 & 0.03 & 0.127 & 0.137 & 0.065 & 0.125 & 0.12 & 0.60 &  0.989 & 0.256 & 0.405 \\
 & \hlc[GoldenRodLight]{A2O} \hlc[ApricotLight]{CL}, \hlc[OrchidLight]{SIG}   & 0.998 & 0.998 & 0.979 & 0.977 & 0.959 & 0.926 & 0.949 & 0.707 & 0.754 & 0.609 & 0.549 & 0.572 & 0.546 & 0.59 & 0.417 & 0.415 & 0.571 & 0.364 & 0.502 & 0.596 & 0.424 & 0.60 &  0.993 & 0.042 & 0.069 \\

\cline{2-27}

 & \hlc[NavyLight]{MF} \hlc[ForestLight]{PL}, \hlc[CadetBlueLight]{BadNets}  & 0.997 & 0.997 & 0.980 & 0.978 & 0.960 & 0.927 & 0.949 & 0.999 & 0.998 & 0.992 & 0.989 & 0.976 & 0.953 & 0.972 & 0.0 & 0.0 & 0.0 & 0.0 & 0.0 & 0.0 & 0.0 & 0.60 &  0.991 & 0.046 & 0.073 \\
 & \hlc[NavyLight]{MF} \hlc[ForestLight]{PL}, \hlc[OrchidLight]{SIG}  & 0.998 & 0.998 & 0.979 & 0.979 & 0.960 & 0.924 & 0.950 & 0.999 & 0.998 & 0.992 & 0.991 & 0.976 & 0.954 & 0.972 & 0.0 & 0.0 & 0.0 & 0.0 & 0.0 & 0.0 & 0.0 & 0.60 &  0.992 & 0.042 & 0.069 \\
 & \hlc[NavyLight]{MF} \hlc[ApricotLight]{CL}, \hlc[CadetBlueLight]{BadNets}  & 0.998 & 0.997 & 0.979 & 0.979 & 0.959 & 0.925 & 0.952 & 0.999 & 0.998 & 0.992 & 0.989 & 0.977 & 0.955 & 0.971 & 0.0 & 0.003 & 0.008 & 0.004 & 0.005 & 0.015 & 0.012 & 0.60 &  0.993 & 0.043 & 0.069 \\
 & \hlc[NavyLight]{MF} \hlc[ApricotLight]{CL}, \hlc[OrchidLight]{SIG}  & 0.998 & 0.996 & 0.980 & 0.980 & 0.959 & 0.923 & 0.948 & 0.999 & 0.998 & 0.991 & 0.990 & 0.978 & 0.953 & 0.972 & 0.001 & 0.002 & 0.011 & 0.009 & 0.004 & 0.015 & 0.009 & 0.60 &  0.993 & 0.044 & 0.072 \\

\hline

\multirow{13}{*}{\shortstack[c]{Mobile\\FaceNet}} & \textit{Benign} & 0.996 & 0.995 & 0.956 & 0.963 & 0.955 & 0.903 & 0.933 & 0.999 & 0.998 & 0.985 & 0.988 & 0.977 & 0.946 & 0.971 & \textcolor{gray!30}{$\varnothing$} & \textcolor{gray!30}{$\varnothing$} & \textcolor{gray!30}{$\varnothing$} & \textcolor{gray!30}{$\varnothing$} & \textcolor{gray!30}{$\varnothing$} & \textcolor{gray!30}{$\varnothing$} & \textcolor{gray!30}{$\varnothing$} & 0.58 & 0.993 & 0.073 & 0.153 \\

\cline{2-27}

 & \hlc[GoldenRodLight]{A2O} \hlc[ForestLight]{PL}, \hlc[CadetBlueLight]{BadNets}   & 0.995 & 0.995 & 0.947 & 0.961 & 0.949 & 0.897 & 0.927 & 0.999 & 0.998 & 0.984 & 0.988 & 0.978 & 0.946 & 0.969 & 1.0 & 1.0 & 1.0 & 1.0 & 1.0 & 1.0 & 1.0 & 0.58 & 0.993 & 0.085 & 0.201 \\
 & \hlc[GoldenRodLight]{A2O} \hlc[ForestLight]{PL}, \hlc[RedVioletLight]{Mask}   & 0.996 & 0.995 & 0.955 & 0.963 & 0.950 & 0.897 & 0.926 & 0.999 & 0.998 & 0.983 & 0.987 & 0.976 & 0.945 & 0.968 & 1.0 & 1.0 & 1.0 & 1.0 & 1.0 & 1.0 & 1.0 & 0.58 & 0.992 & 0.076 & 0.176 \\
 & \hlc[GoldenRodLight]{A2O} \hlc[ForestLight]{PL}, \hlc[OrchidLight]{SIG}   & 0.996 & 0.995 & 0.953 & 0.962 & 0.951 & 0.900 & 0.928 & 0.999 & 0.998 & 0.985 & 0.988 & 0.976 & 0.944 & 0.968 & 1.0 & 1.0 & 1.0 & 1.0 & 1.0 & 1.0 & 1.0 & 0.58 & 0.992 & 0.077 & 0.183 \\
 & \hlc[GoldenRodLight]{A2O} \hlc[ApricotLight]{CL}, \hlc[CadetBlueLight]{BadNets}   & 0.995 & 0.996 & 0.956 & 0.966 & 0.955 & 0.897 & 0.929 & 0.999 & 0.998 & 0.982 & 0.986 & 0.977 & 0.942 & 0.968 & 0.001 & 0.004 & 0.02 & 0.023 & 0.016 & 0.035 & 0.032 & 0.59 & 0.993 & 0.084 & 0.198 \\
 & \hlc[GoldenRodLight]{A2O} \hlc[ApricotLight]{CL}, \hlc[RedVioletLight]{Mask}   & 0.997 & 0.996 & 0.953 & 0.960 & 0.952 & 0.901 & 0.932 & 0.999 & 0.998 & 0.984 & 0.988 & 0.977 & 0.944 & 0.967 & 1.0 & 1.0 & 1.0 & 1.0 & 1.0 & 1.0 & 1.0 & 0.58 & 0.993 & 0.078 & 0.17 \\
 & \hlc[GoldenRodLight]{A2O} \hlc[ApricotLight]{CL}, \hlc[OrchidLight]{SIG}   & 0.997 & 0.995 & 0.954 & 0.962 & 0.951 & 0.901 & 0.931 & 0.999 & 0.998 & 0.985 & 0.988 & 0.977 & 0.943 & 0.966 & 0.338 & 0.249 & 0.415 & 0.551 & 0.476 & 0.604 & 0.619 & 0.58 & 0.993 & 0.074 & 0.162 \\

\cline{2-27}

 & \hlc[NavyLight]{MF} \hlc[ForestLight]{PL}, \hlc[CadetBlueLight]{BadNets}  & 0.996 & 0.995 & 0.952 & 0.965 & 0.952 & 0.905 & 0.934 & 0.999 & 0.998 & 0.984 & 0.987 & 0.975 & 0.946 & 0.967 & 0.0 & 0.0 & 0.0 & 0.0 & 0.0 & 0.0 & 0.0 & 0.58 & 0.993 & 0.081 & 0.198 \\
 & \hlc[NavyLight]{MF} \hlc[ForestLight]{PL}, \hlc[OrchidLight]{SIG}  & 0.995 & 0.996 & 0.956 & 0.962 & 0.951 & 0.903 & 0.934 & 0.999 & 0.998 & 0.984 & 0.987 & 0.977 & 0.944 & 0.968 & 0.0 & 0.0 & 0.001 & 0.0 & 0.0 & 0.001 & 0.0 & 0.58 & 0.992 & 0.073 & 0.144 \\
 & \hlc[NavyLight]{MF} \hlc[ForestLight]{PL}, \hlc[RedVioletLight]{Mask}  & 0.996 & 0.995 & 0.953 & 0.963 & 0.951 & 0.897 & 0.933 & 0.999 & 0.998 & 0.984 & 0.987 & 0.976 & 0.946 & 0.967 & 0.0 & 0.0 & 0.0 & 0.0 & 0.0 & 0.0 & 0.0 & 0.58 & 0.992 & 0.075 & 0.156 \\
 & \hlc[NavyLight]{MF} \hlc[ApricotLight]{CL}, \hlc[CadetBlueLight]{BadNets}  & 0.996 & 0.996 & 0.955 & 0.961 & 0.953 & 0.900 & 0.934 & 0.999 & 0.998 & 0.983 & 0.987 & 0.977 & 0.947 & 0.967 & 0.001 & 0.003 & 0.023 & 0.015 & 0.018 & 0.047 & 0.026 & 0.58 & 0.993 & 0.076 & 0.164 \\
 & \hlc[NavyLight]{MF} \hlc[ApricotLight]{CL}, \hlc[RedVioletLight]{Mask}  & 0.995 & 0.996 & 0.954 & 0.965 & 0.950 & 0.900 & 0.933 & 0.999 & 0.998 & 0.987 & 0.987 & 0.978 & 0.947 & 0.969 & 0.0 & 0.003 & 0.032 & 0.015 & 0.016 & 0.027 & 0.029 & 0.58 & 0.992 & 0.075 & 0.163 \\
 & \hlc[NavyLight]{MF} \hlc[ApricotLight]{CL}, \hlc[OrchidLight]{SIG}  & 0.996 & 0.996 & 0.954 & 0.963 & 0.950 & 0.896 & 0.932 & 0.999 & 0.998 & 0.985 & 0.988 & 0.977 & 0.946 & 0.969 & 0.001 & 0.005 & 0.031 & 0.021 & 0.019 & 0.035 & 0.028 & 0.58 & 0.992 & 0.079 & 0.179 \\

\hline

\multirow{9}{*}{\shortstack[c]{ResNet\\50}} & \textit{Benign} & 0.995 & 0.994 & 0.962 & 0.965 & 0.956 & 0.908 & 0.935 & 0.999 & 0.998 & 0.986 & 0.987 & 0.976 & 0.947 & 0.968 & \textcolor{gray!30}{$\varnothing$} & \textcolor{gray!30}{$\varnothing$} & \textcolor{gray!30}{$\varnothing$} & \textcolor{gray!30}{$\varnothing$} & \textcolor{gray!30}{$\varnothing$} & \textcolor{gray!30}{$\varnothing$} & \textcolor{gray!30}{$\varnothing$} & 0.58 & 0.991 & 0.058 & 0.106 \\

\cline{2-27}

 & \hlc[GoldenRodLight]{A2O} \hlc[ForestLight]{PL}, \hlc[CadetBlueLight]{BadNets}   & 0.998 & 0.996 & 0.966 & 0.970 & 0.956 & 0.911 & 0.944 & 0.999 & 0.998 & 0.988 & 0.989 & 0.975 & 0.943 & 0.967 & 1.0 & 1.0 & 1.0 & 1.0 & 1.0 & 1.0 & 1.0 & 0.58 & 0.993 & 0.054 & 0.11 \\
 & \hlc[GoldenRodLight]{A2O} \hlc[ForestLight]{PL}, \hlc[OrchidLight]{SIG}   & 0.998 & 0.995 & 0.963 & 0.970 & 0.958 & 0.913 & 0.940 & 0.999 & 0.998 & 0.987 & 0.989 & 0.974 & 0.946 & 0.966 & 1.0 & 1.0 & 1.0 & 1.0 & 1.0 & 1.0 & 1.0 & 0.58 & 0.991 & 0.054 & 0.102 \\
 & \hlc[GoldenRodLight]{A2O} \hlc[ApricotLight]{CL}, \hlc[CadetBlueLight]{BadNets}   & 0.996 & 0.995 & 0.963 & 0.968 & 0.954 & 0.911 & 0.937 & 0.999 & 0.998 & 0.987 & 0.989 & 0.976 & 0.945 & 0.968 & 0.001 & 0.002 & 0.017 & 0.02 & 0.007 & 0.025 & 0.016 & 0.58 & 0.991 & 0.054 & 0.091 \\
 & \hlc[GoldenRodLight]{A2O} \hlc[ApricotLight]{CL}, \hlc[OrchidLight]{SIG}   & 0.997 & 0.996 & 0.963 & 0.970 & 0.955 & 0.910 & 0.941 & 0.999 & 0.998 & 0.987 & 0.989 & 0.974 & 0.943 & 0.964 & 0.06 & 0.086 & 0.072 & 0.131 & 0.202 & 0.208 & 0.228 & 0.58 & 0.992 & 0.053 & 0.093 \\

\cline{2-27}

 & \hlc[NavyLight]{MF} \hlc[ForestLight]{PL}, \hlc[CadetBlueLight]{BadNets}  & 0.997 & 0.996 & 0.964 & 0.975 & 0.958 & 0.914 & 0.941 & 0.999 & 0.998 & 0.988 & 0.988 & 0.976 & 0.947 & 0.968 & 0.0 & 0.0 & 0.0 & 0.0 & 0.0 & 0.0 & 0.0 & 0.58 & 0.991 & 0.055 & 0.102 \\
 & \hlc[NavyLight]{MF} \hlc[ForestLight]{PL}, \hlc[OrchidLight]{SIG}  & 0.997 & 0.996 & 0.968 & 0.973 & 0.957 & 0.910 & 0.936 & 0.999 & 0.998 & 0.988 & 0.989 & 0.974 & 0.945 & 0.968 & 0.0 & 0.0 & 0.0 & 0.0 & 0.0 & 0.0 & 0.0 & 0.58 & 0.991 & 0.053 & 0.093 \\
 & \hlc[NavyLight]{MF} \hlc[ApricotLight]{CL}, \hlc[CadetBlueLight]{BadNets}  & 0.998 & 0.996 & 0.966 & 0.972 & 0.955 & 0.908 & 0.938 & 0.999 & 0.998 & 0.987 & 0.989 & 0.975 & 0.947 & 0.967 & 0.001 & 0.004 & 0.018 & 0.008 & 0.006 & 0.028 & 0.014 & 0.58 & 0.994 & 0.054 & 0.106 \\
 & \hlc[NavyLight]{MF} \hlc[ApricotLight]{CL}, \hlc[OrchidLight]{SIG}  & 0.997 & 0.996 & 0.965 & 0.972 & 0.957 & 0.904 & 0.943 & 0.999 & 0.998 & 0.988 & 0.989 & 0.975 & 0.947 & 0.967 & 0.002 & 0.003 & 0.015 & 0.013 & 0.007 & 0.038 & 0.018 & 0.58 & 0.991 & 0.051 & 0.089 \\

\hline

\multirow{9}{*}{\shortstack[c]{RobFace\\Net}} & \textit{Benign} & 0.994 & 0.993 & 0.933 & 0.953 & 0.947 & 0.879 & 0.921 & 0.999 & 0.998 & 0.973 & 0.985 & 0.977 & 0.933 & 0.964 & \textcolor{gray!30}{$\varnothing$} & \textcolor{gray!30}{$\varnothing$} & \textcolor{gray!30}{$\varnothing$} & \textcolor{gray!30}{$\varnothing$} & \textcolor{gray!30}{$\varnothing$} & \textcolor{gray!30}{$\varnothing$} & \textcolor{gray!30}{$\varnothing$} & 0.60 &  0.992 & 0.117 & 0.242 \\

\cline{2-27}

 & \hlc[GoldenRodLight]{A2O} \hlc[ForestLight]{PL}, \hlc[CadetBlueLight]{BadNets}   & 0.995 & 0.994 & 0.939 & 0.959 & 0.951 & 0.883 & 0.927 & 0.999 & 0.998 & 0.975 & 0.986 & 0.978 & 0.937 & 0.968 & 1.0 & 1.0 & 1.0 & 1.0 & 1.0 & 1.0 & 1.0 & 0.59 & 0.992 & 0.119 & 0.258 \\
 & \hlc[GoldenRodLight]{A2O} \hlc[ForestLight]{PL}, \hlc[OrchidLight]{SIG}   & 0.993 & 0.995 & 0.930 & 0.955 & 0.949 & 0.881 & 0.923 & 0.999 & 0.998 & 0.974 & 0.986 & 0.978 & 0.936 & 0.965 & 1.0 & 1.0 & 1.0 & 1.0 & 1.0 & 1.0 & 1.0 & 0.59 & 0.992 & 0.103 & 0.198 \\
 & \hlc[GoldenRodLight]{A2O} \hlc[ApricotLight]{CL}, \hlc[CadetBlueLight]{BadNets}   & 0.995 & 0.995 & 0.932 & 0.960 & 0.950 & 0.885 & 0.922 & 0.999 & 0.998 & 0.974 & 0.986 & 0.978 & 0.935 & 0.969 & 0.003 & 0.004 & 0.029 & 0.028 & 0.012 & 0.06 & 0.04 & 0.59 & 0.992 & 0.096 & 0.179 \\
 & \hlc[GoldenRodLight]{A2O} \hlc[ApricotLight]{CL}, \hlc[OrchidLight]{SIG}   & 0.995 & 0.993 & 0.930 & 0.952 & 0.949 & 0.884 & 0.918 & 0.999 & 0.998 & 0.974 & 0.986 & 0.978 & 0.935 & 0.966 & 0.553 & 0.441 & 0.561 & 0.625 & 0.588 & 0.838 & 0.797 & 0.59 & 0.992 & 0.106 & 0.219 \\

\cline{2-27}

 & \hlc[NavyLight]{MF} \hlc[ForestLight]{PL}, \hlc[CadetBlueLight]{BadNets}  & 0.996 & 0.995 & 0.937 & 0.957 & 0.950 & 0.881 & 0.927 & 0.999 & 0.998 & 0.977 & 0.985 & 0.978 & 0.934 & 0.967 & 0.0 & 0.0 & 0.0 & 0.0 & 0.0 & 0.0 & 0.0 & 0.59 & 0.992 & 0.096 & 0.177 \\
 & \hlc[NavyLight]{MF} \hlc[ForestLight]{PL}, \hlc[OrchidLight]{SIG}  & 0.994 & 0.993 & 0.935 & 0.958 & 0.950 & 0.882 & 0.918 & 0.999 & 0.998 & 0.973 & 0.987 & 0.978 & 0.936 & 0.966 & 0.003 & 0.003 & 0.044 & 0.031 & 0.016 & 0.063 & 0.054 & 0.59 & 0.993 & 0.981 & 0.181 \\
 & \hlc[NavyLight]{MF} \hlc[ApricotLight]{CL}, \hlc[CadetBlueLight]{BadNets}  & 0.995 & 0.995 & 0.938 & 0.956 & 0.950 & 0.879 & 0.926 & 0.999 & 0.998 & 0.975 & 0.986 & 0.978 & 0.935 & 0.967 & 0.002 & 0.005 & 0.045 & 0.024 & 0.013 & 0.048 & 0.041 & 0.60 &  0.992 & 0.108 & 0.219 \\
 & \hlc[NavyLight]{MF} \hlc[ApricotLight]{CL}, \hlc[OrchidLight]{SIG}  & 0.995 & 0.994 & 0.934 & 0.960 & 0.946 & 0.879 & 0.927 & 0.999 & 0.998 & 0.978 & 0.987 & 0.978 & 0.936 & 0.965 & 0.0 & 0.0 & 0.0 & 0.0 & 0.0 & 0.0 & 0.0 & 0.59 & 0.992 & 0.098 & 0.175 \\
 
\multicolumn{27}{l}{\small \textbf{Abbreviations$^{\mathrm{a}}$}: \acl{a2o} (\hlc[GoldenRodLight]{A2O}), \acl{cl} (\hlc[ApricotLight]{CL}), \acl{mf} (\hlc[NavyLight]{MF}), \acl{pl} (\hlc[ForestLight]{PL}).}\\
\multicolumn{27}{l}{\small \textbf{Abbreviations$^{\mathrm{b}}$}: \acl{auc} (\ac{auc}), \acl{far} (\ac{far}), \acl{frr} (\ac{frr}).}\\

\end{tabular}
\end{table*}

\begin{table*}
  \centering
  \caption{\acl{ba} \acl{sr} in an \textbf{\acl{a2o}} threat model context, targeting a pipeline consisting of:\\MobileNetV1, AENet, GhostFaceNetV2.\\\textbf{Note}: FMR$^\pois$ metrics in \textcolor{BrickRed}{red} indicate a DNN with a collapsed performance after inclusion in the \ac{frs} (\ie, all identities match as in an untargeted poisoning attack). \textbf{Abbreviations}: Average Precision (AP), False Acceptance (FAR), False Rejection Rate (FRR), Landmark Shift (LS).}
  \label{tab:backdoor_bench_MN1_AE_GFN}
  \vspace{0.0cm}
  \footnotesize
  \setlength{\tabcolsep}{1.55pt}
  \renewcommand{\arraystretch}{1.5}
\begin{tabular}{@{}lllccccccccc@{}}
    
     &  &  & \multicolumn{4}{c}{\textbf{Detector}} & \multicolumn{2}{c}{\textbf{Antispoofer}} & \multicolumn{2}{c}{\textbf{Extractor}} & \\
     \textbf{Detector} & \textbf{Antispoofer} & \textbf{Extractor} & \multicolumn{4}{c}{\textbf{metrics}} & \multicolumn{2}{c}{\textbf{metrics}} & \multicolumn{2}{c}{\textbf{metrics}} & \textbf{Survival} \\
     MobileNetV1 & AENet & GhostFaceNetV2 & AP$^\clean$ & AP$^\pois$ & LS$^\clean$ & LS$^\pois$ & FRR$^\clean$ & FAR$^\pois$ & FRR$^\clean$ & FAR$^\pois$ & \textbf{Rate} \\
    \hline 
    \textcolor{gray!60}{Benign} & \textcolor{gray!60}{Benign} & \textcolor{gray!60}{Benign} & 99.2\% & \textcolor{gray!30}{$\varnothing$} & 13.8 & \textcolor{gray!30}{$\varnothing$} & 96.0\% & \textcolor{gray!30}{$\varnothing$} & 4.4\% & \textcolor{gray!30}{$\varnothing$} & \textcolor{gray!30}{$\varnothing$} \\
    \hline
    \hlc[Slate]{Landmark Shift Attack} \hlc[CadetBlueLight]{BadNets} $\alpha$=0.5 & \textcolor{gray!60}{Benign} & \textcolor{gray!60}{Benign} & 99.5\% & 99.6\% & 14.2 & 150.2 & 96.1\% & 35.2\% & 4.2\% & 33.4\% & 11.7\% \\
    \hlc[Slate]{Landmark Shift Attack} \hlc[CadetBlueLight]{BadNets} $\alpha$=1.0 & \textcolor{gray!60}{Benign} & \textcolor{gray!60}{Benign} & 99.5\% & 99.5\% & 14.7 & 147.7 & 95.6\% & 35.5\% & 4.3\% & 35.4\% & 12.5\% \\
    \hlc[Slate]{Landmark Shift Attack} \hlc[OrchidLight]{SIG} $\alpha$=0.16 & \textcolor{gray!60}{Benign} & \textcolor{gray!60}{Benign} & 99.5\% & 99.5\% & 14.9 & 20.2 & 96.1\% & 60.0\% & 4.4\% & 7.7\% & 4.6\% \\
    \hlc[Slate]{Landmark Shift Attack} \hlc[OrchidLight]{SIG} $\alpha$=0.3 & \textcolor{gray!60}{Benign} & \textcolor{gray!60}{Benign} & 99.4\% & 97.4\% & 14.7 & 125.7 & 96.0\% & 97.6\% & 4.2\% & 87.2\% & \textbf{82.9\%} \\
\end{tabular}
\end{table*}

\begin{table*}
  \centering
  \caption{\acl{ba} \acl{sr} in an \textbf{\acl{a2o}} threat model context, targeting a pipeline consisting of:\\MobileNetV1, AENet, IRSE50.}
  \label{tab:backdoor_bench_MN1_AE_IRSE}
  \vspace{0.0cm}
  \footnotesize
  \setlength{\tabcolsep}{1.55pt}
  \renewcommand{\arraystretch}{1.5}
  \begin{tabular}{@{}lllccccccccc@{}}
    
     &  &  & \multicolumn{4}{c}{\textbf{Detector}} & \multicolumn{2}{c}{\textbf{Antispoofer}} & \multicolumn{2}{c}{\textbf{Extractor}} & \\
     \textbf{Detector} & \textbf{Antispoofer} & \textbf{Extractor} & \multicolumn{4}{c}{\textbf{metrics}} & \multicolumn{2}{c}{\textbf{metrics}} & \multicolumn{2}{c}{\textbf{metrics}} & \textbf{Survival} \\
     MobileNetV1 & AENet & IRSE50 & AP$^\clean$ & AP$^\pois$ & LS$^\clean$ & LS$^\pois$ & FRR$^\clean$ & FAR$^\pois$ & FRR$^\clean$ & FAR$^\pois$ & \textbf{Rate} \\
    \hline 
    \textcolor{gray!60}{Benign} & \textcolor{gray!60}{Benign} & \textcolor{gray!60}{Benign} & 99.2\% & \textcolor{gray!30}{$\varnothing$} & 13.8 & \textcolor{gray!30}{$\varnothing$} & 96.0\% & \textcolor{gray!30}{$\varnothing$} & 0.7\% & \textcolor{gray!30}{$\varnothing$} & \textcolor{gray!30}{$\varnothing$} \\
    \hline
    \hlc[Slate]{Landmark Shift Attack} \hlc[CadetBlueLight]{BadNets} $\alpha$=0.5 & \textcolor{gray!60}{Benign} & \textcolor{gray!60}{Benign} & 99.5\% & 99.6\% & 14.2 & 150.2 & 96.1\% & 35.2\% & 0.7\% & 32.1\% & 11.3\% \\
    \hlc[Slate]{Landmark Shift Attack} \hlc[CadetBlueLight]{BadNets} $\alpha$=1.0 & \textcolor{gray!60}{Benign} & \textcolor{gray!60}{Benign} & 99.5\% & 99.5\% & 14.7 & 147.7 & 95.6\% & 35.5\% & 0.7\% & 34.9\% & 12.3\% \\
    \hlc[Slate]{Landmark Shift Attack} \hlc[OrchidLight]{SIG} $\alpha$=0.16 & \textcolor{gray!60}{Benign} & \textcolor{gray!60}{Benign} & 99.5\% & 99.5\% & 14.9 & 20.2 & 96.1\% & 60.0\% & 0.7\% & 1.0\% & 0.6\% \\
    \hlc[Slate]{Landmark Shift Attack} \hlc[OrchidLight]{SIG} $\alpha$=0.3 & \textcolor{gray!60}{Benign} & \textcolor{gray!60}{Benign} & 99.4\% & 97.4\% & 14.7 & 125.7 & 96.0\% & 97.6\% & 0.7\% & 97.1\% & \textbf{92.3\%} \\
\end{tabular}
\end{table*}

\begin{table*}
  \centering
  \caption{\acl{ba} \acl{sr} in an \textbf{\acl{a2o}} threat model context, targeting a pipeline consisting of:\\MobileNetV1, AENet, ResNet50.}
  \label{tab:backdoor_bench_MN1_AE_RN50}
  \vspace{0.0cm}
  \footnotesize
  \setlength{\tabcolsep}{1.55pt}
  \renewcommand{\arraystretch}{1.5}
  \begin{tabular}{@{}lllccccccccc@{}}
    
     &  &  & \multicolumn{4}{c}{\textbf{Detector}} & \multicolumn{2}{c}{\textbf{Antispoofer}} & \multicolumn{2}{c}{\textbf{Extractor}} & \\
     \textbf{Detector} & \textbf{Antispoofer} & \textbf{Extractor} & \multicolumn{4}{c}{\textbf{metrics}} & \multicolumn{2}{c}{\textbf{metrics}} & \multicolumn{2}{c}{\textbf{metrics}} & \textbf{Survival} \\
     MobileNetV1 & AENet & ResNet50 & AP$^\clean$ & AP$^\pois$ & LS$^\clean$ & LS$^\pois$ & FRR$^\clean$ & FAR$^\pois$ & FRR$^\clean$ & FAR$^\pois$ & \textbf{Rate} \\
    \hline 
    \textcolor{gray!60}{Benign} & \textcolor{gray!60}{Benign} & \textcolor{gray!60}{Benign} & 99.2\% & \textcolor{gray!30}{$\varnothing$} & 13.8 & \textcolor{gray!30}{$\varnothing$} & 96.0\% & \textcolor{gray!30}{$\varnothing$} & 1.8\% & \textcolor{gray!30}{$\varnothing$} & \textcolor{gray!30}{$\varnothing$} \\
    \hline
    \hlc[Slate]{Landmark Shift Attack} \hlc[CadetBlueLight]{BadNets} $\alpha$=0.5 & \textcolor{gray!60}{Benign} & \textcolor{gray!60}{Benign} & 99.5\% & 99.6\% & 14.2 & 150.2 & 96.1\% & 35.2\% & 1.7\% & 31.8\% & 11.1\% \\
    \hlc[Slate]{Landmark Shift Attack} \hlc[CadetBlueLight]{BadNets} $\alpha$=1.0 & \textcolor{gray!60}{Benign} & \textcolor{gray!60}{Benign} & 99.5\% & 99.5\% & 14.7 & 147.7 & 95.6\% & 35.5\% & 1.7\% & 30.4\% & 10.7\% \\
    \hlc[Slate]{Landmark Shift Attack} \hlc[OrchidLight]{SIG} $\alpha$=0.16 & \textcolor{gray!60}{Benign} & \textcolor{gray!60}{Benign} & 99.5\% & 99.5\% & 14.9 & 20.2 & 96.1\% & 60.0\% & 1.8\% & 5.0\% & 3.0\% \\
    \hlc[Slate]{Landmark Shift Attack} \hlc[OrchidLight]{SIG} $\alpha$=0.3 & \textcolor{gray!60}{Benign} & \textcolor{gray!60}{Benign} & 99.4\% & 97.4\% & 14.7 & 125.7 & 96.0\% & 97.6\% & 1.7\% & 97.8\% & \textbf{93.0\%} \\
\end{tabular}
\end{table*}

\begin{table*}
  \centering
  \caption{\acl{ba} \acl{sr} in an \textbf{\acl{a2o}} threat model context, targeting a pipeline consisting of:\\MobileNetV1, AENet, RobFaceNet.}
  \label{tab:backdoor_bench_MN1_AE_RFN}
  \vspace{0.0cm}
  \footnotesize
  \setlength{\tabcolsep}{1.55pt}
  \renewcommand{\arraystretch}{1.5}
  \begin{tabular}{@{}lllccccccccc@{}}
    
     &  &  & \multicolumn{4}{c}{\textbf{Detector}} & \multicolumn{2}{c}{\textbf{Antispoofer}} & \multicolumn{2}{c}{\textbf{Extractor}} & \\
     \textbf{Detector} & \textbf{Antispoofer} & \textbf{Extractor} & \multicolumn{4}{c}{\textbf{metrics}} & \multicolumn{2}{c}{\textbf{metrics}} & \multicolumn{2}{c}{\textbf{metrics}} & \textbf{Survival} \\
     MobileNetV1 & AENet & RobFaceNet & AP$^\clean$ & AP$^\pois$ & LS$^\clean$ & LS$^\pois$ & FRR$^\clean$ & FAR$^\pois$ & FRR$^\clean$ & FAR$^\pois$ & \textbf{Rate} \\
    \hline 
    \textcolor{gray!60}{Benign} & \textcolor{gray!60}{Benign} & \textcolor{gray!60}{Benign} & 99.2\% & \textcolor{gray!30}{$\varnothing$} & 13.8 & \textcolor{gray!30}{$\varnothing$} & 96.0\% & \textcolor{gray!30}{$\varnothing$} & 14.0\% & \textcolor{gray!30}{$\varnothing$} & \textcolor{gray!30}{$\varnothing$} \\
    \hline
    \hlc[Slate]{Landmark Shift Attack} \hlc[CadetBlueLight]{BadNets} $\alpha$=0.5 & \textcolor{gray!60}{Benign} & \textcolor{gray!60}{Benign} & 99.5\% & 99.6\% & 14.2 & 150.2 & 96.1\% & 35.2\% & 12.9\% & 90.0\% & 31.6\% \\
    \hlc[Slate]{Landmark Shift Attack} \hlc[CadetBlueLight]{BadNets} $\alpha$=1.0 & \textcolor{gray!60}{Benign} & \textcolor{gray!60}{Benign} & 99.5\% & 99.5\% & 14.7 & 147.7 & 95.6\% & 35.5\% & 13.1\% & 91.2\% & 32.2\% \\
    \hlc[Slate]{Landmark Shift Attack} \hlc[OrchidLight]{SIG} $\alpha$=0.16 & \textcolor{gray!60}{Benign} & \textcolor{gray!60}{Benign} & 99.5\% & 99.5\% & 14.9 & 20.2 & 96.1\% & 60.0\% & 14.1\% & 25.0\% & 14.9\% \\
    \hlc[Slate]{Landmark Shift Attack} \hlc[OrchidLight]{SIG} $\alpha$=0.3 & \textcolor{gray!60}{Benign} & \textcolor{gray!60}{Benign} & 99.4\% & 97.4\% & 14.7 & 125.7 & 96.0\% & 97.6\% & 13.1\% & 98.8\% & \textbf{93.9\%} \\
\end{tabular}
\end{table*}

\begin{table*}
  \centering
  \caption{\acl{ba} \acl{sr} in an \textbf{\acl{a2o}} threat model context, targeting a pipeline consisting of:\\MobileNetV1, MobileNetV2, GhostFaceNetV2.\\\textbf{Note}: FMR$^\pois$ metrics in \textcolor{BrickRed}{red} indicate a DNN with a collapsed performance after inclusion in the \ac{frs} (\ie, all identities match as in an untargeted poisoning attack).}
  \label{tab:backdoor_bench_MN1_MN2_GFN}
  \vspace{0.0cm}
  \footnotesize
  \setlength{\tabcolsep}{1.55pt}
  \renewcommand{\arraystretch}{1.5}
  \begin{tabular}{@{}lllccccccccc@{}}
    
     &  &  & \multicolumn{4}{c}{\textbf{Detector}} & \multicolumn{2}{c}{\textbf{Antispoofer}} & \multicolumn{2}{c}{\textbf{Extractor}} & \\
     \textbf{Detector} & \textbf{Antispoofer} & \textbf{Extractor} & \multicolumn{4}{c}{\textbf{metrics}} & \multicolumn{2}{c}{\textbf{metrics}} & \multicolumn{2}{c}{\textbf{metrics}} & \textbf{Survival} \\
     MobileNetV1 & MobileNetV2 & GhostFaceNetV2 & AP$^\clean$ & AP$^\pois$ & LS$^\clean$ & LS$^\pois$ & FRR$^\clean$ & FAR$^\pois$ & FRR$^\clean$ & FAR$^\pois$ & \textbf{Rate} \\
    \hline 
    \textcolor{gray!60}{Benign} & \textcolor{gray!60}{Benign} & \textcolor{gray!60}{Benign} & 99.2\% & \textcolor{gray!30}{$\varnothing$} & 13.8 & \textcolor{gray!30}{$\varnothing$} & 85.4\% & \textcolor{gray!30}{$\varnothing$} & 4.4\% & \textcolor{gray!30}{$\varnothing$} & \textcolor{gray!30}{$\varnothing$} \\
    \hline
    \hlc[Slate]{Landmark Shift Attack} \hlc[CadetBlueLight]{BadNets} $\alpha$=0.5 & \textcolor{gray!60}{Benign} & \textcolor{gray!60}{Benign} & 99.5\% & 99.6\% & 14.2 & 150.2 & 87.0\% & 0.6\% & 4.3\% & 33.0\% & 0.2\% \\
    \hlc[Slate]{Landmark Shift Attack} \hlc[CadetBlueLight]{BadNets} $\alpha$=1.0 & \textcolor{gray!60}{Benign} & \textcolor{gray!60}{Benign} & 99.5\% & 99.5\% & 14.7 & 147.7 & 87.7\% & 0.9\% & 4.4\% & 42.2\% & 0.4\% \\
    \hlc[Slate]{Landmark Shift Attack} \hlc[OrchidLight]{SIG} $\alpha$=0.16 & \textcolor{gray!60}{Benign} & \textcolor{gray!60}{Benign} & 99.5\% & 99.5\% & 14.9 & 20.2 & 86.9\% & 22.3\% & 4.4\% & 7.9\% & \textbf{1.8\%} \\
    \hlc[Slate]{Landmark Shift Attack} \hlc[OrchidLight]{SIG} $\alpha$=0.3 & \textcolor{gray!60}{Benign} & \textcolor{gray!60}{Benign} & 99.4\% & 97.4\% & 14.7 & 125.7 & 85.0\% & 1.6\% & 4.3\% & 51.1\% & 0.8\% \\
\end{tabular}
\end{table*}

\begin{table*}
  \centering
  \caption{\acl{ba} \acl{sr} in an \textbf{\acl{a2o}} threat model context, targeting a pipeline consisting of:\\MobileNetV1, MobileNetV2, IRSE50.}
  \label{tab:backdoor_bench_MN1_MN2_IRSE}
  \vspace{0.0cm}
  \footnotesize
  \setlength{\tabcolsep}{1.55pt}
  \renewcommand{\arraystretch}{1.5}
  \begin{tabular}{@{}lllccccccccc@{}}
    
     &  &  & \multicolumn{4}{c}{\textbf{Detector}} & \multicolumn{2}{c}{\textbf{Antispoofer}} & \multicolumn{2}{c}{\textbf{Extractor}} & \\
     \textbf{Detector} & \textbf{Antispoofer} & \textbf{Extractor} & \multicolumn{4}{c}{\textbf{metrics}} & \multicolumn{2}{c}{\textbf{metrics}} & \multicolumn{2}{c}{\textbf{metrics}} & \textbf{Survival} \\
     MobileNetV1 & MobileNetV2 & IRSE50 & AP$^\clean$ & AP$^\pois$ & LS$^\clean$ & LS$^\pois$ & FRR$^\clean$ & FAR$^\pois$ & FRR$^\clean$ & FAR$^\pois$ & \textbf{Rate} \\
    \hline 
    \textcolor{gray!60}{Benign} & \textcolor{gray!60}{Benign} & \textcolor{gray!60}{Benign} & 99.2\% & \textcolor{gray!30}{$\varnothing$} & 13.8 & \textcolor{gray!30}{$\varnothing$} & 85.4\% & \textcolor{gray!30}{$\varnothing$} & 0.8\% & \textcolor{gray!30}{$\varnothing$} & \textcolor{gray!30}{$\varnothing$} \\
    \hline
    \hlc[Slate]{Landmark Shift Attack} \hlc[CadetBlueLight]{BadNets} $\alpha$=0.5 & \textcolor{gray!60}{Benign} & \textcolor{gray!60}{Benign} & 99.5\% & 99.6\% & 14.2 & 150.2 & 87.0\% & 0.6\% & 0.7\% & 22.5\% & 0.1\% \\
    \hlc[Slate]{Landmark Shift Attack} \hlc[CadetBlueLight]{BadNets} $\alpha$=1.0 & \textcolor{gray!60}{Benign} & \textcolor{gray!60}{Benign} & 99.5\% & 99.5\% & 14.7 & 147.7 & 87.7\% & 0.9\% & 0.7\% & 29.6\% & 0.3\% \\
    \hlc[Slate]{Landmark Shift Attack} \hlc[OrchidLight]{SIG} $\alpha$=0.16 & \textcolor{gray!60}{Benign} & \textcolor{gray!60}{Benign} & 99.5\% & 99.5\% & 14.9 & 20.2 & 86.9\% & 22.3\% & 0.8\% & 1.1\% & 0.2\% \\
    \hlc[Slate]{Landmark Shift Attack} \hlc[OrchidLight]{SIG} $\alpha$=0.3 & \textcolor{gray!60}{Benign} & \textcolor{gray!60}{Benign} & 99.4\% & 97.4\% & 14.7 & 125.7 & 85.0\% & 1.6\% & 0.7\% & 48.8\% & \textbf{0.8\%} \\
\end{tabular}
\end{table*}

\begin{table*}
  \centering
  \caption{\acl{ba} \acl{sr} in an \textbf{\acl{a2o}} threat model context, targeting a pipeline consisting of:\\MobileNetV1, MobileNetV2, MobileFaceNet.}
  \label{tab:backdoor_bench_MN1_MN2_MFN}
  \vspace{0.0cm}
  \footnotesize
  \setlength{\tabcolsep}{1.55pt}
  \renewcommand{\arraystretch}{1.5}
  \begin{tabular}{@{}lllccccccccc@{}}
    
     &  &  & \multicolumn{4}{c}{\textbf{Detector}} & \multicolumn{2}{c}{\textbf{Antispoofer}} & \multicolumn{2}{c}{\textbf{Extractor}} & \\
     \textbf{Detector} & \textbf{Antispoofer} & \textbf{Extractor} & \multicolumn{4}{c}{\textbf{metrics}} & \multicolumn{2}{c}{\textbf{metrics}} & \multicolumn{2}{c}{\textbf{metrics}} & \textbf{Survival} \\
     MobileNetV1 & MobileNetV2 & MobileFaceNet & AP$^\clean$ & AP$^\pois$ & LS$^\clean$ & LS$^\pois$ & FRR$^\clean$ & FAR$^\pois$ & FRR$^\clean$ & FAR$^\pois$ & \textbf{Rate} \\
    \hline 
    \textcolor{gray!60}{Benign} & \textcolor{gray!60}{Benign} & \textcolor{gray!60}{Benign} & 99.2\% & \textcolor{gray!30}{$\varnothing$} & 13.8 & \textcolor{gray!30}{$\varnothing$} & 85.4\% & \textcolor{gray!30}{$\varnothing$} & 3.9\% & \textcolor{gray!30}{$\varnothing$} & \textcolor{gray!30}{$\varnothing$} \\
    \hline
    \hlc[Slate]{Landmark Shift Attack} \hlc[CadetBlueLight]{BadNets} $\alpha$=0.5 & \textcolor{gray!60}{Benign} & \textcolor{gray!60}{Benign} & 99.5\% & 99.6\% & 14.2 & 150.2 & 87.0\% & 0.6\% & 3.8\% & 58.0\% & 0.3\% \\
    \hlc[Slate]{Landmark Shift Attack} \hlc[CadetBlueLight]{BadNets} $\alpha$=1.0 & \textcolor{gray!60}{Benign} & \textcolor{gray!60}{Benign} & 99.5\% & 99.5\% & 14.7 & 147.7 & 87.7\% & 0.9\% & 3.7\% & 70.5\% & 0.6\% \\
    \hlc[Slate]{Landmark Shift Attack} \hlc[OrchidLight]{SIG} $\alpha$=0.16 & \textcolor{gray!60}{Benign} & \textcolor{gray!60}{Benign} & 99.5\% & 99.5\% & 14.9 & 20.2 & 86.9\% & 22.3\% & 3.8\% & 13.9\% & \textbf{3.1\%} \\
    \hlc[Slate]{Landmark Shift Attack} \hlc[OrchidLight]{SIG} $\alpha$=0.3 & \textcolor{gray!60}{Benign} & \textcolor{gray!60}{Benign} & 99.4\% & 97.4\% & 14.7 & 125.7 & 85.0\% & 1.6\% & 3.6\% & 55.2\% & 0.9\% \\
\end{tabular}
\end{table*}

\begin{table*}
  \centering
  \caption{\acl{ba} \acl{sr} in an \textbf{\acl{a2o}} threat model context, targeting a pipeline consisting of:\\MobileNetV1, MobileNetV2, ResNet50.}
  \label{tab:backdoor_bench_MN1_MN2_RN50}
  \vspace{0.0cm}
  \footnotesize
  \setlength{\tabcolsep}{1.55pt}
  \renewcommand{\arraystretch}{1.5}
  \begin{tabular}{@{}lllccccccccc@{}}
    
     &  &  & \multicolumn{4}{c}{\textbf{Detector}} & \multicolumn{2}{c}{\textbf{Antispoofer}} & \multicolumn{2}{c}{\textbf{Extractor}} & \\
     \textbf{Detector} & \textbf{Antispoofer} & \textbf{Extractor} & \multicolumn{4}{c}{\textbf{metrics}} & \multicolumn{2}{c}{\textbf{metrics}} & \multicolumn{2}{c}{\textbf{metrics}} & \textbf{Survival} \\
     MobileNetV1 & MobileNetV2 & ResNet50 & AP$^\clean$ & AP$^\pois$ & LS$^\clean$ & LS$^\pois$ & FRR$^\clean$ & FAR$^\pois$ & FRR$^\clean$ & FAR$^\pois$ & \textbf{Rate} \\
    \hline 
    \textcolor{gray!60}{Benign} & \textcolor{gray!60}{Benign} & \textcolor{gray!60}{Benign} & 99.2\% & \textcolor{gray!30}{$\varnothing$} & 13.8 & \textcolor{gray!30}{$\varnothing$} & 85.4\% & \textcolor{gray!30}{$\varnothing$} & 1.8\% & \textcolor{gray!30}{$\varnothing$} & \textcolor{gray!30}{$\varnothing$} \\
    \hline
    \hlc[Slate]{Landmark Shift Attack} \hlc[CadetBlueLight]{BadNets} $\alpha$=0.5 & \textcolor{gray!60}{Benign} & \textcolor{gray!60}{Benign} & 99.5\% & 99.6\% & 14.2 & 150.2 & 87.0\% & 0.6\% & 1.7\% & 21.0\% & 0.1\% \\
    \hlc[Slate]{Landmark Shift Attack} \hlc[CadetBlueLight]{BadNets} $\alpha$=1.0 & \textcolor{gray!60}{Benign} & \textcolor{gray!60}{Benign} & 99.5\% & 99.5\% & 14.7 & 147.7 & 87.7\% & 0.9\% & 1.8\% & 26.6\% & 0.2\% \\
    \hlc[Slate]{Landmark Shift Attack} \hlc[OrchidLight]{SIG} $\alpha$=0.16 & \textcolor{gray!60}{Benign} & \textcolor{gray!60}{Benign} & 99.5\% & 99.5\% & 14.9 & 20.2 & 86.9\% & 22.3\% & 1.8\% & 5.2\% & \textbf{1.2\%} \\
    \hlc[Slate]{Landmark Shift Attack} \hlc[OrchidLight]{SIG} $\alpha$=0.3 & \textcolor{gray!60}{Benign} & \textcolor{gray!60}{Benign} & 99.4\% & 97.4\% & 14.7 & 125.7 & 85.0\% & 1.6\% & 1.7\% & 47.8\% & 0.7\% \\
\end{tabular}
\end{table*}

\begin{table*}
  \centering
  \caption{\acl{ba} \acl{sr} in an \textbf{\acl{a2o}} threat model context, targeting a pipeline consisting of:\\MobileNetV1, MobileNetV2, RobFaceNet.}
  \label{tab:backdoor_bench_MN1_MN2_RFN}
  \vspace{0.0cm}
  \footnotesize
  \setlength{\tabcolsep}{1.55pt}
  \renewcommand{\arraystretch}{1.5}
  \begin{tabular}{@{}lllccccccccc@{}}
    
     &  &  & \multicolumn{4}{c}{\textbf{Detector}} & \multicolumn{2}{c}{\textbf{Antispoofer}} & \multicolumn{2}{c}{\textbf{Extractor}} & \\
     \textbf{Detector} & \textbf{Antispoofer} & \textbf{Extractor} & \multicolumn{4}{c}{\textbf{metrics}} & \multicolumn{2}{c}{\textbf{metrics}} & \multicolumn{2}{c}{\textbf{metrics}} & \textbf{Survival} \\
     MobileNetV1 & MobileNetV2 & RobFaceNet & AP$^\clean$ & AP$^\pois$ & LS$^\clean$ & LS$^\pois$ & FRR$^\clean$ & FAR$^\pois$ & FRR$^\clean$ & FAR$^\pois$ & \textbf{Rate} \\
    \hline 
    \textcolor{gray!60}{Benign} & \textcolor{gray!60}{Benign} & \textcolor{gray!60}{Benign} & 99.2\% & \textcolor{gray!30}{$\varnothing$} & 13.8 & \textcolor{gray!30}{$\varnothing$} & 85.4\% & \textcolor{gray!30}{$\varnothing$} & 14.4\% & \textcolor{gray!30}{$\varnothing$} & \textcolor{gray!30}{$\varnothing$} \\
    \hline
    \hlc[Slate]{Landmark Shift Attack} \hlc[CadetBlueLight]{BadNets} $\alpha$=0.5 & \textcolor{gray!60}{Benign} & \textcolor{gray!60}{Benign} & 99.5\% & 99.6\% & 14.2 & 150.2 & 87.0\% & 0.6\% & 13.3\% & 59.8\% & 0.4\% \\
    \hlc[Slate]{Landmark Shift Attack} \hlc[CadetBlueLight]{BadNets} $\alpha$=1.0 & \textcolor{gray!60}{Benign} & \textcolor{gray!60}{Benign} & 99.5\% & 99.5\% & 14.7 & 147.7 & 87.7\% & 0.9\% & 13.7\% & 68.2\% & 0.6\% \\
    \hlc[Slate]{Landmark Shift Attack} \hlc[OrchidLight]{SIG} $\alpha$=0.16 & \textcolor{gray!60}{Benign} & \textcolor{gray!60}{Benign} & 99.5\% & 99.5\% & 14.9 & 20.2 & 86.9\% & 22.3\% & 14.6\% & 26.1\% & \textbf{5.8\%} \\
    \hlc[Slate]{Landmark Shift Attack} \hlc[OrchidLight]{SIG} $\alpha$=0.3 & \textcolor{gray!60}{Benign} & \textcolor{gray!60}{Benign} & 99.4\% & 97.4\% & 14.7 & 125.7 & 85.0\% & 1.6\% & 13.5\% & 65.9\% & 1.0\% \\
\end{tabular}
\end{table*}

\begin{table*}
  \centering
  \caption{\acl{ba} \acl{sr} in an \textbf{\acl{a2o}} threat model context, targeting a pipeline consisting of:\\ResNet50, AENet, GhostFaceNetV2.\\\textbf{Note}: FMR$^\pois$ metrics in \textcolor{BrickRed}{red} indicate a DNN with a collapsed performance after inclusion in the \ac{frs} (\ie, all identities match as in an untargeted poisoning attack). \textbf{Note}: The Landmark Shift Attack SIG $\alpha=0.16$ model did not converge and is not reported.}
  \label{tab:backdoor_bench_RN50_AE_GFN}
  \vspace{0.0cm}
  \footnotesize
  \setlength{\tabcolsep}{1.55pt}
  \renewcommand{\arraystretch}{1.5}
  \begin{tabular}{@{}lllccccccccc@{}}
    
     &  &  & \multicolumn{4}{c}{\textbf{Detector}} & \multicolumn{2}{c}{\textbf{Antispoofer}} & \multicolumn{2}{c}{\textbf{Extractor}} & \\
     \textbf{Detector} & \textbf{Antispoofer} & \textbf{Extractor} & \multicolumn{4}{c}{\textbf{metrics}} & \multicolumn{2}{c}{\textbf{metrics}} & \multicolumn{2}{c}{\textbf{metrics}} & \textbf{Survival} \\
     ResNet50 & AENet & GhostFaceNetV2 & AP$^\clean$ & AP$^\pois$ & LS$^\clean$ & LS$^\pois$ & FRR$^\clean$ & FAR$^\pois$ & FRR$^\clean$ & FAR$^\pois$ & \textbf{Rate} \\
    \hline 
   \textcolor{gray!60}{Benign} & \textcolor{gray!60}{Benign} & \textcolor{gray!60}{Benign} & 99.6\% & \textcolor{gray!30}{$\varnothing$} & 16.6 & \textcolor{gray!30}{$\varnothing$} & 94.4\% & \textcolor{gray!30}{$\varnothing$} & 4.4\% & \textcolor{gray!30}{$\varnothing$} & \textcolor{gray!30}{$\varnothing$} \\
    \hline
    \hlc[Slate]{Landmark Shift Attack} \hlc[CadetBlueLight]{BadNets} $\alpha$=0.5 & \textcolor{gray!60}{Benign} & \textcolor{gray!60}{Benign} & 99.5\% & 99.5\% & 12.0 & 153.4 & 95.2\% & 34.4\% & 4.5\% & 44.0\% & 15.1\% \\
    \hlc[Slate]{Landmark Shift Attack} \hlc[CadetBlueLight]{BadNets} $\alpha$=1.0 & \textcolor{gray!60}{Benign} & \textcolor{gray!60}{Benign} & 99.5\% & 99.6\% & 12.1 & 143.4 & 95.3\% & 33.0\% & 4.3\% & 39.8\% & 13.1\% \\
    \hlc[Slate]{Landmark Shift Attack} \hlc[OrchidLight]{SIG} $\alpha$=0.3 & \textcolor{gray!60}{Benign} & \textcolor{gray!60}{Benign} & 99.4\% & 96.8\% & 12.6 & 156.1 & 95.6\% & 96.1\% & 4.6\% & 87.8\% & \textbf{81.7\%} \\
    \hline\hlc[Wheat]{Face Generation Attack} \hlc[CadetBlueLight]{BadNets} $\alpha$=0.5 & \textcolor{gray!60}{Benign} & \textcolor{gray!60}{Benign} & 99.5\% & 99.5\% & 12.0 & 2.8 & 95.5\% & 61.8\% & 4.4\% & 99.5\% & 61.2\% \\
    \hlc[Wheat]{Face Generation Attack} \hlc[CadetBlueLight]{BadNets} $\alpha$=1.0 & \textcolor{gray!60}{Benign} & \textcolor{gray!60}{Benign} & 99.5\% & 99.8\% & 12.2 & 1.9 & 95.4\% & 41.9\% & 4.3\% & 99.6\% & 41.6\% \\
    \hlc[Wheat]{Face Generation Attack} \hlc[OrchidLight]{SIG} $\alpha$=0.16 & \textcolor{gray!60}{Benign} & \textcolor{gray!60}{Benign} & 99.5\% & 92.2\% & 11.9 & 30.4 & 95.4\% & 72.2\% & 4.5\% & 92.6\% & 61.6\% \\
    \hlc[Wheat]{Face Generation Attack} \hlc[OrchidLight]{SIG} $\alpha$=0.3 & \textcolor{gray!60}{Benign} & \textcolor{gray!60}{Benign} & 99.5\% & 98.0\% & 12.6 & 9.3 & 95.2\% & 74.0\% & 4.6\% & 98.9\% & \textbf{71.7\%} \\
    \hline
    \textcolor{gray!60}{Benign} & \hlc[DandelionLight]{Glasses} & \textcolor{gray!60}{Benign} & 99.6\% & 97.9\% & 16.6 & 22.9 & 20.9\% & 86.4\% & 4.5\% & 46.4\% & 39.2\% \\
    \textcolor{gray!60}{Benign} & \hlc[CadetBlueLight]{BadNets} & \textcolor{gray!60}{Benign} & 99.6\% & 99.5\% & 16.6 & 16.6 & 18.3\% & 40.9\% & 4.6\% & 4.5\% & 1.8\% \\
    \textcolor{gray!60}{Benign} & \hlc[OrchidLight]{SIG} & \textcolor{gray!60}{Benign} & 99.6\% & 98.4\% & 16.6 & 24.1 & 12.9\% & 97.2\% & 4.8\% & 62.8\% & \textbf{60.1\%} \\
    \textcolor{gray!60}{Benign} & \hlc[MelonLight]{TrojanNN} & \textcolor{gray!60}{Benign} & 99.6\% & 99.5\% & 16.6 & 16.7 & 8.0\% & 7.2\% & 5.2\% & 4.6\% & 0.3\% \\
    \hline
    \textcolor{gray!60}{Benign} & \textcolor{gray!60}{Benign} & \hlc[RoyalPurpleLight]{FIBA} & 99.6\% & 98.3\% & 16.6 & 26.7 & 94.4\% & 24.6\% & 4.4\% & 96.1\% & \textbf{23.2\%} \\
    \hline\textcolor{gray!60}{Benign} & \textcolor{gray!60}{Benign} & \hlc[GoldenRodLight]{All-to-One}, \hlc[ApricotLight]{Clean-Label}, \hlc[CadetBlueLight]{BadNets} & 99.6\% & 99.5\% & 16.6 & 16.9 & 94.4\% & 18.8\% & 7.7\% & 8.0\% & 1.5\% \\
    \textcolor{gray!60}{Benign} & \textcolor{gray!60}{Benign} & \hlc[GoldenRodLight]{All-to-One}, \hlc[ForestLight]{Poison-Label}, \hlc[CadetBlueLight]{BadNets} & 99.6\% & 99.5\% & 16.6 & 16.9 & 94.4\% & 18.8\% & 4.0\% & 97.9\% & 18.3\% \\
    \textcolor{gray!60}{Benign} & \textcolor{gray!60}{Benign} & \hlc[NavyLight]{Master Face}, \hlc[ApricotLight]{Clean-Label}, \hlc[CadetBlueLight]{BadNets} & 99.6\% & 99.5\% & 16.6 & 16.9 & 94.4\% & 18.8\% & 4.1\% & 4.2\% & 0.8\% \\
    \textcolor{gray!60}{Benign} & \textcolor{gray!60}{Benign} & \hlc[NavyLight]{Master Face}, \hlc[ForestLight]{Poison-Label}, \hlc[CadetBlueLight]{BadNets} & 99.2\% & 99.0\% & 13.8 & 14.2 & 85.4\% & 21.4\% & \textcolor{BrickRed}{99.1\%} & \textcolor{BrickRed}{99.1\%} & \textbf{21.0\%} \\
    \hline\textcolor{gray!60}{Benign} & \textcolor{gray!60}{Benign} & \hlc[GoldenRodLight]{All-to-One}, \hlc[ApricotLight]{Clean-Label}, \hlc[OrchidLight]{SIG} & 99.6\% & 98.4\% & 16.6 & 24.1 & 94.4\% & 81.4\% & 4.7\% & 73.5\% & 58.9\% \\
    \textcolor{gray!60}{Benign} & \textcolor{gray!60}{Benign} & \hlc[GoldenRodLight]{All-to-One}, \hlc[ForestLight]{Poison-Label}, \hlc[OrchidLight]{SIG} & 99.6\% & 98.4\% & 16.6 & 24.1 & 94.4\% & 81.4\% & 5.9\% & 97.8\% & \textbf{78.3\%} \\
    \textcolor{gray!60}{Benign} & \textcolor{gray!60}{Benign} & \hlc[NavyLight]{Master Face}, \hlc[ApricotLight]{Clean-Label}, \hlc[OrchidLight]{SIG} & 99.6\% & 98.4\% & 16.6 & 24.1 & 94.4\% & 81.4\% & 5.3\% & 20.0\% & 16.0\% \\
    \textcolor{gray!60}{Benign} & \textcolor{gray!60}{Benign} & \hlc[NavyLight]{Master Face}, \hlc[ForestLight]{Poison-Label}, \hlc[OrchidLight]{SIG} & 99.2\% & 94.2\% & 13.8 & 23.2 & 85.4\% & 75.2\% & 4.9\% & 96.5\% & 68.4\% \\
    
  \end{tabular}
\end{table*}

\begin{table*}
  \centering
  \caption{\acl{ba} \acl{sr} in an \textbf{\acl{a2o}} threat model context, targeting a pipeline consisting of:\\ResNet50, AENet, IRSE50.\\\textbf{Note}: The Landmark Shift Attack SIG $\alpha=0.16$ model did not converge and is not reported.}
  \label{tab:backdoor_bench_RN50_AE_IRSE}
  \vspace{0.0cm}
  \footnotesize
  \setlength{\tabcolsep}{1.55pt}
  \renewcommand{\arraystretch}{1.5}
  \begin{tabular}{@{}lllccccccccc@{}}
    
     &  &  & \multicolumn{4}{c}{\textbf{Detector}} & \multicolumn{2}{c}{\textbf{Antispoofer}} & \multicolumn{2}{c}{\textbf{Extractor}} & \\
     \textbf{Detector} & \textbf{Antispoofer} & \textbf{Extractor} & \multicolumn{4}{c}{\textbf{metrics}} & \multicolumn{2}{c}{\textbf{metrics}} & \multicolumn{2}{c}{\textbf{metrics}} & \textbf{Survival} \\
     ResNet50 & AENet & IRSE50 & AP$^\clean$ & AP$^\pois$ & LS$^\clean$ & LS$^\pois$ & FRR$^\clean$ & FAR$^\pois$ & FRR$^\clean$ & FAR$^\pois$ & \textbf{Rate} \\
    \hline 
    \textcolor{gray!60}{Benign} & \textcolor{gray!60}{Benign} & \textcolor{gray!60}{Benign} & 99.6\% & \textcolor{gray!30}{$\varnothing$} & 16.6 & \textcolor{gray!30}{$\varnothing$} & 94.4\% & \textcolor{gray!30}{$\varnothing$} & 0.8\% & \textcolor{gray!30}{$\varnothing$} & \textcolor{gray!30}{$\varnothing$} \\
    \hline
    \hlc[Slate]{Landmark Shift Attack} \hlc[CadetBlueLight]{BadNets} $\alpha$=0.5 & \textcolor{gray!60}{Benign} & \textcolor{gray!60}{Benign} & 99.5\% & 99.5\% & 12.0 & 153.4 & 95.2\% & 34.4\% & 0.7\% & 56.9\% & 19.5\% \\
    \hlc[Slate]{Landmark Shift Attack} \hlc[CadetBlueLight]{BadNets} $\alpha$=1.0 & \textcolor{gray!60}{Benign} & \textcolor{gray!60}{Benign} & 99.5\% & 99.6\% & 12.1 & 143.4 & 95.3\% & 33.0\% & 0.7\% & 42.0\% & 13.8\% \\
    \hlc[Slate]{Landmark Shift Attack} \hlc[OrchidLight]{SIG} $\alpha$=0.3 & \textcolor{gray!60}{Benign} & \textcolor{gray!60}{Benign} & 99.4\% & 96.8\% & 12.6 & 156.1 & 95.6\% & 96.1\% & 0.8\% & 96.9\% & \textbf{90.1\%} \\
    \hline\hlc[Wheat]{Face Generation Attack} \hlc[CadetBlueLight]{BadNets} $\alpha$=0.5 & \textcolor{gray!60}{Benign} & \textcolor{gray!60}{Benign} & 99.5\% & 99.5\% & 12.0 & 2.7 & 95.5\% & 62.8\% & 0.7\% & 99.4\% & 62.1\% \\
    \hlc[Wheat]{Face Generation Attack} \hlc[CadetBlueLight]{BadNets} $\alpha$=1.0 & \textcolor{gray!60}{Benign} & \textcolor{gray!60}{Benign} & 99.5\% & 99.9\% & 12.2 & 1.8 & 95.4\% & 41.9\% & 0.7\% & 99.6\% & 41.7\% \\
    \hlc[Wheat]{Face Generation Attack} \hlc[OrchidLight]{SIG} $\alpha$=0.16 & \textcolor{gray!60}{Benign} & \textcolor{gray!60}{Benign} & 99.5\% & 92.6\% & 11.9 & 29.7 & 95.4\% & 73.1\% & 0.7\% & 89.3\% & 60.4\% \\
    \hlc[Wheat]{Face Generation Attack} \hlc[OrchidLight]{SIG} $\alpha$=0.3 & \textcolor{gray!60}{Benign} & \textcolor{gray!60}{Benign} & 99.5\% & 98.0\% & 12.6 & 8.8 & 95.2\% & 74.0\% & 0.8\% & 98.7\% & \textbf{71.6\%} \\
    \hline
    \textcolor{gray!60}{Benign} & \hlc[DandelionLight]{Glasses} & \textcolor{gray!60}{Benign} & 99.6\% & 97.9\% & 16.6 & 22.9 & 20.9\% & 86.4\% & 0.7\% & 25.9\% & \textbf{21.9\%} \\
    \textcolor{gray!60}{Benign} & \hlc[CadetBlueLight]{BadNets} & \textcolor{gray!60}{Benign} & 99.6\% & 99.5\% & 16.6 & 16.6 & 18.3\% & 40.9\% & 0.7\% & 0.6\% & 0.2\% \\
    \textcolor{gray!60}{Benign} & \hlc[OrchidLight]{SIG} & \textcolor{gray!60}{Benign} & 99.6\% & 98.4\% & 16.6 & 24.1 & 12.9\% & 97.2\% & 0.7\% & 21.4\% & 20.5\% \\
    \textcolor{gray!60}{Benign} & \hlc[MelonLight]{TrojanNN} & \textcolor{gray!60}{Benign} & 99.6\% & 99.5\% & 16.6 & 16.7 & 8.0\% & 6.8\% & 0.8\% & 0.7\% & 0.0\% \\
    \hline
    \textcolor{gray!60}{Benign} & \textcolor{gray!60}{Benign} & \hlc[RoyalPurpleLight]{FIBA} & 99.6\% & 98.3\% & 16.6 & 26.7 & 94.4\% & 24.6\% & 0.8\% & 92.2\% & \textbf{22.3\%} \\
    \hline\textcolor{gray!60}{Benign} & \textcolor{gray!60}{Benign} & \hlc[GoldenRodLight]{All-to-One}, \hlc[ApricotLight]{Clean-Label}, \hlc[CadetBlueLight]{BadNets} & 99.6\% & 99.5\% & 16.6 & 16.9 & 94.4\% & 18.8\% & 34.3\% & 39.2\% & 7.3\% \\
    \textcolor{gray!60}{Benign} & \textcolor{gray!60}{Benign} & \hlc[GoldenRodLight]{All-to-One}, \hlc[ForestLight]{Poison-Label}, \hlc[CadetBlueLight]{BadNets} & 99.6\% & 99.5\% & 16.6 & 16.9 & 94.4\% & 18.8\% & 0.8\% & 94.8\% & \textbf{17.7\%} \\
    \textcolor{gray!60}{Benign} & \textcolor{gray!60}{Benign} & \hlc[NavyLight]{Master Face}, \hlc[ApricotLight]{Clean-Label}, \hlc[CadetBlueLight]{BadNets} & 99.6\% & 99.5\% & 16.6 & 16.9 & 94.4\% & 18.8\% & 0.5\% & 0.5\% & 0.1\% \\
    \textcolor{gray!60}{Benign} & \textcolor{gray!60}{Benign} & \hlc[NavyLight]{Master Face}, \hlc[ForestLight]{Poison-Label}, \hlc[CadetBlueLight]{BadNets} & 99.6\% & 99.5\% & 16.6 & 16.9 & 94.4\% & 18.8\% & 0.6\% & 0.6\% & 0.1\% \\
    \hline\textcolor{gray!60}{Benign} & \textcolor{gray!60}{Benign} & \hlc[GoldenRodLight]{All-to-One}, \hlc[ApricotLight]{Clean-Label}, \hlc[OrchidLight]{SIG} & 99.6\% & 98.4\% & 16.6 & 24.1 & 94.4\% & 81.4\% & 0.6\% & 34.7\% & 27.8\% \\
    \textcolor{gray!60}{Benign} & \textcolor{gray!60}{Benign} & \hlc[GoldenRodLight]{All-to-One}, \hlc[ForestLight]{Poison-Label}, \hlc[OrchidLight]{SIG} & 99.6\% & 98.4\% & 16.6 & 24.1 & 94.4\% & 81.4\% & 0.6\% & 48.8\% & 39.1\% \\
    \textcolor{gray!60}{Benign} & \textcolor{gray!60}{Benign} & \hlc[NavyLight]{Master Face}, \hlc[ApricotLight]{Clean-Label}, \hlc[OrchidLight]{SIG} & 99.6\% & 98.4\% & 16.6 & 24.1 & 94.4\% & 81.4\% & 0.5\% & 12.1\% & 9.7\% \\
    \textcolor{gray!60}{Benign} & \textcolor{gray!60}{Benign} & \hlc[NavyLight]{Master Face}, \hlc[ForestLight]{Poison-Label}, \hlc[OrchidLight]{SIG} & 99.6\% & 98.4\% & 16.6 & 24.1 & 94.4\% & 81.4\% & 0.6\% & 89.3\% & \textbf{71.5\%} \\
    
  \end{tabular}
\end{table*}

\begin{table*}
  \centering
  \caption{\acl{ba} \acl{sr} in an \textbf{\acl{a2o}} threat model context, targeting a pipeline consisting of:\\ResNet50, AENet, MobileFaceNet.\\\textbf{Note}: The Landmark Shift Attack SIG $\alpha=0.16$ model did not converge and is not reported.}
  \label{tab:backdoor_bench_RN50_AE_MFN}
  \vspace{0.0cm}
  \footnotesize
  \setlength{\tabcolsep}{1.55pt}
  \renewcommand{\arraystretch}{1.5}
  \begin{tabular}{@{}lllccccccccc@{}}
    
     &  &  & \multicolumn{4}{c}{\textbf{Detector}} & \multicolumn{2}{c}{\textbf{Antispoofer}} & \multicolumn{2}{c}{\textbf{Extractor}} & \\
     \textbf{Detector} & \textbf{Antispoofer} & \textbf{Extractor} & \multicolumn{4}{c}{\textbf{metrics}} & \multicolumn{2}{c}{\textbf{metrics}} & \multicolumn{2}{c}{\textbf{metrics}} & \textbf{Survival} \\
     ResNet50 & AENet & MobileFaceNet & AP$^\clean$ & AP$^\pois$ & LS$^\clean$ & LS$^\pois$ & FRR$^\clean$ & FAR$^\pois$ & FRR$^\clean$ & FAR$^\pois$ & \textbf{Rate} \\
    \hline 
    \textcolor{gray!60}{Benign} & \textcolor{gray!60}{Benign} & \textcolor{gray!60}{Benign} & 99.6\% & \textcolor{gray!30}{$\varnothing$} & 16.6 & \textcolor{gray!30}{$\varnothing$} & 94.4\% & \textcolor{gray!30}{$\varnothing$} & 3.8\% & \textcolor{gray!30}{$\varnothing$} & \textcolor{gray!30}{$\varnothing$} \\
    \hline
    \hlc[Slate]{Landmark Shift Attack} \hlc[CadetBlueLight]{BadNets} $\alpha$=0.5 & \textcolor{gray!60}{Benign} & \textcolor{gray!60}{Benign} & 99.5\% & 99.5\% & 12.0 & 153.4 & 95.2\% & 34.4\% & 3.8\% & 92.4\% & 31.6\% \\
    \hlc[Slate]{Landmark Shift Attack} \hlc[CadetBlueLight]{BadNets} $\alpha$=1.0 & \textcolor{gray!60}{Benign} & \textcolor{gray!60}{Benign} & 99.5\% & 99.6\% & 12.1 & 143.4 & 95.3\% & 33.0\% & 3.6\% & 84.6\% & 27.8\% \\
    \hlc[Slate]{Landmark Shift Attack} \hlc[OrchidLight]{SIG} $\alpha$=0.3 & \textcolor{gray!60}{Benign} & \textcolor{gray!60}{Benign} & 99.4\% & 96.8\% & 12.6 & 156.1 & 95.6\% & 96.1\% & 3.8\% & 97.4\% & \textbf{90.6\%} \\
    \hline\hlc[Wheat]{Face Generation Attack} \hlc[CadetBlueLight]{BadNets} $\alpha$=0.5 & \textcolor{gray!60}{Benign} & \textcolor{gray!60}{Benign} & 99.5\% & 99.6\% & 12.0 & 2.6 & 95.5\% & 62.9\% & 3.7\% & 99.5\% & 62.3\% \\
    \hlc[Wheat]{Face Generation Attack} \hlc[CadetBlueLight]{BadNets} $\alpha$=1.0 & \textcolor{gray!60}{Benign} & \textcolor{gray!60}{Benign} & 99.5\% & 99.9\% & 12.2 & 1.8 & 95.4\% & 42.0\% & 3.6\% & 99.5\% & 41.7\% \\
    \hlc[Wheat]{Face Generation Attack} \hlc[OrchidLight]{SIG} $\alpha$=0.16 & \textcolor{gray!60}{Benign} & \textcolor{gray!60}{Benign} & 99.5\% & 91.6\% & 11.9 & 30.8 & 95.4\% & 72.5\% & 3.7\% & 92.0\% & 61.1\% \\
    \hlc[Wheat]{Face Generation Attack} \hlc[OrchidLight]{SIG} $\alpha$=0.3 & \textcolor{gray!60}{Benign} & \textcolor{gray!60}{Benign} & 99.5\% & 97.9\% & 12.6 & 8.9 & 95.2\% & 72.8\% & 4.1\% & 98.3\% & \textbf{70.1\%} \\
    \hline
    \textcolor{gray!60}{Benign} & \hlc[DandelionLight]{Glasses} & \textcolor{gray!60}{Benign} & 99.6\% & 97.9\% & 16.6 & 22.9 & 20.9\% & 86.4\% & 3.5\% & 60.9\% & 51.5\% \\
    \textcolor{gray!60}{Benign} & \hlc[CadetBlueLight]{BadNets} & \textcolor{gray!60}{Benign} & 99.6\% & 99.5\% & 16.6 & 16.6 & 18.3\% & 40.9\% & 3.4\% & 2.9\% & 1.2\% \\
    \textcolor{gray!60}{Benign} & \hlc[OrchidLight]{SIG} & \textcolor{gray!60}{Benign} & 99.6\% & 98.4\% & 16.6 & 24.1 & 12.9\% & 97.2\% & 3.6\% & 88.8\% & \textbf{84.9\%} \\
    \textcolor{gray!60}{Benign} & \hlc[MelonLight]{TrojanNN} & \textcolor{gray!60}{Benign} & 99.6\% & 99.5\% & 16.6 & 16.6 & 8.0\% & 6.5\% & 4.0\% & 3.6\% & 0.2\% \\
    \hline
    \textcolor{gray!60}{Benign} & \textcolor{gray!60}{Benign} & \hlc[RoyalPurpleLight]{FIBA} & 99.6\% & 98.3\% & 16.6 & 26.7 & 94.4\% & 24.6\% & 3.8\% & 98.4\% & \textbf{23.8\%} \\
    \hline\textcolor{gray!60}{Benign} & \textcolor{gray!60}{Benign} & \hlc[GoldenRodLight]{All-to-One}, \hlc[ApricotLight]{Clean-Label}, \hlc[CadetBlueLight]{BadNets} & 99.6\% & 99.5\% & 16.6 & 16.9 & 94.4\% & 18.8\% & 3.9\% & 4.2\% & 0.8\% \\
    \textcolor{gray!60}{Benign} & \textcolor{gray!60}{Benign} & \hlc[GoldenRodLight]{All-to-One}, \hlc[ForestLight]{Poison-Label}, \hlc[CadetBlueLight]{BadNets} & 99.6\% & 99.5\% & 16.6 & 16.9 & 94.4\% & 18.8\% & 2.5\% & 97.8\% & \textbf{18.3\%} \\
    \textcolor{gray!60}{Benign} & \textcolor{gray!60}{Benign} & \hlc[NavyLight]{Master Face}, \hlc[ApricotLight]{Clean-Label}, \hlc[CadetBlueLight]{BadNets} & 99.6\% & 99.5\% & 16.6 & 16.9 & 94.4\% & 18.8\% & 3.2\% & 3.4\% & 0.6\% \\
    \textcolor{gray!60}{Benign} & \textcolor{gray!60}{Benign} & \hlc[NavyLight]{Master Face}, \hlc[ForestLight]{Poison-Label}, \hlc[CadetBlueLight]{BadNets} & 99.6\% & 99.5\% & 16.6 & 16.9 & 94.4\% & 18.8\% & 2.3\% & 2.9\% & 0.5\% \\
    \hline\textcolor{gray!60}{Benign} & \textcolor{gray!60}{Benign} & \hlc[GoldenRodLight]{All-to-One}, \hlc[ApricotLight]{Clean-Label}, \hlc[RedVioletLight]{Mask} & 99.6\% & 99.2\% & 16.6 & 29.7 & 94.4\% & 21.5\% & 2.8\% & 98.3\% & \textbf{21.0\%} \\
    \textcolor{gray!60}{Benign} & \textcolor{gray!60}{Benign} & \hlc[GoldenRodLight]{All-to-One}, \hlc[ForestLight]{Poison-Label}, \hlc[RedVioletLight]{Mask} & 99.6\% & 99.2\% & 16.6 & 29.7 & 94.4\% & 21.5\% & 2.8\% & 98.6\% & \textbf{21.0\%} \\
    \textcolor{gray!60}{Benign} & \textcolor{gray!60}{Benign} & \hlc[NavyLight]{Master Face}, \hlc[ApricotLight]{Clean-Label}, \hlc[RedVioletLight]{Mask} & 99.6\% & 99.2\% & 16.6 & 29.7 & 94.4\% & 21.5\% & 2.6\% & 20.2\% & 4.3\% \\
    \textcolor{gray!60}{Benign} & \textcolor{gray!60}{Benign} & \hlc[NavyLight]{Master Face}, \hlc[ForestLight]{Poison-Label}, \hlc[RedVioletLight]{Mask} & 99.6\% & 99.2\% & 16.6 & 29.7 & 94.4\% & 21.5\% & 3.2\% & 58.5\% & 12.5\% \\
    \hline\textcolor{gray!60}{Benign} & \textcolor{gray!60}{Benign} & \hlc[GoldenRodLight]{All-to-One}, \hlc[ApricotLight]{Clean-Label}, \hlc[OrchidLight]{SIG} & 99.6\% & 98.4\% & 16.6 & 24.1 & 94.4\% & 81.4\% & 3.3\% & 93.4\% & \textbf{74.8\%} \\
    \textcolor{gray!60}{Benign} & \textcolor{gray!60}{Benign} & \hlc[GoldenRodLight]{All-to-One}, \hlc[ForestLight]{Poison-Label}, \hlc[OrchidLight]{SIG} & 99.6\% & 98.4\% & 16.6 & 24.1 & 94.4\% & 81.4\% & 3.0\% & 96.2\% & 77.1\% \\
    \textcolor{gray!60}{Benign} & \textcolor{gray!60}{Benign} & \hlc[NavyLight]{Master Face}, \hlc[ApricotLight]{Clean-Label}, \hlc[OrchidLight]{SIG} & 99.6\% & 98.4\% & 16.6 & 24.1 & 94.4\% & 81.4\% & 3.2\% & 69.9\% & 56.0\% \\
    \textcolor{gray!60}{Benign} & \textcolor{gray!60}{Benign} & \hlc[NavyLight]{Master Face}, \hlc[ForestLight]{Poison-Label}, \hlc[OrchidLight]{SIG} & 99.6\% & 98.4\% & 16.6 & 24.1 & 94.4\% & 81.4\% & 3.6\% & 52.6\% & 42.1\% \\
    
  \end{tabular}
  
\end{table*}

\begin{table*}
  \centering
  \caption{\acl{ba} \acl{sr} in an \textbf{\acl{a2o}} threat model context, targeting a pipeline consisting of:\\ResNet50, AENet, ResNet50.\\\textbf{Note}: The Landmark Shift Attack SIG $\alpha=0.16$ model did not converge and is not reported.}
  \label{tab:backdoor_bench_RN50_AE_RN50}
  \vspace{0.0cm}
  \footnotesize
  \setlength{\tabcolsep}{1.55pt}
  \renewcommand{\arraystretch}{1.5}
  \begin{tabular}{@{}lllccccccccc@{}}
    
     &  &  & \multicolumn{4}{c}{\textbf{Detector}} & \multicolumn{2}{c}{\textbf{Antispoofer}} & \multicolumn{2}{c}{\textbf{Extractor}} & \\
     \textbf{Detector} & \textbf{Antispoofer} & \textbf{Extractor} & \multicolumn{4}{c}{\textbf{metrics}} & \multicolumn{2}{c}{\textbf{metrics}} & \multicolumn{2}{c}{\textbf{metrics}} & \textbf{Survival} \\
     ResNet50 & AENet & ResNet50 & AP$^\clean$ & AP$^\pois$ & LS$^\clean$ & LS$^\pois$ & FRR$^\clean$ & FAR$^\pois$ & FRR$^\clean$ & FAR$^\pois$ & \textbf{Rate} \\
    \hline 
    \textcolor{gray!60}{Benign} & \textcolor{gray!60}{Benign} & \textcolor{gray!60}{Benign} & 99.6\% & \textcolor{gray!30}{$\varnothing$} & 16.6 & \textcolor{gray!30}{$\varnothing$} & 94.4\% & \textcolor{gray!30}{$\varnothing$} & 1.8\% & \textcolor{gray!30}{$\varnothing$} & \textcolor{gray!30}{$\varnothing$} \\
    \hline
    \hlc[Slate]{Landmark Shift Attack} \hlc[CadetBlueLight]{BadNets} $\alpha$=0.5 & \textcolor{gray!60}{Benign} & \textcolor{gray!60}{Benign} & 99.5\% & 99.5\% & 12.0 & 153.4 & 95.2\% & 34.4\% & 1.8\% & 41.7\% & 14.3\% \\
    \hlc[Slate]{Landmark Shift Attack} \hlc[CadetBlueLight]{BadNets} $\alpha$=1.0 & \textcolor{gray!60}{Benign} & \textcolor{gray!60}{Benign} & 99.5\% & 99.6\% & 12.1 & 143.4 & 95.3\% & 33.0\% & 1.7\% & 33.2\% & 10.9\% \\
    \hlc[Slate]{Landmark Shift Attack} \hlc[OrchidLight]{SIG} $\alpha$=0.3 & \textcolor{gray!60}{Benign} & \textcolor{gray!60}{Benign} & 99.4\% & 96.8\% & 12.6 & 156.1 & 95.6\% & 96.1\% & 1.9\% & 97.2\% & \textbf{90.4\%} \\
    \hline\hlc[Wheat]{Face Generation Attack} \hlc[CadetBlueLight]{BadNets} $\alpha$=0.5 & \textcolor{gray!60}{Benign} & \textcolor{gray!60}{Benign} & 99.5\% & 99.5\% & 12.0 & 3.0 & 95.5\% & 63.2\% & 1.8\% & 99.3\% & 62.4\% \\
    \hlc[Wheat]{Face Generation Attack} \hlc[CadetBlueLight]{BadNets} $\alpha$=1.0 & \textcolor{gray!60}{Benign} & \textcolor{gray!60}{Benign} & 99.5\% & 99.9\% & 12.2 & 1.8 & 95.4\% & 41.6\% & 1.8\% & 99.8\% & 41.5\% \\
    \hlc[Wheat]{Face Generation Attack} \hlc[OrchidLight]{SIG} $\alpha$=0.16 & \textcolor{gray!60}{Benign} & \textcolor{gray!60}{Benign} & 99.5\% & 92.2\% & 11.9 & 29.5 & 95.4\% & 71.9\% & 1.8\% & 91.2\% & 60.5\% \\
    \hlc[Wheat]{Face Generation Attack} \hlc[OrchidLight]{SIG} $\alpha$=0.3 & \textcolor{gray!60}{Benign} & \textcolor{gray!60}{Benign} & 99.5\% & 98.0\% & 12.6 & 9.0 & 95.2\% & 73.9\% & 1.9\% & 98.6\% & \textbf{71.4\%} \\
    \hline
    \textcolor{gray!60}{Benign} & \hlc[DandelionLight]{Glasses} & \textcolor{gray!60}{Benign} & 99.6\% & 97.9\% & 16.6 & 22.9 & 20.9\% & 86.4\% & 1.4\% & 29.9\% & 25.3\% \\
    \textcolor{gray!60}{Benign} & \hlc[CadetBlueLight]{BadNets} & \textcolor{gray!60}{Benign} & 99.6\% & 99.5\% & 16.6 & 16.6 & 18.3\% & 40.9\% & 1.4\% & 1.3\% & 0.5\% \\
    \textcolor{gray!60}{Benign} & \hlc[OrchidLight]{SIG} & \textcolor{gray!60}{Benign} & 99.6\% & 98.4\% & 16.6 & 24.1 & 12.9\% & 97.2\% & 1.4\% & 62.5\% & \textbf{59.8\%} \\
    \textcolor{gray!60}{Benign} & \hlc[MelonLight]{TrojanNN} & \textcolor{gray!60}{Benign} & 99.6\% & 99.5\% & 16.6 & 16.6 & 8.0\% & 6.5\% & 1.6\% & 1.4\% & 0.1\% \\
    \hline
    \textcolor{gray!60}{Benign} & \textcolor{gray!60}{Benign} & \hlc[RoyalPurpleLight]{FIBA} & 99.6\% & 98.3\% & 16.6 & 26.7 & 94.4\% & 24.6\% & 1.8\% & 96.1\% & \textbf{23.2\%} \\
    \hline\textcolor{gray!60}{Benign} & \textcolor{gray!60}{Benign} & \hlc[GoldenRodLight]{All-to-One}, \hlc[ApricotLight]{Clean-Label}, \hlc[CadetBlueLight]{BadNets} & 99.6\% & 99.5\% & 16.6 & 16.9 & 94.4\% & 18.8\% & 1.6\% & 1.4\% & 0.3\% \\
    \textcolor{gray!60}{Benign} & \textcolor{gray!60}{Benign} & \hlc[GoldenRodLight]{All-to-One}, \hlc[ForestLight]{Poison-Label}, \hlc[CadetBlueLight]{BadNets} & 99.6\% & 99.5\% & 16.6 & 16.9 & 94.4\% & 18.8\% & 2.5\% & 97.0\% & \textbf{18.1\%} \\
    \textcolor{gray!60}{Benign} & \textcolor{gray!60}{Benign} & \hlc[NavyLight]{Master Face}, \hlc[ApricotLight]{Clean-Label}, \hlc[CadetBlueLight]{BadNets} & 99.6\% & 99.5\% & 16.6 & 16.9 & 94.4\% & 18.8\% & 1.9\% & 1.8\% & 0.3\% \\
    \textcolor{gray!60}{Benign} & \textcolor{gray!60}{Benign} & \hlc[NavyLight]{Master Face}, \hlc[ForestLight]{Poison-Label}, \hlc[CadetBlueLight]{BadNets} & 99.6\% & 99.5\% & 16.6 & 16.9 & 94.4\% & 18.8\% & 1.9\% & 1.8\% & 0.3\% \\
    \hline\textcolor{gray!60}{Benign} & \textcolor{gray!60}{Benign} & \hlc[GoldenRodLight]{All-to-One}, \hlc[ApricotLight]{Clean-Label}, \hlc[OrchidLight]{SIG} & 99.6\% & 98.4\% & 16.6 & 24.1 & 94.4\% & 81.4\% & 1.8\% & 74.0\% & 59.3\% \\
    \textcolor{gray!60}{Benign} & \textcolor{gray!60}{Benign} & \hlc[GoldenRodLight]{All-to-One}, \hlc[ForestLight]{Poison-Label}, \hlc[OrchidLight]{SIG} & 99.6\% & 98.4\% & 16.6 & 24.1 & 94.4\% & 81.4\% & 1.5\% & 96.8\% & 77.5\% \\
    \textcolor{gray!60}{Benign} & \textcolor{gray!60}{Benign} & \hlc[NavyLight]{Master Face}, \hlc[ApricotLight]{Clean-Label}, \hlc[OrchidLight]{SIG} & 99.6\% & 98.4\% & 16.6 & 24.1 & 94.4\% & 81.4\% & 1.5\% & 38.2\% & 30.6\% \\
    \textcolor{gray!60}{Benign} & \textcolor{gray!60}{Benign} & \hlc[NavyLight]{Master Face}, \hlc[ForestLight]{Poison-Label}, \hlc[OrchidLight]{SIG} & 99.6\% & 98.4\% & 16.6 & 24.1 & 94.4\% & 81.4\% & 1.8\% & 97.1\% & \textbf{77.8\%} \\
    
  \end{tabular}
\end{table*}

\begin{table*}
  \centering
  \caption{\acl{ba} \acl{sr} in an \textbf{\acl{a2o}} threat model context, targeting a pipeline consisting of:\\ResNet50, AENet, RobFaceNet.\\\textbf{Note}: The Landmark Shift Attack SIG $\alpha=0.16$ model did not converge and is not reported.}
  \label{tab:backdoor_bench_RN50_AE_RFN}
  \vspace{0.0cm}
  \footnotesize
  \setlength{\tabcolsep}{1.55pt}
  \renewcommand{\arraystretch}{1.5}
  \begin{tabular}{@{}lllccccccccc@{}}
    
     &  &  & \multicolumn{4}{c}{\textbf{Detector}} & \multicolumn{2}{c}{\textbf{Antispoofer}} & \multicolumn{2}{c}{\textbf{Extractor}} & \\
     \textbf{Detector} & \textbf{Antispoofer} & \textbf{Extractor} & \multicolumn{4}{c}{\textbf{metrics}} & \multicolumn{2}{c}{\textbf{metrics}} & \multicolumn{2}{c}{\textbf{metrics}} & \textbf{Survival} \\
     ResNet50 & AENet & RobFaceNet & AP$^\clean$ & AP$^\pois$ & LS$^\clean$ & LS$^\pois$ & FRR$^\clean$ & FAR$^\pois$ & FRR$^\clean$ & FAR$^\pois$ & \textbf{Rate} \\
    \hline 
    \textcolor{gray!60}{Benign} & \textcolor{gray!60}{Benign} & \textcolor{gray!60}{Benign} & 99.6\% & \textcolor{gray!30}{$\varnothing$} & 16.6 & \textcolor{gray!30}{$\varnothing$} & 94.4\% & \textcolor{gray!30}{$\varnothing$} & 13.3\% & \textcolor{gray!30}{$\varnothing$} & \textcolor{gray!30}{$\varnothing$} \\
    \hline
    \hlc[Slate]{Landmark Shift Attack} \hlc[CadetBlueLight]{BadNets} $\alpha$=0.5 & \textcolor{gray!60}{Benign} & \textcolor{gray!60}{Benign} & 99.5\% & 99.5\% & 12.0 & 153.4 & 95.2\% & 34.4\% & 14.2\% & 93.4\% & 32.0\% \\
    \hlc[Slate]{Landmark Shift Attack} \hlc[CadetBlueLight]{BadNets} $\alpha$=1.0 & \textcolor{gray!60}{Benign} & \textcolor{gray!60}{Benign} & 99.5\% & 99.6\% & 12.1 & 143.4 & 95.3\% & 33.0\% & 13.5\% & 91.5\% & 30.1\% \\
    \hlc[Slate]{Landmark Shift Attack} \hlc[OrchidLight]{SIG} $\alpha$=0.3 & \textcolor{gray!60}{Benign} & \textcolor{gray!60}{Benign} & 99.4\% & 96.8\% & 12.6 & 156.1 & 95.6\% & 96.1\% & 15.1\% & 98.2\% & \textbf{91.4\%} \\
    \hline\hlc[Wheat]{Face Generation Attack} \hlc[CadetBlueLight]{BadNets} $\alpha$=0.5 & \textcolor{gray!60}{Benign} & \textcolor{gray!60}{Benign} & 99.5\% & 99.5\% & 12.0 & 2.9 & 95.5\% & 63.5\% & 14.1\% & 99.4\% & 62.8\% \\
    \hlc[Wheat]{Face Generation Attack} \hlc[CadetBlueLight]{BadNets} $\alpha$=1.0 & \textcolor{gray!60}{Benign} & \textcolor{gray!60}{Benign} & 99.5\% & 99.9\% & 12.2 & 1.9 & 95.4\% & 43.2\% & 13.6\% & 99.8\% & 43.1\% \\
    \hlc[Wheat]{Face Generation Attack} \hlc[OrchidLight]{SIG} $\alpha$=0.16 & \textcolor{gray!60}{Benign} & \textcolor{gray!60}{Benign} & 99.5\% & 92.5\% & 11.9 & 31.1 & 95.4\% & 71.3\% & 14.0\% & 93.8\% & 61.9\% \\
    \hlc[Wheat]{Face Generation Attack} \hlc[OrchidLight]{SIG} $\alpha$=0.3 & \textcolor{gray!60}{Benign} & \textcolor{gray!60}{Benign} & 99.5\% & 97.8\% & 12.6 & 9.4 & 95.2\% & 74.1\% & 15.5\% & 97.0\% & \textbf{70.3\%} \\
    \hline
    \textcolor{gray!60}{Benign} & \hlc[DandelionLight]{Glasses} & \textcolor{gray!60}{Benign} & 99.6\% & 97.9\% & 16.6 & 22.9 & 20.9\% & 86.4\% & 12.4\% & 80.4\% & 68.0\% \\
    \textcolor{gray!60}{Benign} & \hlc[CadetBlueLight]{BadNets} & \textcolor{gray!60}{Benign} & 99.6\% & 99.5\% & 16.6 & 16.6 & 18.3\% & 40.5\% & 12.2\% & 11.3\% & 4.6\% \\
    \textcolor{gray!60}{Benign} & \hlc[OrchidLight]{SIG} & \textcolor{gray!60}{Benign} & 99.6\% & 98.4\% & 16.6 & 24.1 & 12.9\% & 97.2\% & 12.5\% & 95.7\% & \textbf{91.5\%} \\
    \textcolor{gray!60}{Benign} & \hlc[MelonLight]{TrojanNN} & \textcolor{gray!60}{Benign} & 99.6\% & 99.5\% & 16.6 & 16.7 & 8.0\% & 7.0\% & 13.0\% & 12.2\% & 0.8\% \\
    \hline
    \textcolor{gray!60}{Benign} & \textcolor{gray!60}{Benign} & \hlc[RoyalPurpleLight]{FIBA} & 99.6\% & 98.3\% & 16.6 & 26.7 & 94.4\% & 24.6\% & 13.3\% & 98.6\% & \textbf{23.8\%} \\
    \hline\textcolor{gray!60}{Benign} & \textcolor{gray!60}{Benign} & \hlc[GoldenRodLight]{All-to-One}, \hlc[ApricotLight]{Clean-Label}, \hlc[CadetBlueLight]{BadNets} & 99.6\% & 99.5\% & 16.6 & 16.9 & 94.4\% & 18.8\% & 13.1\% & 14.2\% & 2.7\% \\
    \textcolor{gray!60}{Benign} & \textcolor{gray!60}{Benign} & \hlc[GoldenRodLight]{All-to-One}, \hlc[ForestLight]{Poison-Label}, \hlc[CadetBlueLight]{BadNets} & 99.6\% & 99.5\% & 16.6 & 16.9 & 94.4\% & 18.8\% & 11.4\% & 98.0\% & \textbf{18.3\%} \\
    \textcolor{gray!60}{Benign} & \textcolor{gray!60}{Benign} & \hlc[NavyLight]{Master Face}, \hlc[ApricotLight]{Clean-Label}, \hlc[CadetBlueLight]{BadNets} & 99.6\% & 99.5\% & 16.6 & 16.9 & 94.4\% & 18.8\% & 10.7\% & 10.9\% & 2.0\% \\
    \textcolor{gray!60}{Benign} & \textcolor{gray!60}{Benign} & \hlc[NavyLight]{Master Face}, \hlc[ForestLight]{Poison-Label}, \hlc[CadetBlueLight]{BadNets} & 99.6\% & 99.5\% & 16.6 & 16.9 & 94.4\% & 18.8\% & 10.6\% & 12.4\% & 2.3\% \\
    \hline\textcolor{gray!60}{Benign} & \textcolor{gray!60}{Benign} & \hlc[GoldenRodLight]{All-to-One}, \hlc[ApricotLight]{Clean-Label}, \hlc[OrchidLight]{SIG} & 99.6\% & 98.4\% & 16.6 & 24.1 & 94.4\% & 81.4\% & 11.0\% & 94.1\% & 75.4\% \\
    \textcolor{gray!60}{Benign} & \textcolor{gray!60}{Benign} & \hlc[GoldenRodLight]{All-to-One}, \hlc[ForestLight]{Poison-Label}, \hlc[OrchidLight]{SIG} & 99.6\% & 98.4\% & 16.6 & 24.1 & 94.4\% & 81.4\% & 12.4\% & 97.0\% & \textbf{77.7\%} \\
    \textcolor{gray!60}{Benign} & \textcolor{gray!60}{Benign} & \hlc[NavyLight]{Master Face}, \hlc[ApricotLight]{Clean-Label}, \hlc[OrchidLight]{SIG} & 99.6\% & 98.4\% & 16.6 & 24.1 & 94.4\% & 81.4\% & 10.1\% & 82.5\% & 66.1\% \\
    \textcolor{gray!60}{Benign} & \textcolor{gray!60}{Benign} & \hlc[NavyLight]{Master Face}, \hlc[ForestLight]{Poison-Label}, \hlc[OrchidLight]{SIG} & 99.6\% & 98.4\% & 16.6 & 24.1 & 94.4\% & 81.4\% & 9.1\% & 94.9\% & 76.0\% \\
    
  \end{tabular}
\end{table*}

\begin{table*}
  \centering
  \caption{\acl{ba} \acl{sr} in an \textbf{\acl{a2o}} threat model context, targeting a pipeline consisting of:\\ResNet50, MobileNetV2, GhostFaceNetV2.\\\textbf{Note}: FMR$^\pois$ metrics in \textcolor{BrickRed}{red} indicate a DNN with a collapsed performance after inclusion in the \ac{frs} (\ie, all identities match as in an untargeted poisoning attack); the Landmark Shift Attack SIG $\alpha=0.16$ model did not converge and is not reported.}
  \label{tab:backdoor_bench_RN50_MN2_GFN}
  \vspace{0.0cm}
  \footnotesize
  \setlength{\tabcolsep}{1.55pt}
  \renewcommand{\arraystretch}{1.5}
  \begin{tabular}{@{}lllccccccccc@{}}
    
     &  &  & \multicolumn{4}{c}{\textbf{Detector}} & \multicolumn{2}{c}{\textbf{Antispoofer}} & \multicolumn{2}{c}{\textbf{Extractor}} & \\
     \textbf{Detector} & \textbf{Antispoofer} & \textbf{Extractor} & \multicolumn{4}{c}{\textbf{metrics}} & \multicolumn{2}{c}{\textbf{metrics}} & \multicolumn{2}{c}{\textbf{metrics}} & \textbf{Survival} \\
     ResNet50 & MobileNetV2 & GhostFaceNetV2 & AP$^\clean$ & AP$^\pois$ & LS$^\clean$ & LS$^\pois$ & FRR$^\clean$ & FAR$^\pois$ & FRR$^\clean$ & FAR$^\pois$ & \textbf{Rate} \\
    \hline 
    \textcolor{gray!60}{Benign} & \textcolor{gray!60}{Benign} & \textcolor{gray!60}{Benign} & 99.6\% & \textcolor{gray!30}{$\varnothing$} & 16.6 & \textcolor{gray!30}{$\varnothing$} & 83.6\% & \textcolor{gray!30}{$\varnothing$} & 4.5\% & \textcolor{gray!30}{$\varnothing$} & \textcolor{gray!30}{$\varnothing$} \\
    \hline
    \hlc[Slate]{Landmark Shift Attack} \hlc[CadetBlueLight]{BadNets} $\alpha$=0.5 & \textcolor{gray!60}{Benign} & \textcolor{gray!60}{Benign} & 99.5\% & 99.5\% & 12.0 & 153.4 & 84.4\% & 0.7\% & 4.7\% & 40.1\% & 0.3\% \\
    \hlc[Slate]{Landmark Shift Attack} \hlc[CadetBlueLight]{BadNets} $\alpha$=1.0 & \textcolor{gray!60}{Benign} & \textcolor{gray!60}{Benign} & 99.5\% & 99.6\% & 12.1 & 143.4 & 84.5\% & 1.3\% & 4.5\% & 33.0\% & 0.4\% \\
    \hlc[Slate]{Landmark Shift Attack} \hlc[OrchidLight]{SIG} $\alpha$=0.3 & \textcolor{gray!60}{Benign} & \textcolor{gray!60}{Benign} & 99.4\% & 96.8\% & 12.6 & 156.1 & 84.5\% & 2.3\% & 4.8\% & 45.3\% & \textbf{1.0\%} \\
    \hline\hlc[Wheat]{Face Generation Attack} \hlc[CadetBlueLight]{BadNets} $\alpha$=0.5 & \textcolor{gray!60}{Benign} & \textcolor{gray!60}{Benign} & 99.5\% & 99.5\% & 12.0 & 3.1 & 86.1\% & 9.2\% & 4.6\% & 95.2\% & 8.7\% \\
    \hlc[Wheat]{Face Generation Attack} \hlc[CadetBlueLight]{BadNets} $\alpha$=1.0 & \textcolor{gray!60}{Benign} & \textcolor{gray!60}{Benign} & 99.5\% & 99.9\% & 12.2 & 1.6 & 84.7\% & 3.8\% & 4.5\% & 98.7\% & 3.7\% \\
    \hlc[Wheat]{Face Generation Attack} \hlc[OrchidLight]{SIG} $\alpha$=0.16 & \textcolor{gray!60}{Benign} & \textcolor{gray!60}{Benign} & 99.5\% & 91.9\% & 11.9 & 32.0 & 84.0\% & 90.9\% & 4.6\% & 93.8\% & 78.4\% \\
    \hlc[Wheat]{Face Generation Attack} \hlc[OrchidLight]{SIG} $\alpha$=0.3 & \textcolor{gray!60}{Benign} & \textcolor{gray!60}{Benign} & 99.5\% & 98.2\% & 12.6 & 8.7 & 87.5\% & 85.0\% & 4.7\% & 99.2\% & \textbf{82.8\%} \\
    \hline
    \textcolor{gray!60}{Benign} & \hlc[DandelionLight]{Glasses} & \textcolor{gray!60}{Benign} & 99.6\% & 97.9\% & 16.6 & 22.9 & 32.5\% & 69.0\% & 4.3\% & 45.7\% & 30.9\% \\
    \textcolor{gray!60}{Benign} & \hlc[CadetBlueLight]{BadNets} & \textcolor{gray!60}{Benign} & 99.6\% & 99.5\% & 16.6 & 16.6 & 13.4\% & 72.1\% & 4.6\% & 4.4\% & 3.2\% \\
    \textcolor{gray!60}{Benign} & \hlc[OrchidLight]{SIG} & \textcolor{gray!60}{Benign} & 99.6\% & 98.4\% & 16.6 & 24.1 & 10.7\% & 87.4\% & 4.5\% & 63.8\% & \textbf{54.9\%} \\
    \textcolor{gray!60}{Benign} & \hlc[MelonLight]{TrojanNN} & \textcolor{gray!60}{Benign} & 99.6\% & 99.5\% & 16.6 & 16.6 & 17.0\% & 22.6\% & 4.5\% & 4.4\% & 1.0\% \\
    \hline
    \textcolor{gray!60}{Benign} & \textcolor{gray!60}{Benign} & \hlc[RoyalPurpleLight]{FIBA} & 99.6\% & 98.3\% & 16.6 & 26.7 & 83.6\% & 27.0\% & 4.5\% & 96.3\% & \textbf{25.6\%} \\
    \hline\textcolor{gray!60}{Benign} & \textcolor{gray!60}{Benign} & \hlc[GoldenRodLight]{All-to-One}, \hlc[ApricotLight]{Clean-Label}, \hlc[CadetBlueLight]{BadNets} & 99.6\% & 99.5\% & 16.6 & 16.9 & 83.6\% & 20.4\% & 8.1\% & 8.1\% & 1.6\% \\
    \textcolor{gray!60}{Benign} & \textcolor{gray!60}{Benign} & \hlc[GoldenRodLight]{All-to-One}, \hlc[ForestLight]{Poison-Label}, \hlc[CadetBlueLight]{BadNets} & 99.6\% & 99.5\% & 16.6 & 16.9 & 83.6\% & 20.4\% & 4.2\% & 98.3\% & \textbf{20.0\%} \\
    \textcolor{gray!60}{Benign} & \textcolor{gray!60}{Benign} & \hlc[NavyLight]{Master Face}, \hlc[ApricotLight]{Clean-Label}, \hlc[CadetBlueLight]{BadNets} & 99.6\% & 99.5\% & 16.6 & 16.9 & 83.6\% & 20.4\% & 4.2\% & 4.2\% & 0.9\% \\
    \textcolor{gray!60}{Benign} & \textcolor{gray!60}{Benign} & \hlc[NavyLight]{Master Face}, \hlc[ForestLight]{Poison-Label}, \hlc[CadetBlueLight]{BadNets} & 99.6\% & 99.5\% & 16.6 & 16.9 & 83.6\% & 20.4\% & \textcolor{BrickRed}{98.4\%} & \textcolor{BrickRed}{98.3\%} & \textbf{20.0\%} \\
    \hline\textcolor{gray!60}{Benign} & \textcolor{gray!60}{Benign} & \hlc[GoldenRodLight]{All-to-One}, \hlc[ApricotLight]{Clean-Label}, \hlc[OrchidLight]{SIG} & 99.6\% & 98.4\% & 16.6 & 24.1 & 83.6\% & 68.6\% & 4.9\% & 74.6\% & 50.4\% \\
    \textcolor{gray!60}{Benign} & \textcolor{gray!60}{Benign} & \hlc[GoldenRodLight]{All-to-One}, \hlc[ForestLight]{Poison-Label}, \hlc[OrchidLight]{SIG} & 99.6\% & 98.4\% & 16.6 & 24.1 & 83.6\% & 68.6\% & 6.1\% & 98.3\% & 66.4\% \\
    \textcolor{gray!60}{Benign} & \textcolor{gray!60}{Benign} & \hlc[NavyLight]{Master Face}, \hlc[ApricotLight]{Clean-Label}, \hlc[OrchidLight]{SIG} & 99.6\% & 98.4\% & 16.6 & 24.1 & 83.6\% & 68.6\% & 5.5\% & 19.2\% & 13.0\% \\
    \textcolor{gray!60}{Benign} & \textcolor{gray!60}{Benign} & \hlc[NavyLight]{Master Face}, \hlc[ForestLight]{Poison-Label}, \hlc[OrchidLight]{SIG} & 99.6\% & 98.4\% & 16.6 & 24.1 & 83.6\% & 68.6\% & 5.1\% & 98.5\% & \textbf{66.5\%} \\
    
  \end{tabular}
\end{table*}

\begin{table*}
  \centering
  \caption{\acl{ba} \acl{sr} in an \textbf{\acl{a2o}} threat model context, targeting a pipeline consisting of:\\ResNet50, MobileNetV2, IRSE50.\\\textbf{Note}: The Landmark Shift Attack SIG $\alpha=0.16$ model did not converge and is not reported.}
  \label{tab:backdoor_bench_RN50_MN2_IRSE}
  \vspace{0.0cm}
  \footnotesize
  \setlength{\tabcolsep}{1.55pt}
  \renewcommand{\arraystretch}{1.5}
  \begin{tabular}{@{}lllccccccccc@{}}
    
     &  &  & \multicolumn{4}{c}{\textbf{Detector}} & \multicolumn{2}{c}{\textbf{Antispoofer}} & \multicolumn{2}{c}{\textbf{Extractor}} & \\
     \textbf{Detector} & \textbf{Antispoofer} & \textbf{Extractor} & \multicolumn{4}{c}{\textbf{metrics}} & \multicolumn{2}{c}{\textbf{metrics}} & \multicolumn{2}{c}{\textbf{metrics}} & \textbf{Survival} \\
     ResNet50 & MobileNetV2 & IRSE50 & AP$^\clean$ & AP$^\pois$ & LS$^\clean$ & LS$^\pois$ & FRR$^\clean$ & FAR$^\pois$ & FRR$^\clean$ & FAR$^\pois$ & \textbf{Rate} \\
    \hline 
    \textcolor{gray!60}{Benign} & \textcolor{gray!60}{Benign} & \textcolor{gray!60}{Benign} & 99.6\% & \textcolor{gray!30}{$\varnothing$} & 16.6 & \textcolor{gray!30}{$\varnothing$} & 83.6\% & \textcolor{gray!30}{$\varnothing$} & 0.8\% & \textcolor{gray!30}{$\varnothing$} & \textcolor{gray!30}{$\varnothing$} \\
    \hline
    \hlc[Slate]{Landmark Shift Attack} \hlc[CadetBlueLight]{BadNets} $\alpha$=0.5 & \textcolor{gray!60}{Benign} & \textcolor{gray!60}{Benign} & 99.5\% & 99.5\% & 12.0 & 153.4 & 84.4\% & 0.7\% & 0.8\% & 33.9\% & 0.2\% \\
    \hlc[Slate]{Landmark Shift Attack} \hlc[CadetBlueLight]{BadNets} $\alpha$=1.0 & \textcolor{gray!60}{Benign} & \textcolor{gray!60}{Benign} & 99.5\% & 99.6\% & 12.1 & 143.4 & 84.5\% & 1.3\% & 0.7\% & 24.6\% & 0.3\% \\
    \hlc[Slate]{Landmark Shift Attack} \hlc[OrchidLight]{SIG} $\alpha$=0.3 & \textcolor{gray!60}{Benign} & \textcolor{gray!60}{Benign} & 99.4\% & 96.8\% & 12.6 & 156.1 & 84.5\% & 2.3\% & 0.8\% & 58.6\% & \textbf{1.3\%} \\
    \hline\hlc[Wheat]{Face Generation Attack} \hlc[CadetBlueLight]{BadNets} $\alpha$=0.5 & \textcolor{gray!60}{Benign} & \textcolor{gray!60}{Benign} & 99.5\% & 99.6\% & 12.0 & 2.9 & 86.1\% & 9.0\% & 0.8\% & 94.1\% & 8.4\% \\
    \hlc[Wheat]{Face Generation Attack} \hlc[CadetBlueLight]{BadNets} $\alpha$=1.0 & \textcolor{gray!60}{Benign} & \textcolor{gray!60}{Benign} & 99.5\% & 99.9\% & 12.2 & 1.9 & 84.7\% & 3.6\% & 0.7\% & 96.0\% & 3.5\% \\
    \hlc[Wheat]{Face Generation Attack} \hlc[OrchidLight]{SIG} $\alpha$=0.16 & \textcolor{gray!60}{Benign} & \textcolor{gray!60}{Benign} & 99.5\% & 91.7\% & 11.9 & 31.9 & 84.0\% & 90.3\% & 0.8\% & 91.7\% & 75.9\% \\
    \hlc[Wheat]{Face Generation Attack} \hlc[OrchidLight]{SIG} $\alpha$=0.3 & \textcolor{gray!60}{Benign} & \textcolor{gray!60}{Benign} & 99.5\% & 98.0\% & 12.6 & 8.9 & 87.5\% & 85.4\% & 0.8\% & 98.9\% & \textbf{82.8\%} \\
    \hline
    \textcolor{gray!60}{Benign} & \hlc[DandelionLight]{Glasses} & \textcolor{gray!60}{Benign} & 99.6\% & 97.9\% & 16.6 & 22.9 & 32.5\% & 69.0\% & 0.7\% & 25.6\% & 17.3\% \\
    \textcolor{gray!60}{Benign} & \hlc[CadetBlueLight]{BadNets} & \textcolor{gray!60}{Benign} & 99.6\% & 99.5\% & 16.6 & 16.6 & 13.4\% & 70.8\% & 0.7\% & 0.5\% & 0.4\% \\
    \textcolor{gray!60}{Benign} & \hlc[OrchidLight]{SIG} & \textcolor{gray!60}{Benign} & 99.6\% & 98.4\% & 16.6 & 24.1 & 10.7\% & 87.4\% & 0.7\% & 21.5\% & \textbf{18.5\%} \\
    \textcolor{gray!60}{Benign} & \hlc[MelonLight]{TrojanNN} & \textcolor{gray!60}{Benign} & 99.6\% & 99.5\% & 16.6 & 16.7 & 17.0\% & 22.0\% & 0.7\% & 0.7\% & 0.2\% \\
    \hline
    \textcolor{gray!60}{Benign} & \textcolor{gray!60}{Benign} & \hlc[RoyalPurpleLight]{FIBA} & 99.6\% & 98.3\% & 16.6 & 26.7 & 83.6\% & 27.0\% & 0.8\% & 92.8\% & \textbf{24.6\%} \\
    \hline\textcolor{gray!60}{Benign} & \textcolor{gray!60}{Benign} & \hlc[GoldenRodLight]{All-to-One}, \hlc[ApricotLight]{Clean-Label}, \hlc[CadetBlueLight]{BadNets} & 99.6\% & 99.5\% & 16.6 & 16.9 & 83.6\% & 20.4\% & 36.5\% & 40.9\% & 8.3\% \\
    \textcolor{gray!60}{Benign} & \textcolor{gray!60}{Benign} & \hlc[GoldenRodLight]{All-to-One}, \hlc[ForestLight]{Poison-Label}, \hlc[CadetBlueLight]{BadNets} & 99.6\% & 99.5\% & 16.6 & 16.9 & 83.6\% & 20.4\% & 0.9\% & 97.4\% & \textbf{19.8\%} \\
    \textcolor{gray!60}{Benign} & \textcolor{gray!60}{Benign} & \hlc[NavyLight]{Master Face}, \hlc[ApricotLight]{Clean-Label}, \hlc[CadetBlueLight]{BadNets} & 99.6\% & 99.5\% & 16.6 & 16.9 & 83.6\% & 20.4\% & 0.5\% & 0.5\% & 0.1\% \\
    \textcolor{gray!60}{Benign} & \textcolor{gray!60}{Benign} & \hlc[NavyLight]{Master Face}, \hlc[ForestLight]{Poison-Label}, \hlc[CadetBlueLight]{BadNets} & 99.6\% & 99.5\% & 16.6 & 16.9 & 83.6\% & 20.4\% & 0.6\% & 0.6\% & 0.1\% \\
    \hline\textcolor{gray!60}{Benign} & \textcolor{gray!60}{Benign} & \hlc[GoldenRodLight]{All-to-One}, \hlc[ApricotLight]{Clean-Label}, \hlc[OrchidLight]{SIG} & 99.6\% & 98.4\% & 16.6 & 24.1 & 83.6\% & 68.6\% & 0.6\% & 36.5\% & 24.6\% \\
    \textcolor{gray!60}{Benign} & \textcolor{gray!60}{Benign} & \hlc[GoldenRodLight]{All-to-One}, \hlc[ForestLight]{Poison-Label}, \hlc[OrchidLight]{SIG} & 99.6\% & 98.4\% & 16.6 & 24.1 & 83.6\% & 68.6\% & 0.6\% & 51.6\% & 34.8\% \\
    \textcolor{gray!60}{Benign} & \textcolor{gray!60}{Benign} & \hlc[NavyLight]{Master Face}, \hlc[ApricotLight]{Clean-Label}, \hlc[OrchidLight]{SIG} & 99.6\% & 98.4\% & 16.6 & 24.1 & 83.6\% & 68.6\% & 0.6\% & 12.7\% & 8.6\% \\
    \textcolor{gray!60}{Benign} & \textcolor{gray!60}{Benign} & \hlc[NavyLight]{Master Face}, \hlc[ForestLight]{Poison-Label}, \hlc[OrchidLight]{SIG} & 99.6\% & 98.4\% & 16.6 & 24.1 & 83.6\% & 68.6\% & 0.6\% & 93.9\% & \textbf{63.4\%} \\
    
  \end{tabular}
\end{table*}

\begin{table*}
  \centering
  \caption{\acl{ba} \acl{sr} in an \textbf{\acl{a2o}} threat model context, targeting a pipeline consisting of:\\ResNet50, MobileNetV2, MobileFaceNet.\\\textbf{Note}: The Landmark Shift Attack SIG $\alpha=0.16$ model did not converge and is not reported.}
  \label{tab:backdoor_bench_RN50_MN2_MFN}
  \vspace{0.0cm}
  \footnotesize
  \setlength{\tabcolsep}{1.55pt}
  \renewcommand{\arraystretch}{1.5}
  \begin{tabular}{@{}lllccccccccc@{}}
    
     &  &  & \multicolumn{4}{c}{\textbf{Detector}} & \multicolumn{2}{c}{\textbf{Antispoofer}} & \multicolumn{2}{c}{\textbf{Extractor}} & \\
     \textbf{Detector} & \textbf{Antispoofer} & \textbf{Extractor} & \multicolumn{4}{c}{\textbf{metrics}} & \multicolumn{2}{c}{\textbf{metrics}} & \multicolumn{2}{c}{\textbf{metrics}} & \textbf{Survival} \\
     ResNet50 & MobileNetV2 & MobileFaceNet & AP$^\clean$ & AP$^\pois$ & LS$^\clean$ & LS$^\pois$ & FRR$^\clean$ & FAR$^\pois$ & FRR$^\clean$ & FAR$^\pois$ & \textbf{Rate} \\
    \hline 
    \textcolor{gray!60}{Benign} & \textcolor{gray!60}{Benign} & \textcolor{gray!60}{Benign} & 99.6\% & \textcolor{gray!30}{$\varnothing$} & 16.6 & \textcolor{gray!30}{$\varnothing$} & 83.6\% & \textcolor{gray!30}{$\varnothing$} & 4.0\% & \textcolor{gray!30}{$\varnothing$} & \textcolor{gray!30}{$\varnothing$} \\
    \hline
    \hlc[Slate]{Landmark Shift Attack} \hlc[CadetBlueLight]{BadNets} $\alpha$=0.5 & \textcolor{gray!60}{Benign} & \textcolor{gray!60}{Benign} & 99.5\% & 99.5\% & 12.0 & 153.4 & 84.4\% & 0.7\% & 4.0\% & 65.7\% & 0.5\% \\
    \hlc[Slate]{Landmark Shift Attack} \hlc[CadetBlueLight]{BadNets} $\alpha$=1.0 & \textcolor{gray!60}{Benign} & \textcolor{gray!60}{Benign} & 99.5\% & 99.6\% & 12.1 & 143.4 & 84.5\% & 1.3\% & 3.8\% & 63.3\% & 0.8\% \\
    \hlc[Slate]{Landmark Shift Attack} \hlc[OrchidLight]{SIG} $\alpha$=0.3 & \textcolor{gray!60}{Benign} & \textcolor{gray!60}{Benign} & 99.4\% & 96.8\% & 12.6 & 156.1 & 84.5\% & 2.3\% & 4.0\% & 63.5\% & \textbf{1.4\%} \\
    \hline\hlc[Wheat]{Face Generation Attack} \hlc[CadetBlueLight]{BadNets} $\alpha$=0.5 & \textcolor{gray!60}{Benign} & \textcolor{gray!60}{Benign} & 99.5\% & 99.5\% & 12.0 & 2.8 & 86.1\% & 9.0\% & 3.9\% & 95.9\% & 8.6\% \\
    \hlc[Wheat]{Face Generation Attack} \hlc[CadetBlueLight]{BadNets} $\alpha$=1.0 & \textcolor{gray!60}{Benign} & \textcolor{gray!60}{Benign} & 99.5\% & 99.9\% & 12.2 & 1.9 & 84.7\% & 4.6\% & 3.8\% & 95.8\% & 4.4\% \\
    \hlc[Wheat]{Face Generation Attack} \hlc[OrchidLight]{SIG} $\alpha$=0.16 & \textcolor{gray!60}{Benign} & \textcolor{gray!60}{Benign} & 99.5\% & 91.9\% & 11.9 & 32.6 & 84.0\% & 90.8\% & 3.9\% & 93.8\% & 78.3\% \\
    \hlc[Wheat]{Face Generation Attack} \hlc[OrchidLight]{SIG} $\alpha$=0.3 & \textcolor{gray!60}{Benign} & \textcolor{gray!60}{Benign} & 99.5\% & 98.0\% & 12.6 & 8.9 & 87.5\% & 86.1\% & 4.2\% & 98.6\% & \textbf{83.2\%} \\
    \hline
    \textcolor{gray!60}{Benign} & \hlc[DandelionLight]{Glasses} & \textcolor{gray!60}{Benign} & 99.6\% & 97.9\% & 16.6 & 22.9 & 32.5\% & 69.0\% & 3.3\% & 60.4\% & 40.8\% \\
    \textcolor{gray!60}{Benign} & \hlc[CadetBlueLight]{BadNets} & \textcolor{gray!60}{Benign} & 99.6\% & 99.5\% & 16.6 & 16.6 & 13.4\% & 71.4\% & 3.6\% & 3.0\% & 2.1\% \\
    \textcolor{gray!60}{Benign} & \hlc[OrchidLight]{SIG} & \textcolor{gray!60}{Benign} & 99.6\% & 98.4\% & 16.6 & 24.1 & 10.7\% & 87.4\% & 3.6\% & 90.6\% & \textbf{77.9\%} \\
    \textcolor{gray!60}{Benign} & \hlc[MelonLight]{TrojanNN} & \textcolor{gray!60}{Benign} & 99.6\% & 99.5\% & 16.6 & 16.7 & 17.0\% & 23.0\% & 3.6\% & 3.5\% & 0.8\% \\
    \hline
    \textcolor{gray!60}{Benign} & \textcolor{gray!60}{Benign} & \hlc[RoyalPurpleLight]{FIBA} & 99.6\% & 98.3\% & 16.6 & 26.7 & 83.6\% & 27.0\% & 4.0\% & 98.8\% & \textbf{26.2\%} \\
    \hline\textcolor{gray!60}{Benign} & \textcolor{gray!60}{Benign} & \hlc[GoldenRodLight]{All-to-One}, \hlc[ApricotLight]{Clean-Label}, \hlc[CadetBlueLight]{BadNets} & 99.6\% & 99.5\% & 16.6 & 16.9 & 83.6\% & 20.4\% & 4.1\% & 4.4\% & 0.9\% \\
    \textcolor{gray!60}{Benign} & \textcolor{gray!60}{Benign} & \hlc[GoldenRodLight]{All-to-One}, \hlc[ForestLight]{Poison-Label}, \hlc[CadetBlueLight]{BadNets} & 99.6\% & 99.5\% & 16.6 & 16.9 & 83.6\% & 20.4\% & 2.6\% & 98.2\% & \textbf{19.9\%} \\
    \textcolor{gray!60}{Benign} & \textcolor{gray!60}{Benign} & \hlc[NavyLight]{Master Face}, \hlc[ApricotLight]{Clean-Label}, \hlc[CadetBlueLight]{BadNets} & 99.6\% & 99.5\% & 16.6 & 16.9 & 83.6\% & 20.4\% & 3.3\% & 3.5\% & 0.7\% \\
    \textcolor{gray!60}{Benign} & \textcolor{gray!60}{Benign} & \hlc[NavyLight]{Master Face}, \hlc[ForestLight]{Poison-Label}, \hlc[CadetBlueLight]{BadNets} & 99.6\% & 99.5\% & 16.6 & 16.9 & 83.6\% & 20.4\% & 2.3\% & 2.9\% & 0.6\% \\
    \hline\textcolor{gray!60}{Benign} & \textcolor{gray!60}{Benign} & \hlc[GoldenRodLight]{All-to-One}, \hlc[ApricotLight]{Clean-Label}, \hlc[RedVioletLight]{Mask} & 99.6\% & 99.2\% & 16.6 & 29.7 & 83.6\% & 23.3\% & 3.0\% & 98.9\% & \textbf{22.9\%} \\
    \textcolor{gray!60}{Benign} & \textcolor{gray!60}{Benign} & \hlc[GoldenRodLight]{All-to-One}, \hlc[ForestLight]{Poison-Label}, \hlc[RedVioletLight]{Mask} & 99.6\% & 99.2\% & 16.6 & 29.7 & 83.6\% & 23.3\% & 2.9\% & 99.2\% & \textbf{22.9\%} \\
    \textcolor{gray!60}{Benign} & \textcolor{gray!60}{Benign} & \hlc[NavyLight]{Master Face}, \hlc[ApricotLight]{Clean-Label}, \hlc[RedVioletLight]{Mask} & 99.6\% & 99.2\% & 16.6 & 29.7 & 83.6\% & 23.3\% & 2.8\% & 20.7\% & 4.8\% \\
    \textcolor{gray!60}{Benign} & \textcolor{gray!60}{Benign} & \hlc[NavyLight]{Master Face}, \hlc[ForestLight]{Poison-Label}, \hlc[RedVioletLight]{Mask} & 99.6\% & 99.2\% & 16.6 & 29.7 & 83.6\% & 23.3\% & 3.4\% & 59.9\% & 13.8\% \\
    \hline\textcolor{gray!60}{Benign} & \textcolor{gray!60}{Benign} & \hlc[GoldenRodLight]{All-to-One}, \hlc[ApricotLight]{Clean-Label}, \hlc[OrchidLight]{SIG} & 99.6\% & 98.4\% & 16.6 & 24.1 & 83.6\% & 68.6\% & 3.5\% & 95.4\% & 64.4\% \\
    \textcolor{gray!60}{Benign} & \textcolor{gray!60}{Benign} & \hlc[GoldenRodLight]{All-to-One}, \hlc[ForestLight]{Poison-Label}, \hlc[OrchidLight]{SIG} & 99.6\% & 98.4\% & 16.6 & 24.1 & 83.6\% & 68.6\% & 3.2\% & 98.0\% & \textbf{66.2\%} \\
    \textcolor{gray!60}{Benign} & \textcolor{gray!60}{Benign} & \hlc[NavyLight]{Master Face}, \hlc[ApricotLight]{Clean-Label}, \hlc[OrchidLight]{SIG} & 99.6\% & 98.4\% & 16.6 & 24.1 & 83.6\% & 68.6\% & 3.4\% & 71.1\% & 48.0\% \\
    \textcolor{gray!60}{Benign} & \textcolor{gray!60}{Benign} & \hlc[NavyLight]{Master Face}, \hlc[ForestLight]{Poison-Label}, \hlc[OrchidLight]{SIG} & 99.6\% & 98.4\% & 16.6 & 24.1 & 83.6\% & 68.6\% & 3.8\% & 55.5\% & 37.5\% \\
    
  \end{tabular}
\end{table*}

\begin{table*}
  \centering
  \caption{\acl{ba} \acl{sr} in an \textbf{\acl{a2o}} threat model context, targeting a pipeline consisting of:\\ResNet50, MobileNetV2, ResNet50.\\\textbf{Note}: The Landmark Shift Attack SIG $\alpha=0.16$ model did not converge and is not reported.}
  \label{tab:backdoor_bench_RN50_MN2_RN50}
  \vspace{0.0cm}
  \footnotesize
  \setlength{\tabcolsep}{1.55pt}
  \renewcommand{\arraystretch}{1.5}
  \begin{tabular}{@{}lllccccccccc@{}}
    
     &  &  & \multicolumn{4}{c}{\textbf{Detector}} & \multicolumn{2}{c}{\textbf{Antispoofer}} & \multicolumn{2}{c}{\textbf{Extractor}} & \\
     \textbf{Detector} & \textbf{Antispoofer} & \textbf{Extractor} & \multicolumn{4}{c}{\textbf{metrics}} & \multicolumn{2}{c}{\textbf{metrics}} & \multicolumn{2}{c}{\textbf{metrics}} & \textbf{Survival} \\
     ResNet50 & MobileNetV2 & ResNet50 & AP$^\clean$ & AP$^\pois$ & LS$^\clean$ & LS$^\pois$ & FRR$^\clean$ & FAR$^\pois$ & FRR$^\clean$ & FAR$^\pois$ & \textbf{Rate} \\
    \hline 
    \textcolor{gray!60}{Benign} & \textcolor{gray!60}{Benign} & \textcolor{gray!60}{Benign} & 99.6\% & \textcolor{gray!30}{$\varnothing$} & 16.6 & \textcolor{gray!30}{$\varnothing$} & 83.6\% & \textcolor{gray!30}{$\varnothing$} & 1.8\% & \textcolor{gray!30}{$\varnothing$} & \textcolor{gray!30}{$\varnothing$} \\
    \hline
    \hlc[Slate]{Landmark Shift Attack} \hlc[CadetBlueLight]{BadNets} $\alpha$=0.5 & \textcolor{gray!60}{Benign} & \textcolor{gray!60}{Benign} & 99.5\% & 99.5\% & 12.0 & 153.4 & 84.4\% & 0.7\% & 1.9\% & 21.1\% & 0.1\% \\
    \hlc[Slate]{Landmark Shift Attack} \hlc[CadetBlueLight]{BadNets} $\alpha$=1.0 & \textcolor{gray!60}{Benign} & \textcolor{gray!60}{Benign} & 99.5\% & 99.6\% & 12.1 & 143.4 & 84.5\% & 1.3\% & 1.8\% & 20.7\% & 0.3\% \\
    \hlc[Slate]{Landmark Shift Attack} \hlc[OrchidLight]{SIG} $\alpha$=0.3 & \textcolor{gray!60}{Benign} & \textcolor{gray!60}{Benign} & 99.4\% & 96.8\% & 12.6 & 156.1 & 84.5\% & 2.3\% & 1.9\% & 57.0\% & \textbf{1.3\%} \\
    \hline\hlc[Wheat]{Face Generation Attack} \hlc[CadetBlueLight]{BadNets} $\alpha$=0.5 & \textcolor{gray!60}{Benign} & \textcolor{gray!60}{Benign} & 99.5\% & 99.5\% & 12.0 & 3.2 & 86.1\% & 9.1\% & 1.8\% & 93.7\% & 8.5\% \\
    \hlc[Wheat]{Face Generation Attack} \hlc[CadetBlueLight]{BadNets} $\alpha$=1.0 & \textcolor{gray!60}{Benign} & \textcolor{gray!60}{Benign} & 99.5\% & 99.9\% & 12.2 & 1.6 & 84.7\% & 4.0\% & 1.8\% & 97.6\% & 3.9\% \\
    \hlc[Wheat]{Face Generation Attack} \hlc[OrchidLight]{SIG} $\alpha$=0.16 & \textcolor{gray!60}{Benign} & \textcolor{gray!60}{Benign} & 99.5\% & 92.1\% & 11.9 & 32.4 & 84.0\% & 89.4\% & 1.8\% & 93.4\% & 76.9\% \\
    \hlc[Wheat]{Face Generation Attack} \hlc[OrchidLight]{SIG} $\alpha$=0.3 & \textcolor{gray!60}{Benign} & \textcolor{gray!60}{Benign} & 99.5\% & 98.0\% & 12.6 & 8.7 & 87.5\% & 86.7\% & 2.0\% & 99.0\% & \textbf{84.1\%} \\
    \hline
    \textcolor{gray!60}{Benign} & \hlc[DandelionLight]{Glasses} & \textcolor{gray!60}{Benign} & 99.6\% & 97.9\% & 16.6 & 22.9 & 32.5\% & 69.0\% & 1.4\% & 29.4\% & 19.9\% \\
    \textcolor{gray!60}{Benign} & \hlc[CadetBlueLight]{BadNets} & \textcolor{gray!60}{Benign} & 99.6\% & 99.5\% & 16.6 & 16.6 & 13.4\% & 70.8\% & 1.6\% & 1.4\% & 1.0\% \\
    \textcolor{gray!60}{Benign} & \hlc[OrchidLight]{SIG} & \textcolor{gray!60}{Benign} & 99.6\% & 98.4\% & 16.6 & 24.1 & 10.7\% & 87.4\% & 1.4\% & 64.7\% & \textbf{55.6\%} \\
    \textcolor{gray!60}{Benign} & \hlc[MelonLight]{TrojanNN} & \textcolor{gray!60}{Benign} & 99.6\% & 99.5\% & 16.6 & 16.7 & 17.0\% & 22.3\% & 1.4\% & 1.4\% & 0.3\% \\
    \hline
    \textcolor{gray!60}{Benign} & \textcolor{gray!60}{Benign} & \hlc[RoyalPurpleLight]{FIBA} & 99.6\% & 98.3\% & 16.6 & 26.7 & 83.6\% & 27.0\% & 1.8\% & 96.5\% & \textbf{25.6\%} \\
    \hline\textcolor{gray!60}{Benign} & \textcolor{gray!60}{Benign} & \hlc[GoldenRodLight]{All-to-One}, \hlc[ApricotLight]{Clean-Label}, \hlc[CadetBlueLight]{BadNets} & 99.6\% & 99.5\% & 16.6 & 16.9 & 83.6\% & 20.4\% & 1.7\% & 1.4\% & 0.3\% \\
    \textcolor{gray!60}{Benign} & \textcolor{gray!60}{Benign} & \hlc[GoldenRodLight]{All-to-One}, \hlc[ForestLight]{Poison-Label}, \hlc[CadetBlueLight]{BadNets} & 99.6\% & 99.5\% & 16.6 & 16.9 & 83.6\% & 20.4\% & 2.6\% & 98.2\% & \textbf{19.9\%} \\
    \textcolor{gray!60}{Benign} & \textcolor{gray!60}{Benign} & \hlc[NavyLight]{Master Face}, \hlc[ApricotLight]{Clean-Label}, \hlc[CadetBlueLight]{BadNets} & 99.6\% & 99.5\% & 16.6 & 16.9 & 83.6\% & 20.4\% & 2.0\% & 1.8\% & 0.4\% \\
    \textcolor{gray!60}{Benign} & \textcolor{gray!60}{Benign} & \hlc[NavyLight]{Master Face}, \hlc[ForestLight]{Poison-Label}, \hlc[CadetBlueLight]{BadNets} & 99.6\% & 99.5\% & 16.6 & 16.9 & 83.6\% & 20.4\% & 2.0\% & 1.8\% & 0.4\% \\
    \hline\textcolor{gray!60}{Benign} & \textcolor{gray!60}{Benign} & \hlc[GoldenRodLight]{All-to-One}, \hlc[ApricotLight]{Clean-Label}, \hlc[OrchidLight]{SIG} & 99.6\% & 98.4\% & 16.6 & 24.1 & 83.6\% & 68.6\% & 1.9\% & 77.8\% & 52.5\% \\
    \textcolor{gray!60}{Benign} & \textcolor{gray!60}{Benign} & \hlc[GoldenRodLight]{All-to-One}, \hlc[ForestLight]{Poison-Label}, \hlc[OrchidLight]{SIG} & 99.6\% & 98.4\% & 16.6 & 24.1 & 83.6\% & 68.6\% & 1.6\% & 98.4\% & \textbf{66.4\%} \\
    \textcolor{gray!60}{Benign} & \textcolor{gray!60}{Benign} & \hlc[NavyLight]{Master Face}, \hlc[ApricotLight]{Clean-Label}, \hlc[OrchidLight]{SIG} & 99.6\% & 98.4\% & 16.6 & 24.1 & 83.6\% & 68.6\% & 1.6\% & 40.2\% & 27.1\% \\
    \textcolor{gray!60}{Benign} & \textcolor{gray!60}{Benign} & \hlc[NavyLight]{Master Face}, \hlc[ForestLight]{Poison-Label}, \hlc[OrchidLight]{SIG} & 99.6\% & 98.4\% & 16.6 & 24.1 & 83.6\% & 68.6\% & 1.9\% & 98.4\% & \textbf{66.4\%} \\
    
  \end{tabular}
\end{table*}

\begin{table*}
  \centering
  \caption{\acl{ba} \acl{sr} in an \textbf{\acl{a2o}} threat model context, targeting a pipeline consisting of:\\ResNet50, MobileNetV2, RobFaceNet.\\\textbf{Note}: The Landmark Shift Attack SIG $\alpha=0.16$ model did not converge and is not reported.}
  \label{tab:backdoor_bench_RN50_MN2_RFN}
  \vspace{0.0cm}
  \footnotesize
  \setlength{\tabcolsep}{1.55pt}
  \renewcommand{\arraystretch}{1.5}
  \begin{tabular}{@{}lllccccccccc@{}}
    
     &  &  & \multicolumn{4}{c}{\textbf{Detector}} & \multicolumn{2}{c}{\textbf{Antispoofer}} & \multicolumn{2}{c}{\textbf{Extractor}} & \\
     \textbf{Detector} & \textbf{Antispoofer} & \textbf{Extractor} & \multicolumn{4}{c}{\textbf{metrics}} & \multicolumn{2}{c}{\textbf{metrics}} & \multicolumn{2}{c}{\textbf{metrics}} & \textbf{Survival} \\
     ResNet50 & MobileNetV2 & RobFaceNet & AP$^\clean$ & AP$^\pois$ & LS$^\clean$ & LS$^\pois$ & FRR$^\clean$ & FAR$^\pois$ & FRR$^\clean$ & FAR$^\pois$ & \textbf{Rate} \\
    \hline 
    \textcolor{gray!60}{Benign} & \textcolor{gray!60}{Benign} & \textcolor{gray!60}{Benign} & 99.6\% & \textcolor{gray!30}{$\varnothing$} & 16.6 & \textcolor{gray!30}{$\varnothing$} & 83.6\% & \textcolor{gray!30}{$\varnothing$} & 14.2\% & \textcolor{gray!30}{$\varnothing$} & \textcolor{gray!30}{$\varnothing$} \\
    \hline
    \hlc[Slate]{Landmark Shift Attack} \hlc[CadetBlueLight]{BadNets} $\alpha$=0.5 & \textcolor{gray!60}{Benign} & \textcolor{gray!60}{Benign} & 99.5\% & 99.5\% & 12.0 & 153.4 & 84.4\% & 0.7\% & 15.0\% & 60.7\% & 0.4\% \\
    \hlc[Slate]{Landmark Shift Attack} \hlc[CadetBlueLight]{BadNets} $\alpha$=1.0 & \textcolor{gray!60}{Benign} & \textcolor{gray!60}{Benign} & 99.5\% & 99.6\% & 12.1 & 143.4 & 84.5\% & 1.3\% & 14.2\% & 64.8\% & 0.8\% \\
    \hlc[Slate]{Landmark Shift Attack} \hlc[OrchidLight]{SIG} $\alpha$=0.3 & \textcolor{gray!60}{Benign} & \textcolor{gray!60}{Benign} & 99.4\% & 96.8\% & 12.6 & 156.1 & 84.5\% & 2.3\% & 15.8\% & 71.6\% & \textbf{1.6\%} \\
    \hline\hlc[Wheat]{Face Generation Attack} \hlc[CadetBlueLight]{BadNets} $\alpha$=0.5 & \textcolor{gray!60}{Benign} & \textcolor{gray!60}{Benign} & 99.5\% & 99.5\% & 12.0 & 2.5 & 86.1\% & 8.2\% & 14.7\% & 96.6\% & 7.9\% \\
    \hlc[Wheat]{Face Generation Attack} \hlc[CadetBlueLight]{BadNets} $\alpha$=1.0 & \textcolor{gray!60}{Benign} & \textcolor{gray!60}{Benign} & 99.5\% & 99.9\% & 12.2 & 1.8 & 84.7\% & 3.9\% & 14.4\% & 97.9\% & 3.8\% \\
    \hlc[Wheat]{Face Generation Attack} \hlc[OrchidLight]{SIG} $\alpha$=0.16 & \textcolor{gray!60}{Benign} & \textcolor{gray!60}{Benign} & 99.5\% & 92.5\% & 11.9 & 30.7 & 84.0\% & 91.8\% & 14.8\% & 95.2\% & 80.8\% \\
    \hlc[Wheat]{Face Generation Attack} \hlc[OrchidLight]{SIG} $\alpha$=0.3 & \textcolor{gray!60}{Benign} & \textcolor{gray!60}{Benign} & 99.5\% & 97.8\% & 12.6 & 9.2 & 87.5\% & 85.6\% & 15.9\% & 98.0\% & \textbf{82.0\%} \\
    \hline
    \textcolor{gray!60}{Benign} & \hlc[DandelionLight]{Glasses} & \textcolor{gray!60}{Benign} & 99.6\% & 97.9\% & 16.6 & 22.9 & 32.5\% & 69.0\% & 12.3\% & 79.9\% & 54.0\% \\
    \textcolor{gray!60}{Benign} & \hlc[CadetBlueLight]{BadNets} & \textcolor{gray!60}{Benign} & 99.6\% & 99.5\% & 16.6 & 16.5 & 13.4\% & 71.8\% & 13.3\% & 11.8\% & 8.4\% \\
    \textcolor{gray!60}{Benign} & \hlc[OrchidLight]{SIG} & \textcolor{gray!60}{Benign} & 99.6\% & 98.4\% & 16.6 & 24.1 & 10.7\% & 87.4\% & 13.0\% & 96.6\% & \textbf{83.1\%} \\
    \textcolor{gray!60}{Benign} & \hlc[MelonLight]{TrojanNN} & \textcolor{gray!60}{Benign} & 99.6\% & 99.5\% & 16.6 & 16.7 & 17.0\% & 22.5\% & 13.1\% & 13.9\% & 3.1\% \\
    \hline
    \textcolor{gray!60}{Benign} & \textcolor{gray!60}{Benign} & \hlc[RoyalPurpleLight]{FIBA} & 99.6\% & 98.3\% & 16.6 & 26.7 & 83.6\% & 27.0\% & 14.2\% & 99.1\% & \textbf{26.3\%} \\
    \hline\textcolor{gray!60}{Benign} & \textcolor{gray!60}{Benign} & \hlc[GoldenRodLight]{All-to-One}, \hlc[ApricotLight]{Clean-Label}, \hlc[CadetBlueLight]{BadNets} & 99.6\% & 99.5\% & 16.6 & 16.9 & 83.6\% & 20.4\% & 14.3\% & 14.8\% & 3.0\% \\
    \textcolor{gray!60}{Benign} & \textcolor{gray!60}{Benign} & \hlc[GoldenRodLight]{All-to-One}, \hlc[ForestLight]{Poison-Label}, \hlc[CadetBlueLight]{BadNets} & 99.6\% & 99.5\% & 16.6 & 16.9 & 83.6\% & 20.4\% & 12.2\% & 98.5\% & \textbf{20.0\%} \\
    \textcolor{gray!60}{Benign} & \textcolor{gray!60}{Benign} & \hlc[NavyLight]{Master Face}, \hlc[ApricotLight]{Clean-Label}, \hlc[CadetBlueLight]{BadNets} & 99.6\% & 99.5\% & 16.6 & 16.9 & 83.6\% & 20.4\% & 11.6\% & 11.4\% & 2.3\% \\
    \textcolor{gray!60}{Benign} & \textcolor{gray!60}{Benign} & \hlc[NavyLight]{Master Face}, \hlc[ForestLight]{Poison-Label}, \hlc[CadetBlueLight]{BadNets} & 99.6\% & 99.5\% & 16.6 & 16.9 & 83.6\% & 20.4\% & 11.5\% & 13.0\% & 2.6\% \\
    \hline\textcolor{gray!60}{Benign} & \textcolor{gray!60}{Benign} & \hlc[GoldenRodLight]{All-to-One}, \hlc[ApricotLight]{Clean-Label}, \hlc[OrchidLight]{SIG} & 99.6\% & 98.4\% & 16.6 & 24.1 & 83.6\% & 68.6\% & 11.8\% & 97.0\% & 65.5\% \\
    \textcolor{gray!60}{Benign} & \textcolor{gray!60}{Benign} & \hlc[GoldenRodLight]{All-to-One}, \hlc[ForestLight]{Poison-Label}, \hlc[OrchidLight]{SIG} & 99.6\% & 98.4\% & 16.6 & 24.1 & 83.6\% & 68.6\% & 13.5\% & 98.4\% & \textbf{66.4\%} \\
    \textcolor{gray!60}{Benign} & \textcolor{gray!60}{Benign} & \hlc[NavyLight]{Master Face}, \hlc[ApricotLight]{Clean-Label}, \hlc[OrchidLight]{SIG} & 99.6\% & 98.4\% & 16.6 & 24.1 & 83.6\% & 68.6\% & 10.9\% & 86.2\% & 58.2\% \\
    \textcolor{gray!60}{Benign} & \textcolor{gray!60}{Benign} & \hlc[NavyLight]{Master Face}, \hlc[ForestLight]{Poison-Label}, \hlc[OrchidLight]{SIG} & 99.6\% & 98.4\% & 16.6 & 24.1 & 83.6\% & 68.6\% & 9.7\% & 97.9\% & 66.1\% \\
    
  \end{tabular}
\end{table*}

\end{document}